\documentclass{article}

\usepackage{microtype}
\usepackage{graphicx}
\usepackage{booktabs} 

\usepackage{hyperref}

\usepackage{amsmath}
\usepackage{bm}
\usepackage{amsfonts}
\usepackage{physics}
\usepackage{subfig}
\usepackage{multirow}
\usepackage{amsthm}
\usepackage[accepted]{icml2021}

\newcommand{\thickhat}[1]{\mathbf{\hat{\text{$#1$}}}}

\newcommand{\thicktilde}[1]{\mathbf{\tilde{\text{$#1$}}}}

\newtheorem{remark}{Remark}
\newtheorem{lemma}{Lemma}

\newtheorem{prop}{Proposition}

\newtheorem{innercustomprop}{Proposition}
\newenvironment{customprop}[1]
  {\renewcommand\theinnercustomprop{#1}\innercustomprop}
  {\endinnercustomprop}

\newtheorem{innercustomlemma}{Lemma}
\newenvironment{customlemma}[1]
  {\renewcommand\theinnercustomlemma{#1}\innercustomlemma}
  {\endinnercustomlemma}

\icmltitlerunning{Learning Stochastic Behaviour from Aggregate Data}

\begin{document}

\twocolumn[
\icmltitle{Learning Stochastic Behaviour from Aggregate Data}




\begin{icmlauthorlist}
\icmlauthor{Shaojun Ma}{to}
\icmlauthor{Shu Liu}{to}
\icmlauthor{Hongyuan Zha}{goo}
\icmlauthor{Haomin Zhou}{to}

\end{icmlauthorlist}

\icmlaffiliation{to}{Department of Mathematics, Georgia Institute of Technology, Atlanta, GA 30332, USA}
\icmlaffiliation{goo}{School of Data Science, Shenzhen Research Institute of Big Data, The Chinese University of Hong Kong, Shenzhen, China, the research of Hongyuan Zha is supported in part by a grant from Shenzhen Research Institute of Big Data}

\icmlcorrespondingauthor{Shaojun Ma}{shaojunma@gatech.edu}


\vskip 0.3in
]



\printAffiliationsAndNotice{}  
\begin{abstract}
Learning nonlinear dynamics from aggregate data is a challenging problem because the full trajectory of each individual is not available, namely, the individual observed at one time may not be observed at the next time point, or the identity of individual is unavailable. This is in sharp contrast to learning dynamics with full trajectory data, on which the majority of existing methods are based. We propose a novel method using the weak form of Fokker Planck Equation (FPE) --- a partial differential equation --- to describe the density evolution of data in a sampled form, which is then combined with Wasserstein generative adversarial network (WGAN) in the training process. In such a sample-based framework we are able to learn the nonlinear dynamics from aggregate data without explicitly solving 
FPE. We demonstrate our approach in the context of a series of synthetic and real-world data sets.
\end{abstract}

\section{Introduction}
In the context of dynamical systems, \textbf{Aggregate data} refers to a data format in which the full trajectory of each individual modeled by the evolution of state is not available, 
but rather a sample from the distribution of state at a certain time point is available. Typical examples include data sets collected for DNA evolution, social gathering, density  in control problems, and bird migration, during the evolution of which it is impossible to follow an individual inter-temporally. In those applications, some observed individuals at one time point may be un-observable at the next time spot, or when the individual identities are blocked or unavailable due to various technical and ethical reasons. Rather than inferring the exact information for each individual, the main objective of learning dynamics in aggregate data is to recover and predict the evolution of distribution of all individuals together. \textbf{Trajectory data}, in contrast, is a kind of data that we are able to acquire the information of each individual all the time. Although some studies also considered the case that partial trajectories are missing, the identities of those individuals, whenever they are observable, are always assumed available. For example, stock price, weather, customer behaviors and most training data sets for computer vision and natural language processing are considered as trajectory data. There are many existing models to learn dynamics of full-trajectory data. Typical ones include Hamiltonian neural networks \cite{hnn}, Hidden Markov Model (HMM) \cite{hmm1, eddy1996hidden}, Kalman Filter (KF) \cite{kf1, harvey1990forecasting, originalkf} and Particle Filter (PF) \cite{pf1, djuric2003particle}, as well as the models built upon HMM, KF and PF \cite{hmmvar, pf2, hefny2015supervised, langford2009learning}. They require full trajectories of each individual, which may not be applicable in the aggregate data situations. On the other side, only a few methods are proposed on aggregated data in the recent learning literature. In the work of \citet{hashinn}, authors assumed that the hidden dynamic of particles follows a stochastic differential equation (SDE), in particular, they used a recurrent neural network to parameterize the drift term. Furthermore, \citet{yisen2018aggregate} improved traditional HMM model by using an SDE to describe the evolving process of hidden states and \citet{hmmaggregate} updated HMM parameters through aggregate observations. 

We propose to learn the dynamics of  density through the weak form of Fokker Planck Equation (FPE), which is a parabolic partial differential equation (PDE) governing many dynamical systems subject to random noise perturbations, including the typical SDE models in existing studies. Our learning is accomplished by minimizing the Wasserstein distance between predicted distribution given by FPE and the empirical distribution from data samples. Meanwhile we utilize neural networks to handle higher dimensional cases. More importantly, by leveraging the framework of Wasserstein Generative Adversarial Network (WGAN) \cite{wgan}, our model is capable of approximating the distribution of samples at different time points without solving the SDE or FPE. More specifically, we treat the drift coefficient, the goal of learning, in the FPE as a generator, and the test function in the weak form of FPE as a discriminator. In other words, our method can also be regarded as a data-driven method to estimate transport coefficient in FPE, which corresponds to the drift terms in SDEs. Additionally, though we treat diffusion term as a constant in our model, it is straightforward to generalize it to be a neural network as well, which can be an extension of this work. We would like to mention that several methods of solving SDE and FPE \cite{deepforsde, deepforsde1, parafpe} adopt opposite ways to our method, they utilize neural networks to estimate the distribution $P(x,t)$ with given drift and diffusion terms. 

In conclusion, our contributions are: 1) We develop an algorithm that learns the drift term of a SDE via minimizing the Wasserstein discrepancy between the observed aggregate data and our generated data.
2) By leveraging a {\it weak} form of FPE, we are able to compute the Wasserstein distance directly without solving the FPE.
3) Finally, we demonstrate the accuracy and the effectiveness of our algorithm via several synthetic and real-world examples.

\section{Proposed Method}
\subsection{Fokker Planck Equation for the density evolution}
\begin{figure}[ht]
\begin{center}
\centerline{\includegraphics[width=0.75\columnwidth]{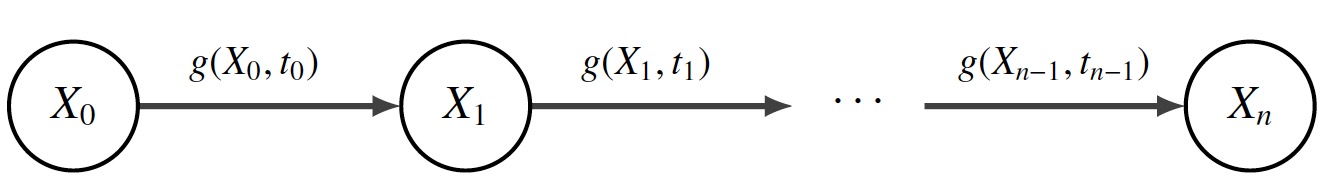}}
\caption{State model of the stochastic process $X_t$}
\label{fig:modelintro}
\end{center}
 \vskip -0.2in
\end{figure}

We assume the individuals evolve in a pattern in the space $\mathbb{R}^D$ as shown in Figure \ref{fig:modelintro}. One example satisfying such process is the  stochastic differential equation(SDE), which is also known as the It\^{o} process \cite{oksendal2003stochastic}: $d\bm{X}_t = g(\bm{X}_t,t)dt+ \sigma d\bm{W}_t$.
Here $d\bm{X}_t$ represents an infinitesimal change of $\{\bm{X}_t\}$ along with time increment $dt$, $g(\cdot,t) = (g^1(\cdot, t), ..., g^D(\cdot, t))^T$ is the drift term (drifting vector field) that drives the dynamics of SDE, $\sigma$ is the diffusion constant, $\{\bm{W}_t\}$ is the standard Brownian Motion. 

\vspace{-0.5em}

The probability density of $\{\bm{X}_t\}$ is governed by the Fokker Planck Equation(FPE) \cite{fokkerplanckbook}: 
\begin{lemma}
Suppose $\{\boldsymbol{X}_t\}$ solves the SDE $d\bm{X}_t = g(\bm{X}_t,t)dt+ \sigma d\bm{W}_t$, denote $p(\cdot, t)$ as the probability density of the random variable $\bm{X}_t$. Then $p(x,t)$ solves the following equation:
\begin{align}
&\frac{\partial p(\bm{x},t)}{\partial t} =\nonumber\\ &\sum_{i=1}^D-\frac{\partial}{\partial{x_i}}\biggl[g^i(\bm{x}, t)p(\bm{x},t)\biggr]+\frac{1}{2}\sigma^2\sum_{i = 1}^{D}\frac{\partial^2}{\partial{x_i}^2}p(\bm{x},t).
\label{eqn:fpe}
\end{align}
\label{lem:fpe}
\end{lemma}
\vspace{-0.6em}
As a linear evolution PDE, FPE describes the evolution of density function of the stochastic process driven by a SDE. Due to this reason, FPE plays a crucial role in stochastic calculus, statistical physics and modeling \citep{quantum1, lowdim, thefpe}. Its importance is also drawing more attention among statistic and machine learning communities \citep{variationalgd, datadrivensb, varinf}. In this paper, we utilize the weak form of FPE as a basis to study hidden dynamics of the time evolving aggregated data without solving FPE. 

Our task can be described as: assume that the individuals evolve with the process indicated by Figure \ref{fig:modelintro}, which can be simulated by It\^{o} process. Then given observations $\bm{x}_t$ along time axis, we aim to recover the drift coefficient $g(\bm{x},t)$ in FPE, and thus we are able to recover and predict the density evolution of such dynamic. For simplicity we treat $g(\bm{x},t)$ as a function uncorrelated to time $t$, namely, $g(\bm{x},t) = g(\bm{x})$. Notice that though evolving process of individuals can be simulated by It\^{o} process, in reality since we lose identity information of individuals, the observed data become aggregate data, thus we need a new way other than traditional methods to study the swarm's distribution.

We also remark that in the work of \citet{hashinn}, based on Jordan-Kinderlehrer-Otto (JKO) \citep{jko} theorem, they utilize RNN to approximate potential function and measure Sinkhorn distance (an approximation to Wasserstein distance). In our work, we assume that the density follows Fokker Planck equation, but we don't solve it directly. Instead, we take the weak form of Fokker Planck equation and compute everything in sample form, which coincides with a similar form of WGAN at the observations. Particularly, we treat FPE as the dynamic regularizer for the marginal fitting problem, therefore is fundamentally different from previous methods. As a byproduct, our numerical scheme allows to freely choose the time step $\Delta t$, which is not restricted to the given time stamp of observations. $\Delta t$ is used to control the error bound.

\subsection{Weak Form of Fokker Planck Equation}
Given FPE stated in Lemma \ref{lem:fpe}, if we multiply a test function $f \in H_0^1(\mathbb{R}^D)$ on both sides of the FPE, where $H_0^1(\mathbb{R}^D)$ denote the Sobolev space. Integration on both sides:
\vspace{0.61em}
\begin{align}
    \int\frac{\partial p}{\partial t}f(\bm{x})d\bm{x} = &\int \sum_{i=1}^D-\frac{\partial}{\partial{x_i}}\left[g^i(\bm{x})p(\bm{x},t)\right]f(\bm{x})d\bm{x} \nonumber\\
    &+\frac{1}{2}\sigma^2\int\sum_{i=1}^{D}\frac{\partial^2}{\partial{x_i}^2}p(\bm{x},t)f(\bm{x})d\bm{x}.\nonumber
\end{align}
\vspace{-0.6em}
\begin{align}
    \int \frac{\partial p}{\partial t}f(\bm{x})d\bm{x} = &\int \sum_{i=1}^D g^i(\bm{x}) \frac{\partial}{\partial{x_i}}f(\bm{x}) p(\bm{x},t)d\bm{x} \nonumber\\
    &+\frac{1}{2}\sigma^2\int\sum_{i = 1}^{D}\frac{\partial^2}{\partial{x_i}^2} f(\bm{x}) p(\bm{x},t) d\bm{x}.\nonumber
\end{align}

The first advantage of weak solution is that the solution of a PDE usually requires strong regularity and thus may not exist in the classical sense for a certain group of equations, however, the weak solution has fewer regularity requirements and thus their existence are guaranteed for a much larger classes of equations. The second advantage is that the weak formulation may provide new perspectives for numerically solving PDEs \cite{zienkiewicz1971finite, sirignano2018dgm, wan}.

Suppose the observed samples at time points $t_{m-1}$ and $t_{m}$ follow the true densities $\thickhat{p}(\cdot,t_{m-1})$ and $\thickhat{p}(\cdot,t_{m})$ respectively. Let's consider the following SDE:
\begin{align}
  &d\thicktilde{\bm{X}_t} = g_\omega(\thicktilde{\bm{X}_t})dt + \sigma d\bm{W}_t,\nonumber\\
  &\text{where}\quad t_{m-1}\leq t \leq t_m, \quad \thicktilde{\bm{X}}_{t_{m-1}} \sim \thickhat{p}(\cdot, t_{m-1}).  \label{SDE a step}
\end{align}
Here $g_\omega$ is an approximation to the real drift term $g$. In our research, we treat $g_\omega$ as a neural network with parameters $\omega$. Stochastic process $\thicktilde{\bm{X}_t}$ has a density function, denoted by $\thicktilde{p}(\cdot,t)$, which is different from the observed density. Hence, it is natural to compute and minimize the discrepancy between the approximated density $\thicktilde{p}(\cdot,t_m)$ and true density $\thickhat{p}(\cdot,t_m)$, within which we optimize $g_\omega$ and thus recover the true drift term $g$.  

In our research, we choose the Wasserstein-1 distance as our discrepancy function \citep{villani2008optimal} \cite{wgan}. Applying Kantorovich-Rubinstein duality \cite{villani2008optimal} leads to $W_1(\thickhat{p}(\cdot,t_m),\thicktilde{p}(\cdot, t_m)) =$
\[\sup_{\norm{\nabla f}\leq 1}\biggl\{\mathbb{E}_{\bm{x}_r\sim \thickhat{p}(\bm{x},t_{m})}[f(\bm{x}_r)] - \mathbb{E}_{\bm{x}_g\sim \thicktilde{p}(\bm{x},t_{m})}[f(\bm{x}_g)]\biggr\}.
\label{eqn:wgan1}
\]

The first term $\mathbb{E}_{\bm{x}_r\sim \thickhat{p}(\bm{x},t_{m})}[f(\bm{x}_r)]$ can be conveniently computed by Monte-Carlo method since we are already provided with the real data points $\bm{x}_r\sim\thickhat{p}(\cdot, t_m)$. To evaluate the second term, we first approximate $\thicktilde{p}(\cdot,t_m)$ by trapezoidal rule \cite{atkinson2008introduction}: $\thicktilde{p}(\bm{x},t_m) \approx$
\begin{align}
\thickhat{p}(\bm{x},t_{m-1})+\frac{\Delta t}{2}\left(\frac{\partial \thickhat{p}(\bm{x},t_{m-1})}{\partial t}+\frac{\partial \thicktilde{p}(\bm{x},t_m)}{\partial t}\right),
\label{eqn:pevolve}
\end{align}
where $\Delta t = t_m-t_{m-1}$. Then we compute:
\begin{align}
    &\mathbb{E}_{\bm{x}_g\sim \thicktilde{p}(\cdot,t_{m})}[f(\bm{x}_g)] \approx \int  f(\bm{x})\thickhat{p}(\bm{x},t_{m-1})d\bm{x}+ \nonumber\\
    &\frac{\Delta t}{2}\biggl( \int \frac{\partial \thickhat{p}(\bm{x},t_{m-1})}{\partial t} f(\bm{x}) d\bm{x} + \int \frac{\partial \thicktilde{p}(\bm{x},t_m)}{\partial t} f(\bm{x})  d\bm{x} \biggr). \label{eq1}
\end{align}
In the above Equation (\ref{eq1}), the second and the third term on the right-hand side can be reformulated via the weak form of FPE. This gives us a new formulation for $W_1(\thickhat{p}(\cdot,t_m),\thicktilde{p}(\cdot, t_m))$, which can by computed by using Monte-Carlo method. In fact, the first and the second terms in \eqref{eq1} can be directly computed via data points from $\thickhat{p}(\cdot, t_{m-1})$. For the third term, we need to generate samples from $\thicktilde{p}(\cdot, t_{m})$. To achieve this, we apply Euler-Maruyama scheme \cite{kloeden2013numerical} to SDE (\ref{SDE a step}) in order to acquire our desired samples $\tilde{\bm{x}}_{t_m}$: 
\begin{align}
    &\tilde{\bm{x}}_{t_m} = \hat{\bm{x}}_{t_{m-1}} + g_\omega(\hat{\bm{x}}_{t_{m-1}})\Delta t + \sigma\sqrt{\Delta t}\bm{z},\nonumber\\
    &\text{where} \quad \bm{z}\sim \mathcal{N}(0, I), \quad \hat{\bm{x}}_{t_{m-1}}\sim \thickhat{p}(\cdot, t_{m-1}).
    \label{eqn:nsde}
\end{align}
Here $\mathcal{N}(0, I)$ is the standard Gaussian distribution on $\mathbb{R}^D$. Now we summarize these results in Proposition 1:

\begin{prop}
For a set of points $X=\{\bm{x}^{(1)},...,\bm{x}^{(N)}\}$ in $\mathbb{R}^D$.  We denote $\mathcal{F}_f(X)$ as:
\begin{align}
\frac{1}{N}\sum\limits_{k=1}^N\Biggl(\sum_{i=1}^Dg_\omega^{i}(\bm{x}^{(k)})\frac{\partial}{\partial x_i}f(\bm{x}^{(k)})+\frac{1}{2}\sigma^2\sum_{i=1}^D\frac{\partial^2}{\partial x_i^2 }f(\bm{x}^{(k)})\Biggr),\nonumber
\end{align}
then at time point $t_{m}$, the Wasserstein distance between $\thickhat{p}(\cdot, t_m)$ and $\thicktilde{p}(\cdot, t_{m})$ can be approximated by:
\begin{align}
   W_1(&\thickhat{p}(\cdot, t_m),  \thicktilde{p}(\cdot, t_m)) \approx \sup_{\norm{\nabla{f}}\leq 1}\biggl\{\frac{1}{N}\sum_{k=1}^N f(\thickhat{\bm{x}}_{t_{m}}^{(k)}) \nonumber\\
   &- \frac{1}{N}\sum\limits_{k=1}^N f(\thickhat{\bm{x}}_{t_{m-1}}^{(k)}) - \frac{\Delta t}{2}\biggl(\mathcal{F}_f(\thickhat{X}_{m-1}) + \mathcal{F}_f(\thicktilde{X}_m)\biggr)\biggr\}.\nonumber
\end{align}

Here $\{\thickhat{\bm{x}}_{t_{m-1}}^{(k)}\}\sim \thickhat{p}(\cdot, t_{m-1})$, $\{\thickhat{\bm{x}}_{t_{m}}^{(k)}\}\sim \thickhat{p}(\cdot, t_{m})$. We denote $\thickhat{X}_{m-1}=\{\thickhat{\bm{x}}_{t_{m-1}}^{(1)},...,\thickhat{\bm{x}}_{t_{m-1}}^{(N)}\}$, $\thicktilde{X}_m = \{\thicktilde{\bm{x}}_{t_m}^{(1)},...,\thicktilde{\bm{x}}_{t_m}^{(N)}\}$, where each $\thicktilde{\bm{x}}_{t_m}^{(k)}$ is computed by Euler-Maruyama scheme.
\label{prop:loss1}
\end{prop}

\subsection{Wasserstein Distance on Time Series}
In real cases, it is not realistic to observe the data at arbitrary two consecutive time nodes, especially when $\Delta t$ is small. To make our model more flexible, we extend our formulation so that we are able to plug in observed data at arbitrary time points. To be more precise, suppose we observe data set $\thickhat{X}_{t_n}=\{ \thickhat{\bm{x}}_{t_n}^{(1)}, ..., \thickhat{\bm{x}}_{t_n}^{(N)} \}$ at $J+1$ different time points $t_0, t_1,..., t_J$. And we denote the generated data set as $\tilde{X}_{t_n} = \{\thicktilde{\bm{x}}_{t_n}^{(1)},...,\thicktilde{\bm{x}}_{t_n}^{(N)}\}$, here each $\tilde{\bm{x}}_{t_n}^{(\cdot)}$ is derived from the $n$-step Euler-Maruyama scheme: 
\begin{align}
&\tilde{\bm{x}}_{t_j} = \tilde{\bm{x}}_{t_{j-1}} + g_\omega(\tilde{\bm{x}}_{t_{j-1}})\Delta t + \sigma \sqrt{\Delta t} \bm{z},\nonumber\\
&\textrm{where} \quad \bm{z}\sim\mathcal{N}(0,I), \quad 0\leq j\leq n, \quad \tilde{\bm{x}}_{t_0}\sim \thickhat{p}(\cdot, t_0). 
\label{n-step Euler Maruyama}
\end{align}

Let us denote $\tilde{p}(\cdot, t)$ as the solution to FPE (\ref{eqn:fpe}) with $g$ replaced by $g_\omega$ and with initial condition $\tilde{p}(\cdot, t_0) = \thickhat{p}(\cdot, t_0)$, then the approximation formula for evaluating the Wasserstein distance $W_1(\hat{p}(\cdot, t_n), \tilde{p}(\cdot, t_n))$ is provided in the following proposition:
\begin{prop}
Suppose we keep all the notations defined as above, then we have the approximation:
\begin{align}
&W_1(\thickhat{p}(\cdot, t_n), \thicktilde{p}(\cdot, t_n)) \approx \sup_{\norm{\nabla{f}}\leq 1} \biggl\{\frac{1}{N}\sum_{k=1}^N f(\thickhat{\bm{x}}_{t_n}^{(k)}) \nonumber\\
&- \frac{1}{N}\sum\limits_{k=1}^N f(\thickhat{\bm{x}}_{t_0}^{(k)}) - \frac{\Delta t}{2}\biggl(\mathcal{F}_f(\thickhat{X}_{0}) + \mathcal{F}_f(\thicktilde{X}_{n})\nonumber\\
&+ 2\sum_{s=1}^{n-1}\mathcal{F}_f(\thicktilde{X}_{s})\biggl)\biggr\}.\nonumber
\end{align}
\label{prop:loss2}
\end{prop}
\vspace{-1em}
\textbf{Minimizing the Objective Function:} Base on Proposition \ref{prop:loss2}, we obtain objective function by summing up the accumulated Wasserstein distances among $J$ observations along the time axis. Thus, our goal is to minimize the following objection function:
\begin{align}
   \min_{g_\omega}
   &\Biggl\{
   \sum_{n=1}^J \sup_{\norm{\nabla{f_n}}\leq 1}
   \biggl\{\frac{1}{N}\sum_{k=1}^N f_n(\thickhat{\bm{x}}_{t_n}^{(k)}) - \frac{1}{N}\sum\limits_{k=1}^N f_n(\thickhat{\bm{x}}_{t_0}^{(k)}) \nonumber\\
   &- \frac{\Delta t}{2}\biggl(\mathcal{F}_{f_n}(\thickhat{X}_{0}) + \mathcal{F}_{f_n}(\thicktilde{X}_{n}) + 2\sum_{s=1}^{n-1}\mathcal{F}_{f_n}(\thicktilde{X}_{s})\biggl)
   \biggr\}\Biggr\}.\nonumber
\end{align}
Notice that since we have observations on $J$ distinct time points, for each time point we compute Wasserstein distance with the help of the dual function $f_n$, thus we involve $J$ test functions in total. In our actual implementation, we will choose these dual functions as neural networks. We call our algorithm Fokker Planck Process(FPP), the entire procedure is shown in Algorithm \ref{alg:1}. We also provide an error analysis in Appendix.
\begin{remark}
When the time interval $\Delta t = t_j - t_i$ between two observations at $X_i$ and $X_j(i<j)$ is large. In order to guarantee the accuracy of $\tilde{X}_s$, we can separate $\Delta t$ into multiple smaller intervals, namely, $\Delta t = K h$, where $K$ the number of intervals and $h$ is the interval length. Then we evaluate \eqref{n-step Euler Maruyama} on the finer meshes to obtain more accurate samples $\{\tilde{x}_s^{(1)},...,\tilde{x}_s^{(N)}\}$ at specific time $s$.
\end{remark}
\begin{remark}
The drift function recovered by our framework may not be unique, see Section 4 for more details.
\end{remark}

\begin{algorithm}[tb]
\caption{Fokker Planck Process Algorithm}
\label{alg:1}
\begin{algorithmic}[1]
\REQUIRE{Initialize $f_{\theta_n}$ ($1\leq n \leq J$), $g_\omega$ }
\REQUIRE{Set $\epsilon_{f_n}$ as the inner loop learning rate for $f_{\theta_n}$ and $\epsilon_g$ as the outer loop learning rate for $g_\omega$}
\FOR{$\#$ training iterations}
\FOR{k steps}
\FOR{observed time $t_s$ in $\{ t_1,...,t_J\}$}
\STATE Compute the generated data set $\tilde{X}_{t_s}$ from Euler-Maruyama scheme (\ref{n-step Euler Maruyama}) for $1\leq s \leq J$
\STATE Acquire data sets $\thickhat{X}_{t_s}=\{\thickhat{\bm{x}}_{t_s}^{(1)},...,\thickhat{\bm{x}}_{t_s}^{(N)}\}$ from real distribution $\thickhat{p}(\cdot, t_s)$ for $1 \leq s \leq J$
\ENDFOR
\STATE For each dual function $f_{\theta_n}$, compute:
    $\mathcal{F}_n=\mathcal{F}_{f_{\theta_n}}(\thickhat{X}_{t_0})+\mathcal{F}_{f_{\theta_n}}(\tilde{X}_{t_n}) + 2 \sum_{s=1}^{n-1}\mathcal{F}_{f_{\theta_n}}(\tilde{X}_{t_s})$
\STATE Update each $f_{\theta_n}$ by:\\
$\theta_n\leftarrow\theta_n+\epsilon_{f_n}\nabla_{\theta}\biggl(\frac{1}{N}\sum_{k=1}^N f_{\theta_n}(\thickhat{\bm{x}}_{t_n}^{(k)})-\frac{1}{N}\sum_{k=1}^N f_{\theta_n}(\thickhat{\bm{x}}_{t_0}^{(k)})-\frac{\Delta t}{2}\mathcal{F}_n\biggr)$
\ENDFOR
\STATE Update $g_{\omega}$ by:\\
$\omega \leftarrow \omega-\epsilon_g \nabla_\omega \biggl(\sum_{n=1}^J\big(\frac{1}{N}f_{\theta_n}(\thickhat{\bm{x}}_{t_n}^{(k)})-
\frac{1}{N}f_{\theta_n}(\thickhat{\bm{x}}_{t_0}^{(k)})$\\
$-\frac{\Delta t}{2}\mathcal{F}_n\big)\biggr)$  \label{gradient wrt g }
\ENDFOR
\end{algorithmic}
\end{algorithm}

\vspace{-1em}
\section{Experiments}
In this section, we evaluate our model on various synthetic and realistic data sets by employing Algorithm \ref{alg:1}. We generate samples $\thicktilde{\bm{x}}_t$ and make all predictions base on Equation (\ref{eqn:nsde}) starting with $\thickhat{\bm{x}}_0$.

\textbf{Baselines:}
We compare our model with two recently proposed methods. One model (NN) adopts recurrent neural network(RNN) to learn dynamics directly from observations of aggregate data \cite{hashinn}. The other one model (LEGEND) learns dynamics in a HMM framework \cite{yisen2018aggregate}. The baselines in our experiments are two typical representatives that have state-of-the-art performance on learning aggregate data. Furthermore, though we simulate the evolving process of the data as a SDE, which is on the same track with NN, as mentioned before, NN trains its RNN via optimizing Sinkhorn distance \cite{sinkhorn}, our model starts with a view of weak form of PDE, focuses more on WGAN framework and easier computation.

\subsection{Synthetic Data}
We first evaluate our model on three synthetic data sets which are generated
by three artificial dynamics: \textbf{Synthetic-1}, \textbf{Synthetic-2} and \textbf{Synthetic-3}.

\textbf{Experiment Setup:}
In all synthetic data experiments, we set the drift term $g$ and the discriminator $f$ as two simple fully-connected networks. The $g$ network has one hidden layer and the $f$ network has three hidden layers. Each layer has 32 nodes for both $g$ and $f$. The only one activation function we choose is Tanh. Notice that since we need to calculate $\frac{\partial^2f}{\partial x^2}$, the activation function of $f$ must be twice differentiable to avoid loss of weight gradient. In terms of training process, we use the Adam optimizer \cite{kingma2014adam} with learning rate $10^{-4}$. Furthermore, we use spectral normalization to realize $\lVert \nabla{f} \rVert \leq 1$\cite{spectral}. We initialize the weights with Xavier initialization\cite{xvaier} and train our model by Algorithm 1. We set the data size at each time point is $N=2000$, treat $1200$ data points as the training set and the other $800$ data points as the test set, $\Delta t$ is set to be 0.01.

\begin{figure}[t!]
    \centering
     \subfloat[][Syn-1: at 10$\Delta t$]{\includegraphics[width=.33\linewidth]{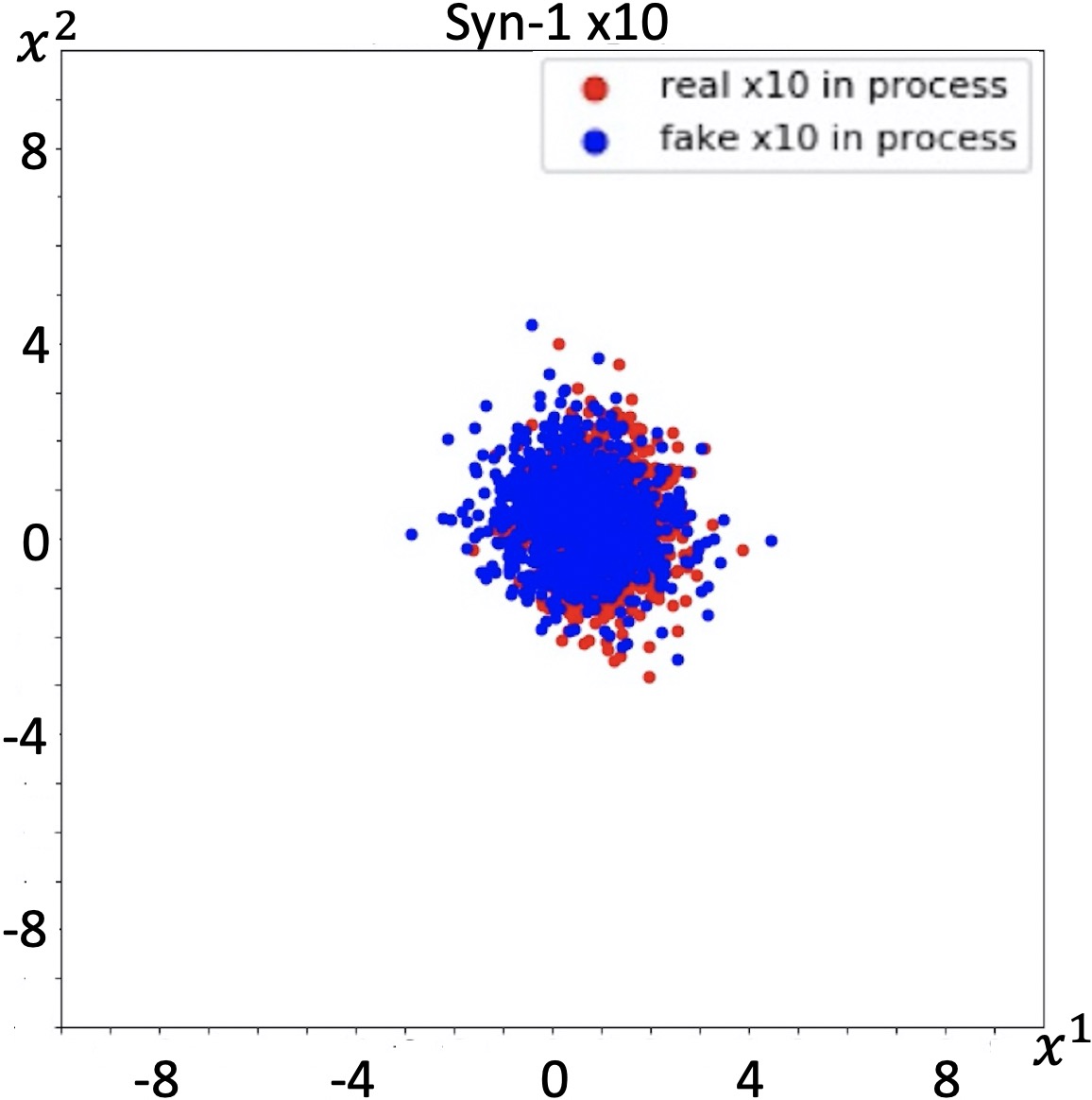}}
     \subfloat[][Syn-1: at 50$\Delta t$]{\includegraphics[width=.33\linewidth]{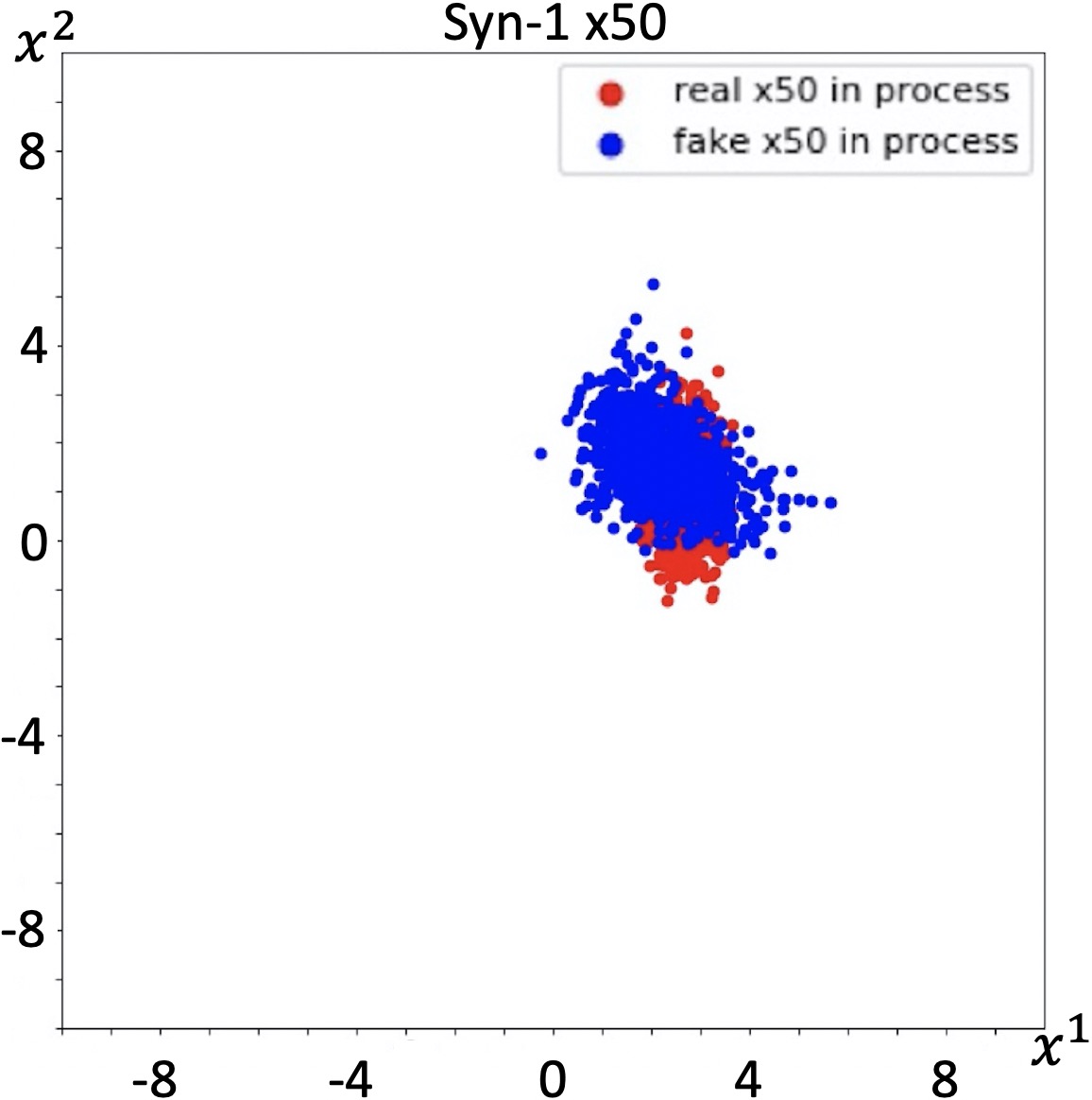}}
     \subfloat[][Syn-1: at 500$\Delta t$]{\includegraphics[width=.33\linewidth]{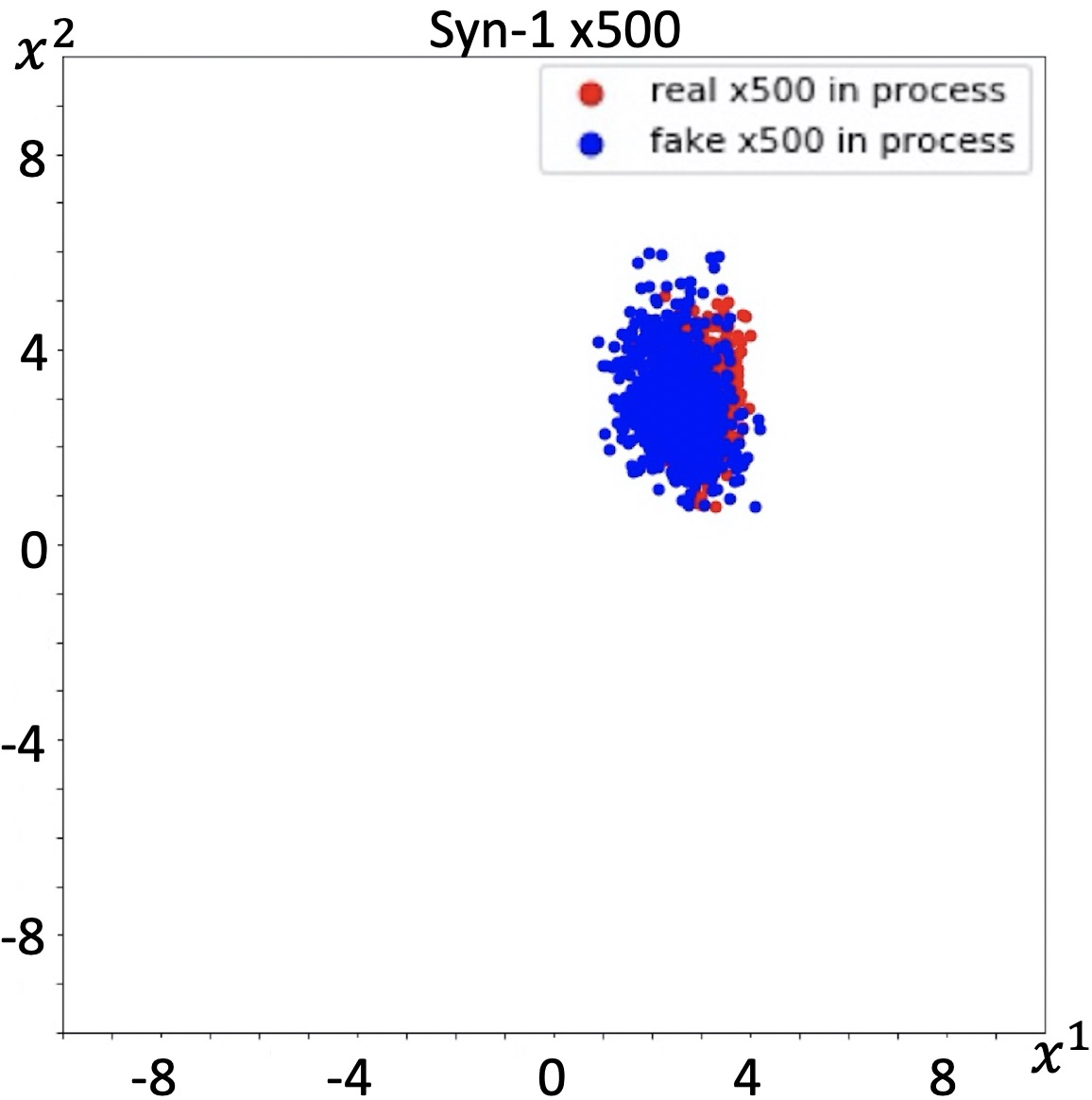}}\\
     \subfloat[][Syn-2: at 30$\Delta t$]{\includegraphics[width=.33\linewidth]{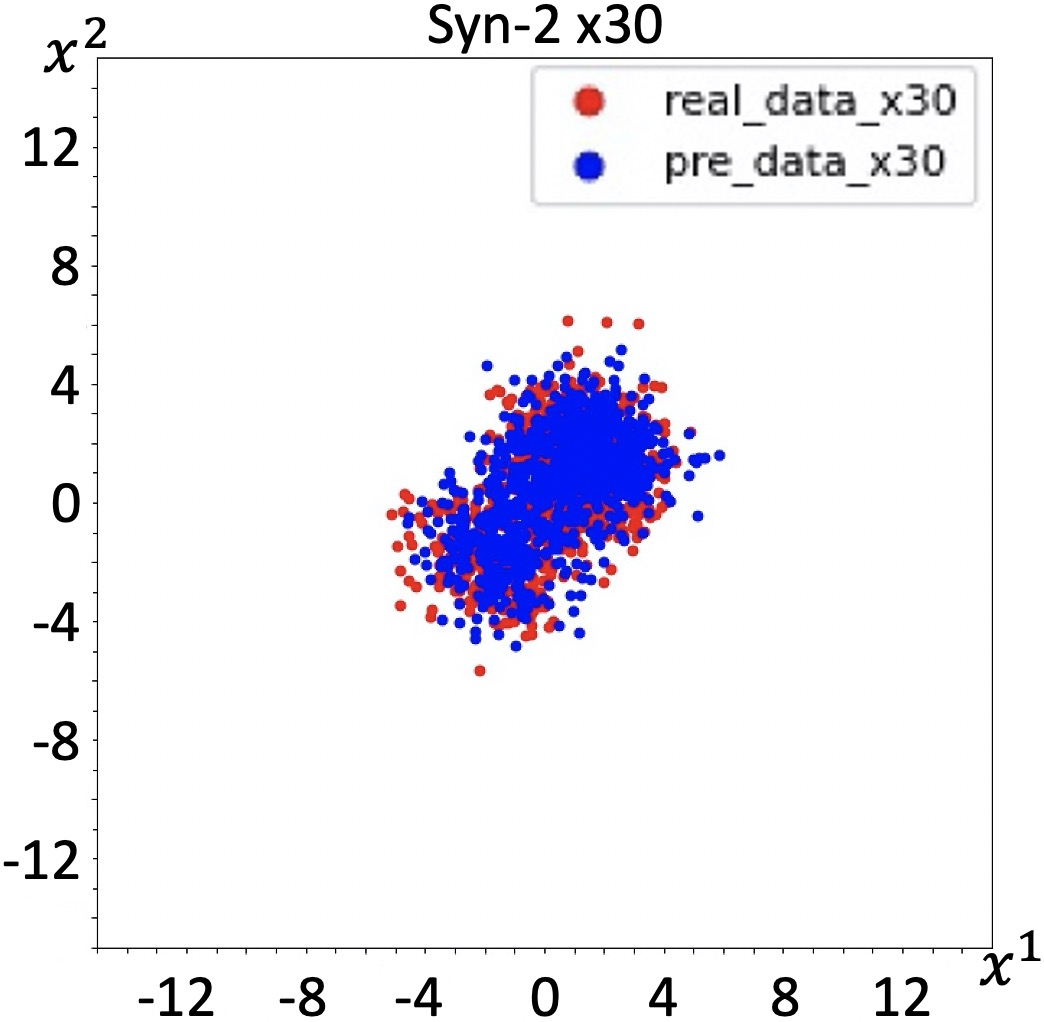}}
     \subfloat[][Syn-2: at 50$\Delta t$]{\includegraphics[width=.33\linewidth]{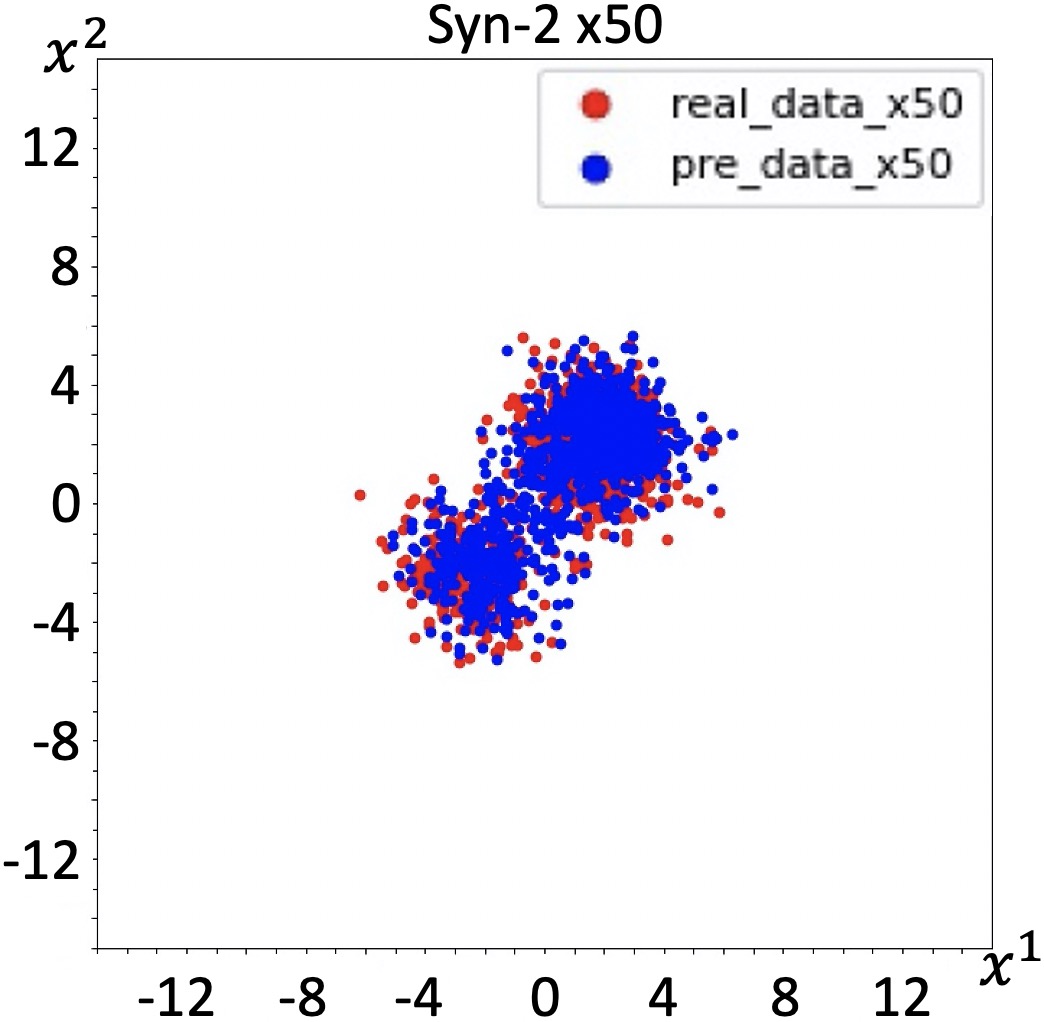}}
     \subfloat[][Syn-2: at 100$\Delta t$]{\includegraphics[width=.33\linewidth]{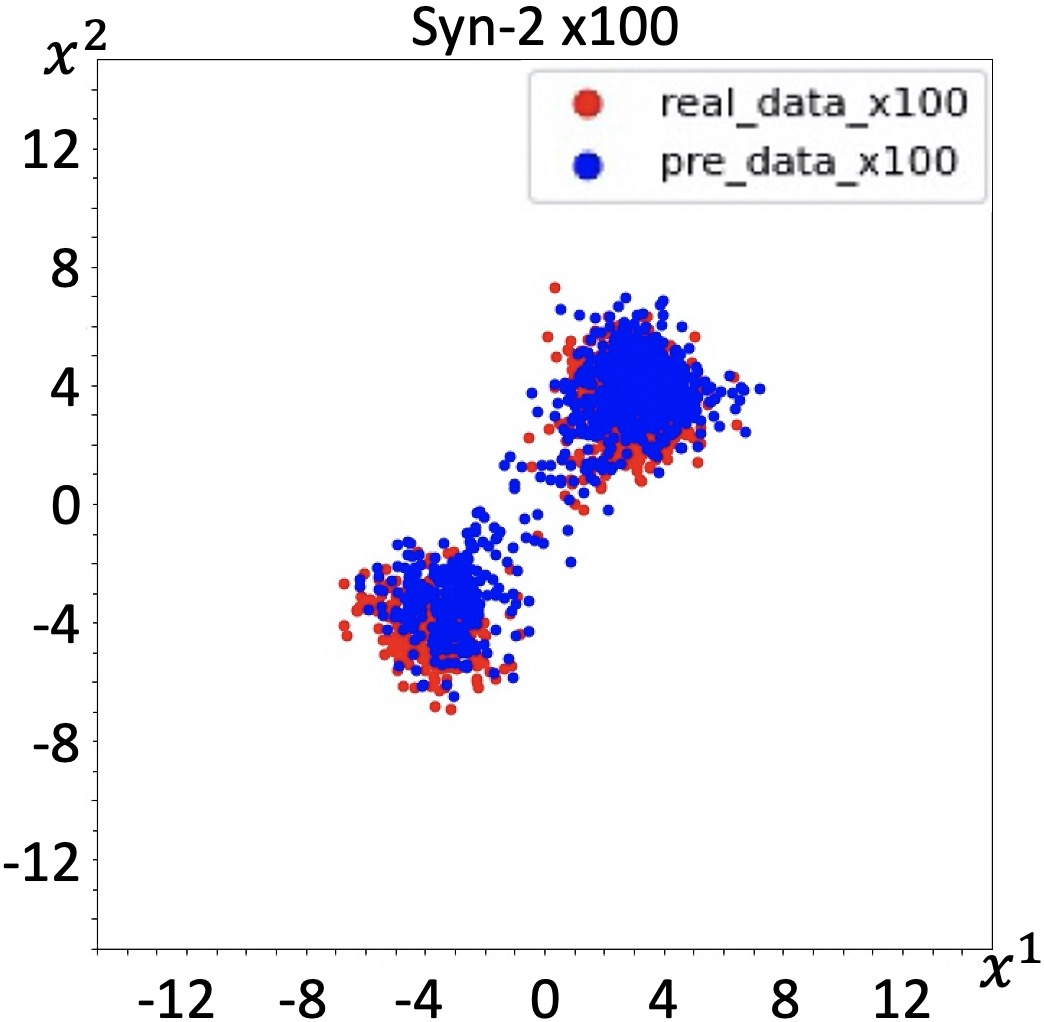}}\\
     \subfloat[][Syn-3: at 10$\Delta t$]{\includegraphics[width=.33\linewidth]{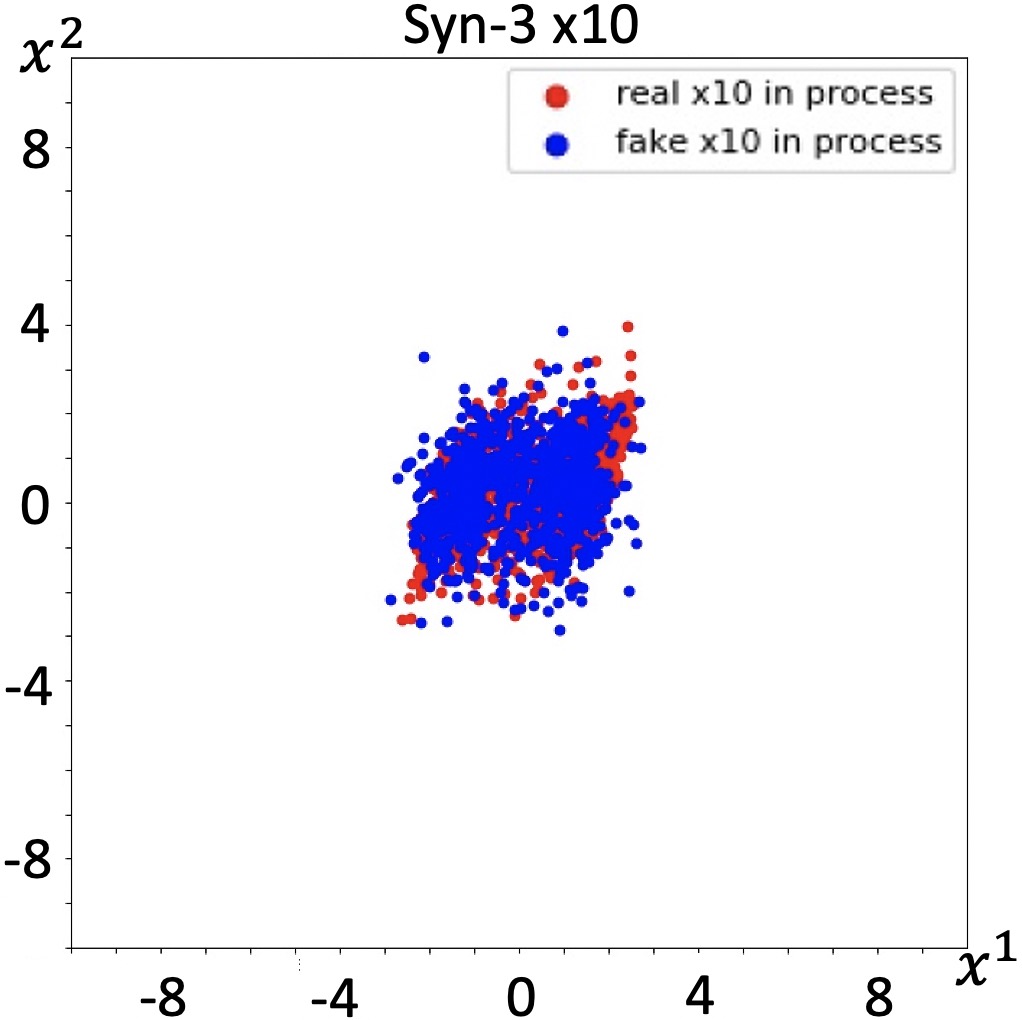}}
     \subfloat[][Syn-3: at 30$\Delta t$]{\includegraphics[width=.33\linewidth]{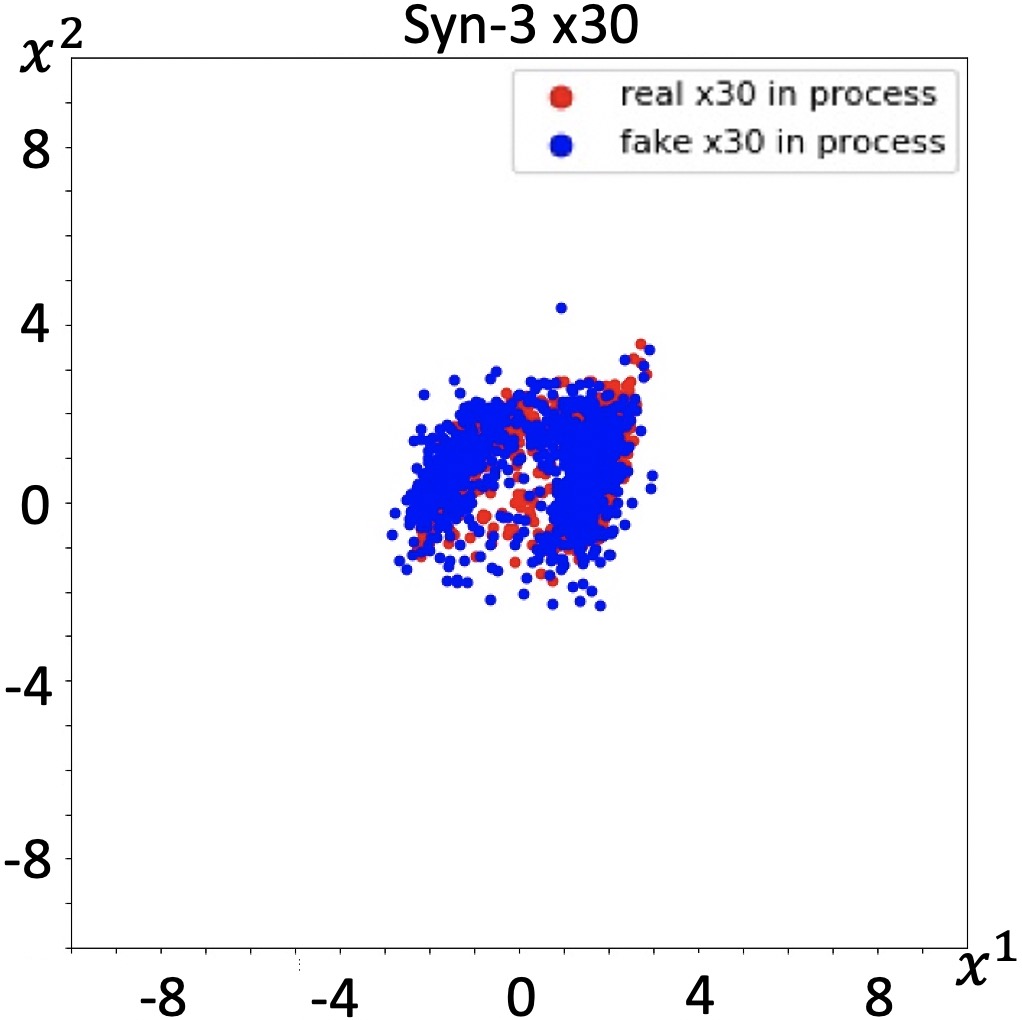}}
     \subfloat[][Syn-3: at 50$\Delta t$]{\includegraphics[width=.33\linewidth]{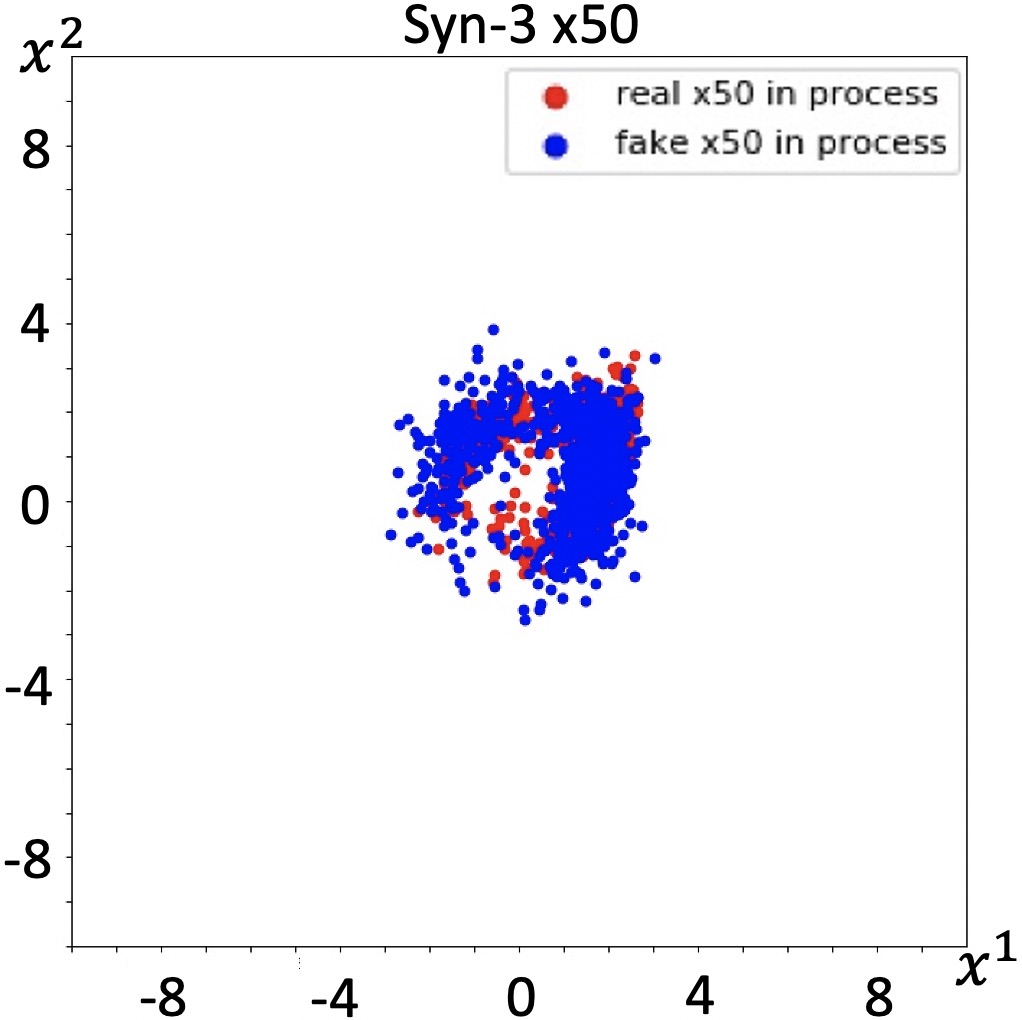}}
\caption{Comparison of generated data(blue) and ground truth(red) of Synthetic-1((a) to (c)), Synthetic-2((d) to (f)) and Synthetic-3((g) to (i)). In each case, it finally converges to a stationary distribution.}
\label{fig:syn}
\end{figure}

\textbf{Synthetic-1:}
\begin{align}
    &\thickhat{\bm{x}}_0 \sim \mathcal{N}(0,\bm{\Sigma}_0),\nonumber\\ &\thickhat{\bm{x}}_{t+\Delta t} = \thickhat{\bm{x}}_{t} - (\bm{A}\thickhat{\bm{x}}_{t} + \bm{b})\Delta t + \sigma \sqrt{\Delta t}\mathcal{N}(0, 1).\nonumber
\end{align}
In Synthetic-1, the data is following a simple linear dynamic, we set $\bm{A} = [(4,0),(0,1)], \bm{b}= [-12,\ -12]^T$, $\sigma = 1$, $\bm{\Sigma}_0 = \bm{I_2}$. We utilize true $\bm{x}_0$, $\bm{x}_{20}$ and $\bm{x}_{200}$ in training process and predict the distributions of $\bm{x}_{10}$, $\bm{x}_{50}$ and $\bm{x}_{500}$. As visualized in Figure \ref{fig:syn}, from $(a)$ to $(c)$, the generated data(blue) covers all areas of ground truth(red), the original Gaussian distribution converges to the target Gaussian distribution as we expect.

\textbf{Synthetic-2:}
\begin{align}
    &\thickhat{\bm{x}}_0 \sim \mathcal{N}(0,\bm{\Sigma}_0), \quad \thickhat{\bm{x}}_{t+\Delta t} = \thickhat{\bm{x}}_{t} - \bm{G} \Delta t + \sigma \sqrt{\Delta t}\mathcal{N}(0, 1),\nonumber\\
    &\text{where $\bm{G}$ is given in Appendix.}\nonumber
\end{align}
In Synthetic-2, the data is following a complex nonlinear dynamic. We let $\sigma = \sigma_1 = \sigma_2 = 4$, $\mu_1 = [12, 15]^T$ and $\mu_2 = [-15, -15]^T$(defined in Appendix). We utilize true $\bm{x}_{10}$, $\bm{x}_{40}$ and $\bm{x}_{80}$ in training process and predict $\bm{x}_{30}$, $\bm{x}_{50}$ and $\bm{x}_{100}$. The results are shown in Figure \ref{fig:syn}, from $(d)$ to $(f)$, the generated data(blue) covers all areas of ground truth(red), generated samples split and converge to a mixed Gaussian as the ground truth suggests.

\textbf{Synthetic-3~(Nonlinear Van der Pol oscillator \cite{vanderpol}:)}
\begin{align}
    &\thickhat{\bm{x}}_0 \sim \mathcal{N}(0,\bm{\Sigma}_0),\nonumber\\
    &\thickhat{x}_{t+\Delta t}^1= \thickhat{x}_{t}^1 + 10\left(\thickhat{x}_{t}^2-\frac{1}{3}(\thickhat{x}_{t}^1)^3+\thickhat{x}_{t}^1\right)\Delta t + \sigma \sqrt{\Delta t}\mathcal{N}(0, 1),\nonumber\\
    &\thickhat{x}_{t+\Delta t}^2 = \thickhat{x}_t^2 + 3(1-\thickhat{x}_t^1)\Delta t + \sigma \sqrt{\Delta t}\mathcal{N}(0, 1).\nonumber
\end{align}
In Synthetic-3, we let $\sigma = 1$ and utilize true $\bm{x}_3$, $\bm{x}_{7}$ and $\bm{x}_{20}$ in training process then predict the distributions of $\bm{x}_{10}$, $\bm{x}_{30}$ and $\bm{x}_{50}$. As presented in Figure \ref{fig:syn}, from $(g)$ to $(i)$, the generated data(blue) covers all areas of ground truth(red), the distributions we predict are following the true stochastic oscillator's pattern.

\textbf{Remark 3:}
In Syn-2 and Syn-3, $\thickhat{x}_{t}^i$ represents the $i$-th dimension of $\thickhat{\bm{x}}_t$. \textbf{We further state that} in Syn-1 and Syn-3, the training data is coming from the same $\bm{x}_0$ respectively. In Syn-2 the training data is coming from different $\bm{x}_0$, namely, the training data $\bm{x}_{10}$, $\bm{x}_{40}$ and $\bm{x}_{80}$ are generated from three different sets of $\bm{x}_0$. We also consider cases in higher dimensions: D = 6 and 10.
To be more precise, we couple three 2-D dynamical systems to create the 6-D dynamical system and five 2-D systems to create the 10-D example. 
We compare our model with the two baseline models by using Wasserstein distance as error metric for the  low-dimensional (D = 2) and high-dimensional (D = 6, 10) cases. As reported in Table \ref{tab:PPer}, our model achieves lower Wasserstein error than the two baseline models in all cases. Clearly all the drift functions in the synthetic data sets cause the change of the distributions. In Section 4 we discuss a special case when the drift term does not change the distribution.

\subsection{Realistic Data -- RNA Sequence of Single Cell}
\begin{figure}[t!]
     \centering
     \subfloat[][D2 of Mt1]{\includegraphics[width=.4\linewidth]{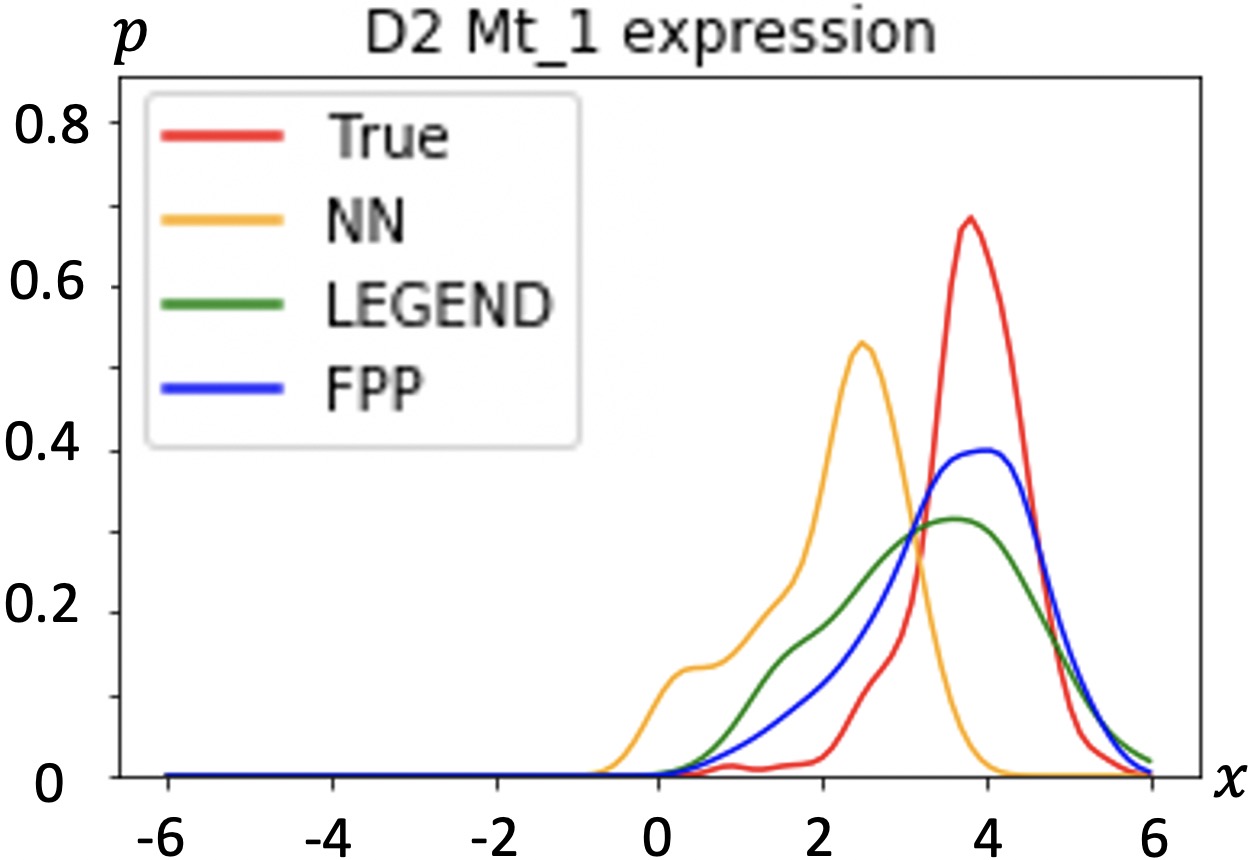}\label{d2m1}}\qquad
     \subfloat[][D7 of Mt1]{\includegraphics[width=.4\linewidth]{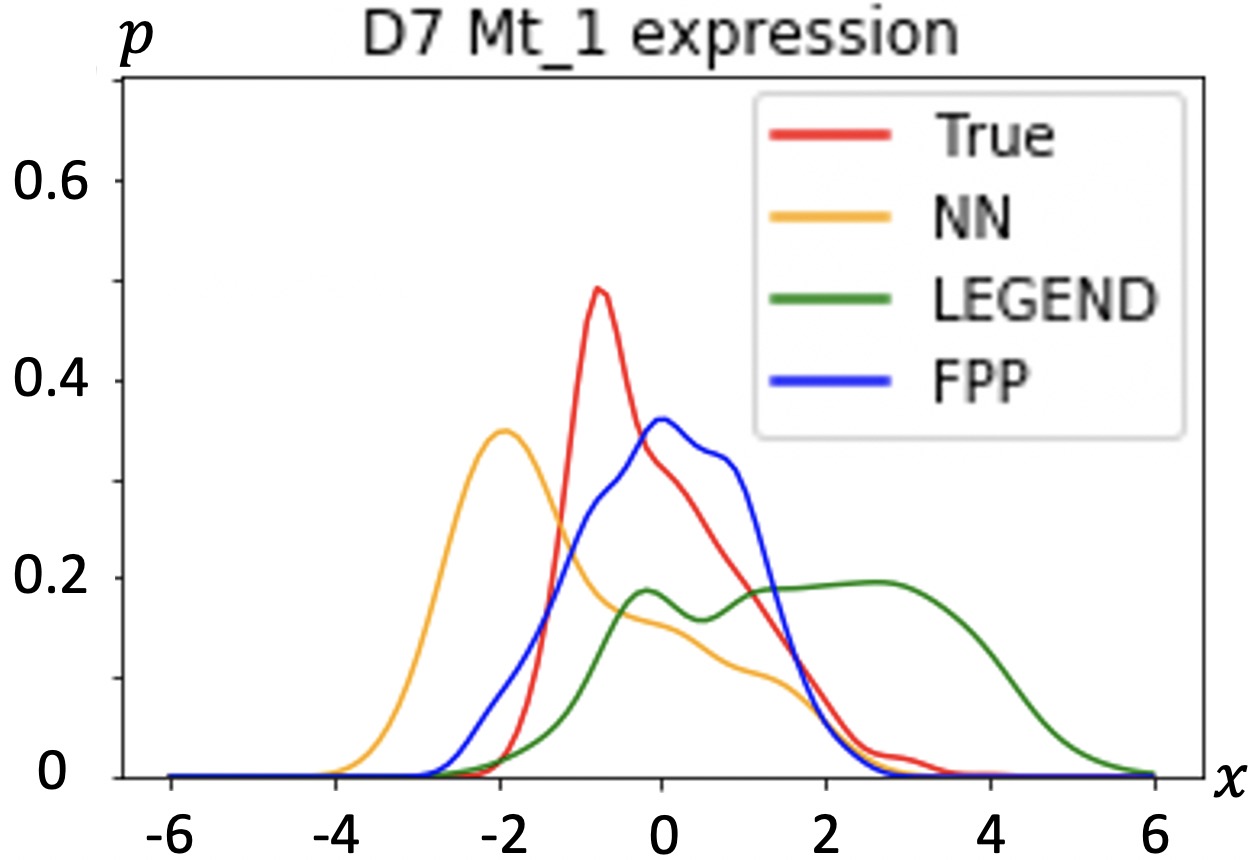}\label{d2m2}}\\
     \subfloat[][D2 of Mt2]{\includegraphics[width=.4\linewidth]{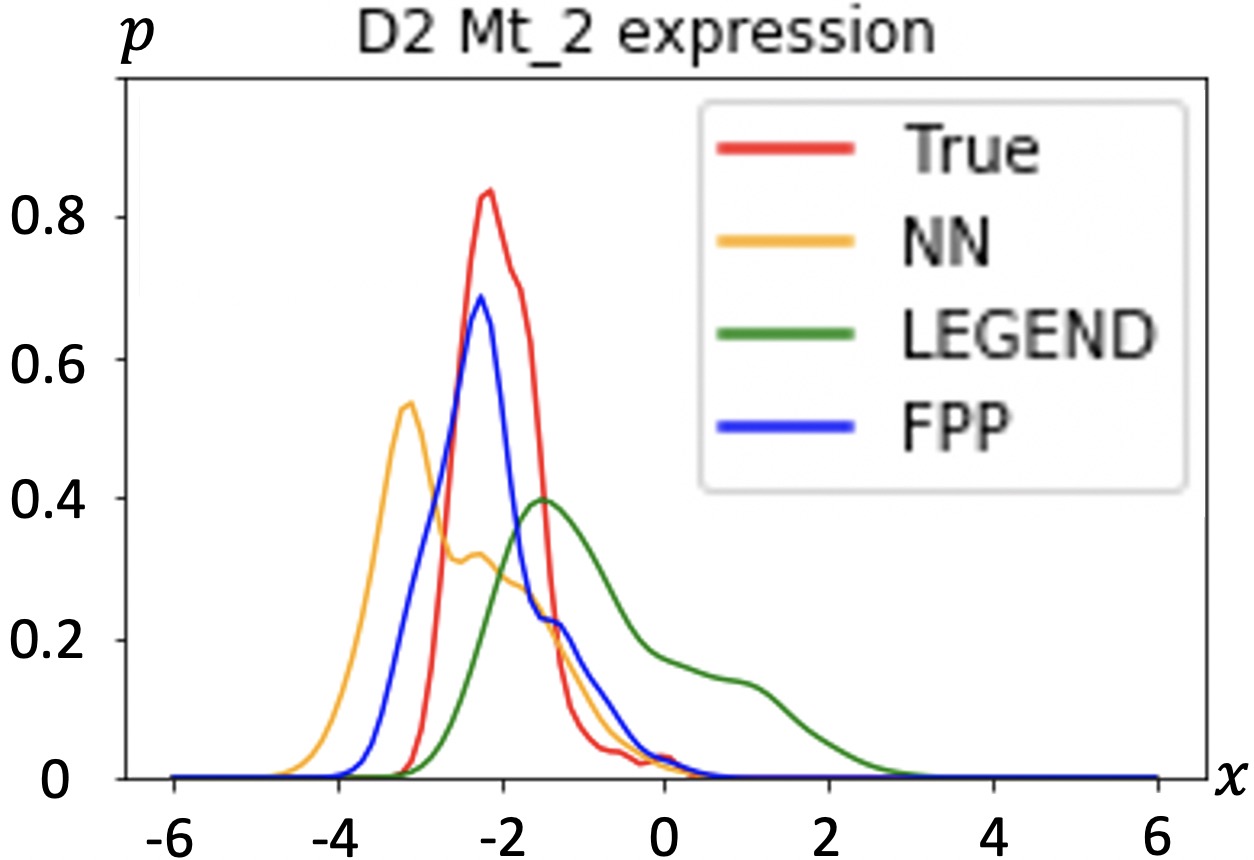}\label{d7m1}}\qquad
     \subfloat[][D7 of Mt2]{\includegraphics[width=.4\linewidth]{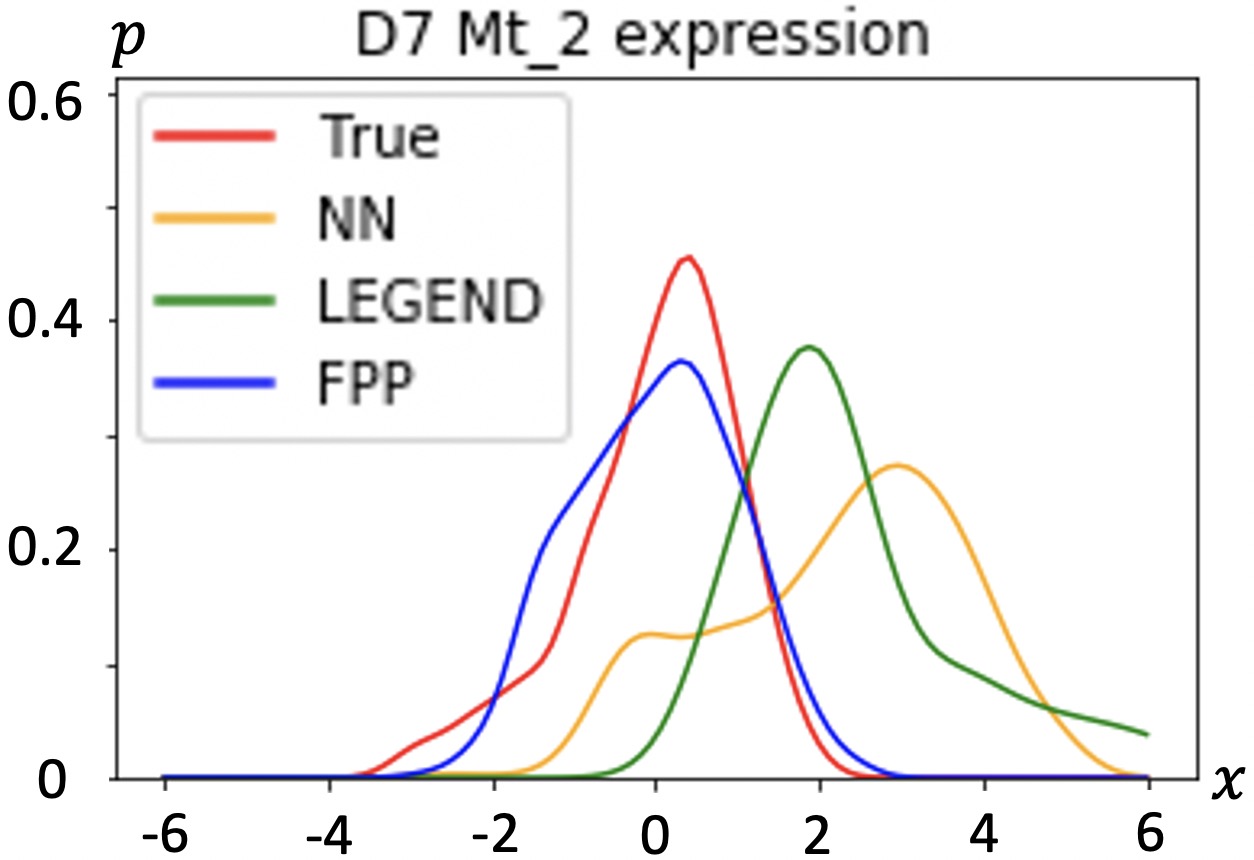}\label{d7m2}}\\
     \subfloat[][Corr on D4]{\includegraphics[width=.4\linewidth]{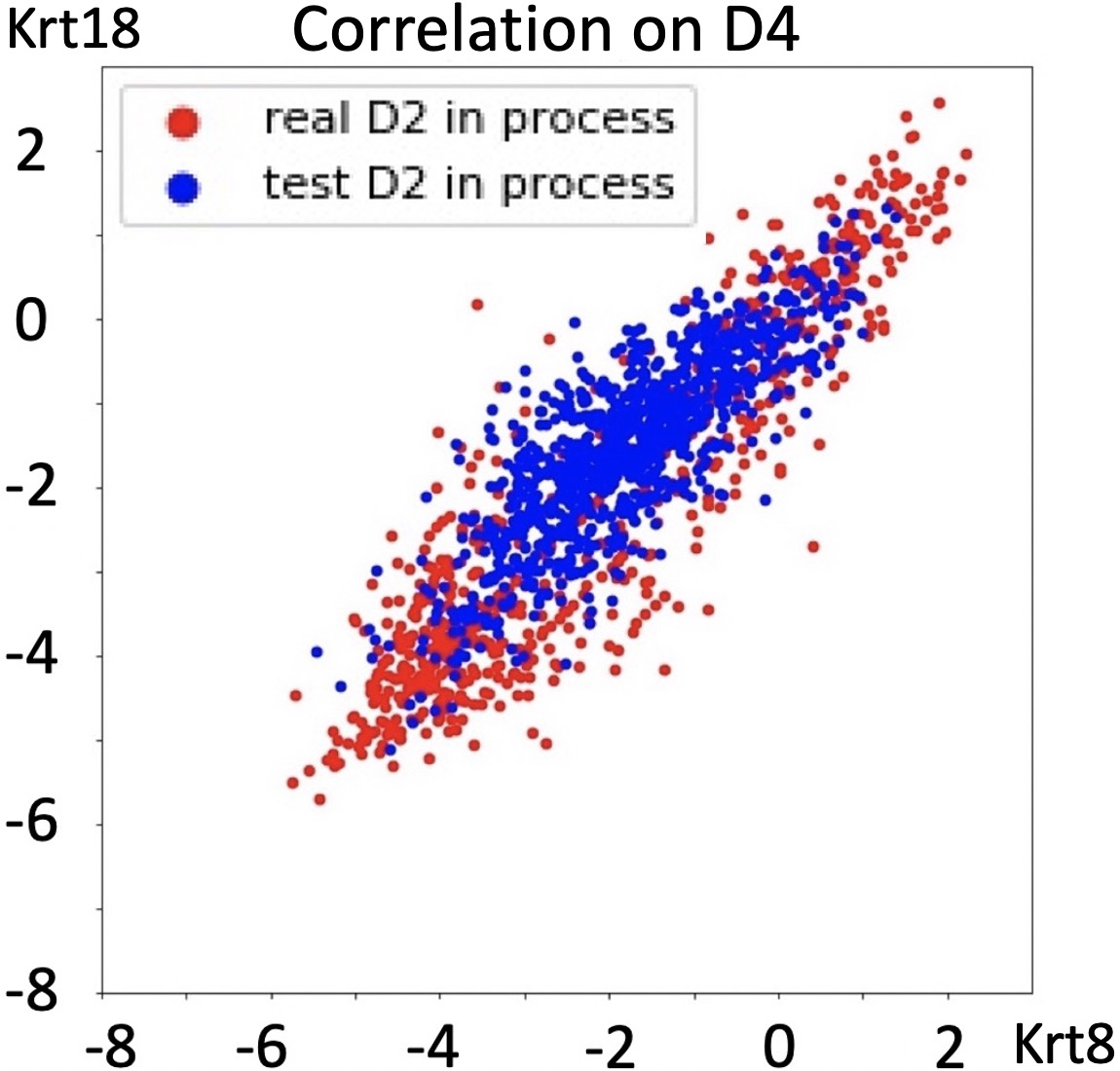}\label{d2corr}}\qquad
     \subfloat[][Corr on D7]{\includegraphics[width=.4\linewidth]{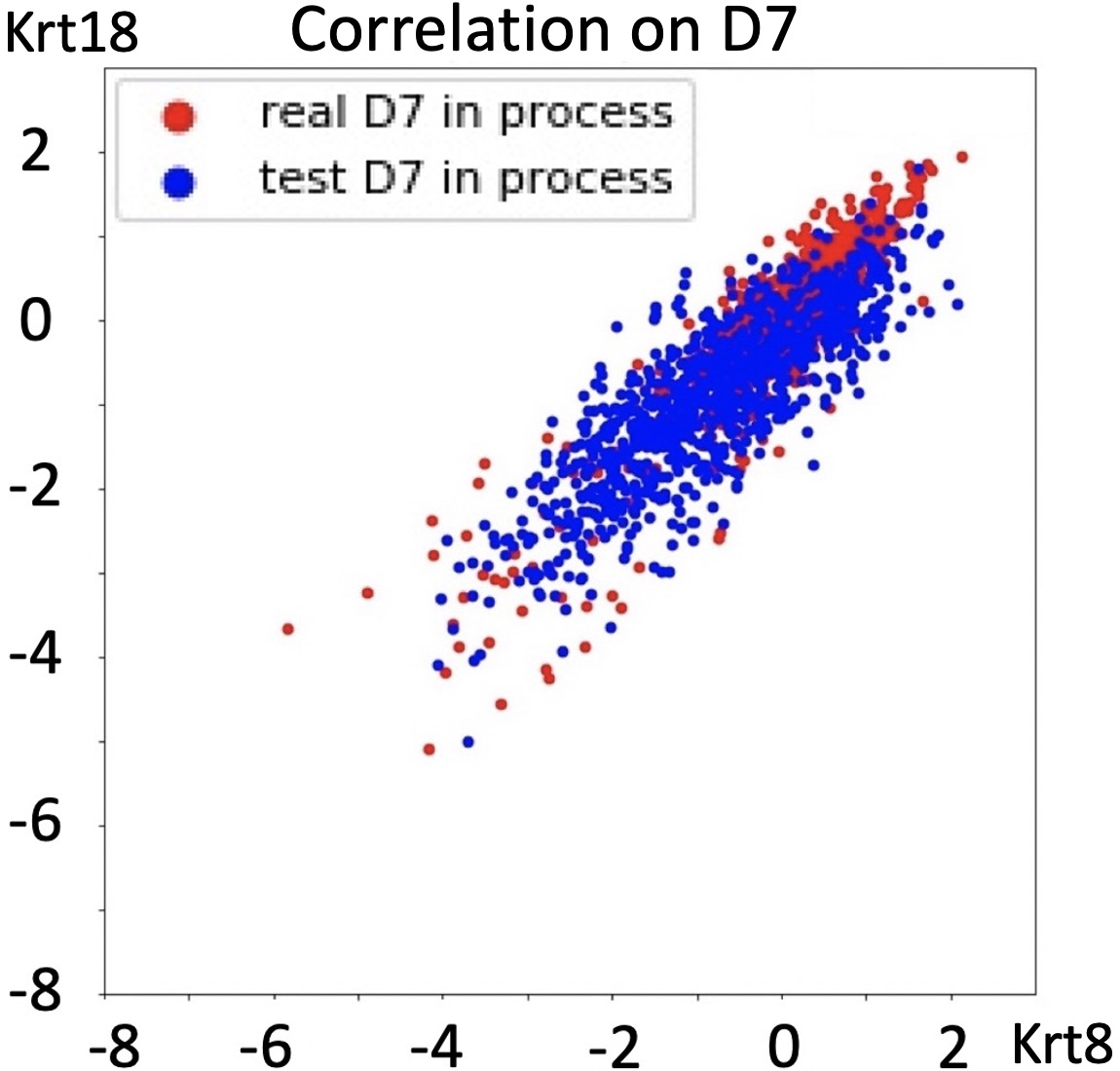}\label{d7corr}}\\
      \subfloat[][W-loss of Mt1, D2]{\includegraphics[width=.4\linewidth]{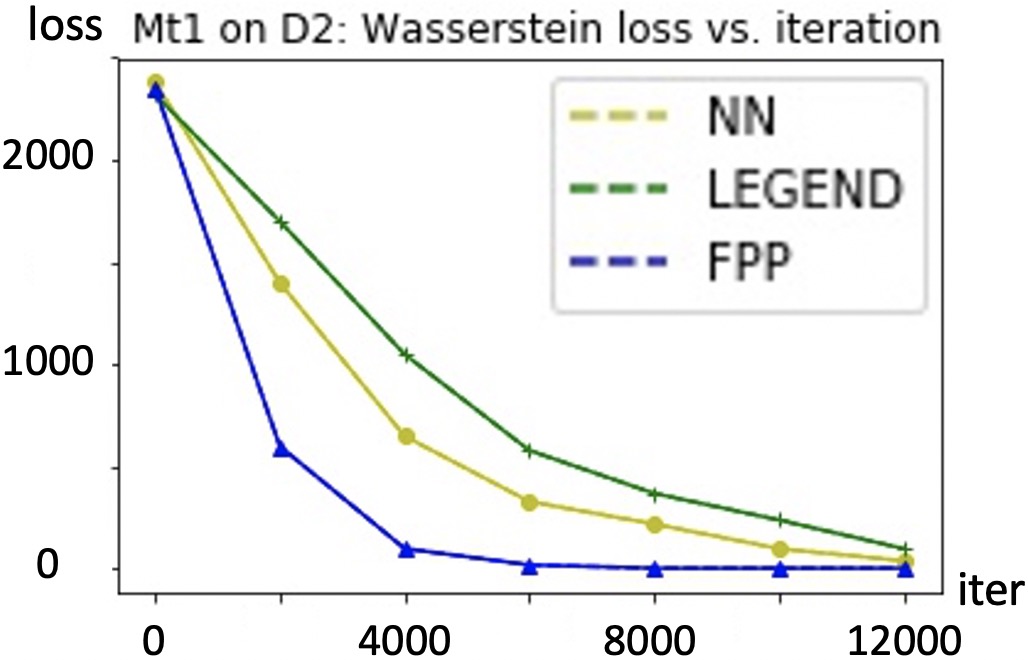}\label{wloss2}}\qquad
     \subfloat[][W-loss of Mt1, D7]{\includegraphics[width=.4\linewidth]{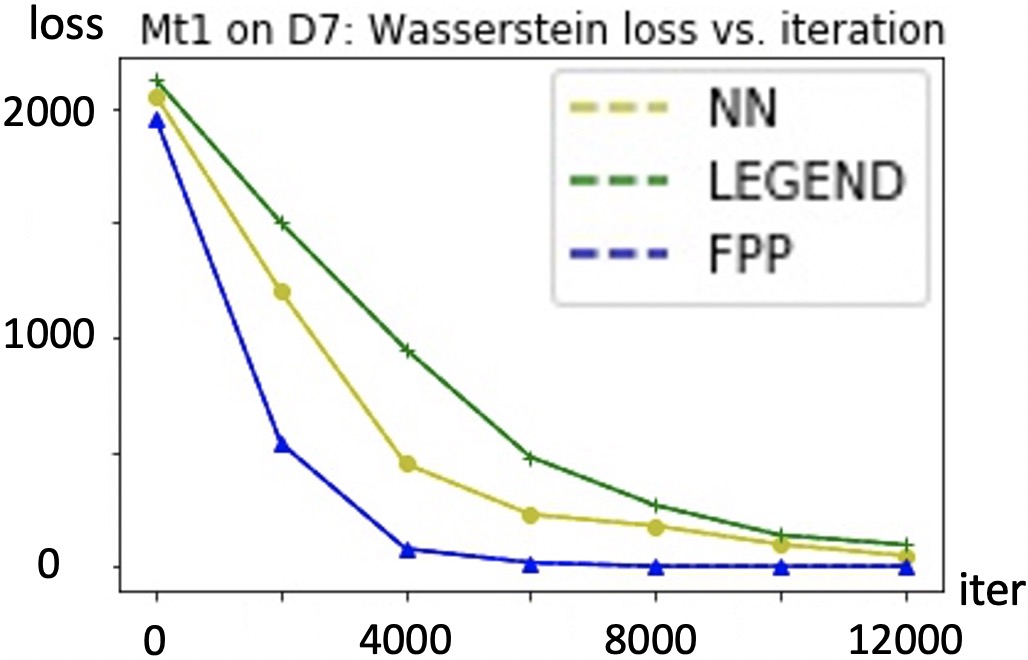}\label{wloss7}}
\caption{(a) to (d): The performance comparisions among different models on D2 and D7 of Mt1 and Mt2. (e) and (f): True (red) and predicted (blue) correlations between Mt1(x-axis) and Mt2(y-axis) on D2 (left) and D7 (right). (g) and (h): Wasserstein loss of Mt1 on D2 and D7 vs iterations.}
\label{fig:gene}
\end{figure}

In this section, we evaluate our model on a realistic biology data set called Single-cell RNA-seq\cite{klein2015droplet}, which is typically used for learning the evolvement of cell differentiation. The cell population begins to differentiate at day 0 (D0). Single-cell RNA-seq observations are then sampled at day 0 (D0), day 2 (D2), day 4 (D4) and day 7 (D7). At each time point, the expression of 24,175 genes of several hundreds cells are measured (933, 303, 683 and 798 cells on D0, D2, D4 and D7 respectively). Notice that there is only whole group's distribution but no trajectory information of each gene on different days. We pick 10 gene markers out of 24,175 to make a 10 dimensional data set. In the first task we treat gene expression at D0, D4 and D7 as training data to learn the hidden dynamic and predict the distribution of gene expression at D2. In the second task we train the model with gene expression at D0, D2 and D4, then predict the distribution of gene expression at D7. We plot the prediction results of two out of ten markers, i.e. Mt1 and Mt2 in Figure \ref{fig:gene}.

\textbf{Experiment Setup:}
We set both $f$ and $g$ as fully connected three-hidden-layers neural networks, each layer has 64 nodes. The only activation function we choose is Tanh. The other setups of neural networks and training process are the same with the ones we use in Synthetic data. Notice that in realistic cases, $\Delta t$ and $T/\Delta t$ become hyperparameters, here we choose $\Delta t = 0.05$, $T/\Delta t = 35$, which means the data evolves 10$\Delta t$ from D0 to D2 , then 10$\Delta t$ from D2 to D4 and finally 15$\Delta t$ from D4 to D7. For preprocessing, we apply standard normalization procedures \cite{Hicks025528} to correct batch effects and use non-negative matrix factorization to impute missing expression levels\cite{hashinn,yisen2018aggregate}.

\begin{table}[t!]
\centering
\caption{The Wasserstein error of different models on
Synthetic-1/2/3 and RNA-sequence data sets.}
\setlength\tabcolsep{4pt}
\begin{tabular}{l|c|c|c c c}
\hline\hline
Data & Task & Dimension & NN & LEGEND & Ours
\\ [0.5ex]
\hline
\multirow{6}{*}{Syn--1} & \multirow{3}{*}{$\bm{x}_{50}$} & 2  &                               1.37 & 0.44 & \textbf{0.05} \\ 
                            &                     & 6 & 4.79 & 2.32 & \textbf{0.06} \\
                            &               & 10 & 9.13 & 2.89 & \textbf{0.10} \\ \cline{2-6}
                            & \multirow{3}{*}{$\bm{x}_{500}$}& 2  & 0.84 & 0.18 &\textbf{0.03} \\
                            &                   & 6   & 3.28 & 0.30 & \textbf{0.03} \\
                            &                  & 10   & 8.05  & 1.79 &\textbf{0.09} \\
\hline
\multirow{6}{*}{Syn--2} & \multirow{3}{*}{$\bm{x}_{50}$} & 2  & 4.72 & 2.84 & \textbf{0.02} \\
                            &                     & 6 & 6.47 & 5.33 & \textbf{0.14} \\
                            &         & 10 & 12.58 & 7.21 & \textbf{0.22} \\ \cline{2-6}
                            & \multirow{3}{*}{$\bm{x}_{100}$}& 2  & 3.83 & 2.98 &\textbf{0.04} \\
                            &                   & 6   & 8.83 & 3.17 &\textbf{0.19}\\
                            &                  & 10   & 14.11  & 5.65 &\textbf{0.32} \\
\hline
\multirow{6}{*}{Syn--3} & \multirow{3}{*}{$\bm{x}_{30}$} & 2  & 4.13 & 1.29 &\textbf{0.08} \\
                            &                     & 6 & 6.40 & 3.16 &\textbf{0.17}\\
                            &            & 10 & 11.76 & 8.53 &\textbf{0.25}\\ \cline{2-6}
                            & \multirow{3}{*}{$\bm{x}_{50}$}& 2  & 3.05 & 0.87 &\textbf{0.12} \\\
                            &                   & 6   & 6.72 & 1.52 & \textbf{0.16} \\
                            &                  & 10   & 9.81  & 3.55 & \textbf{0.23} \\
\hline\hline
\multirow{2}{*}{RNA-Mt1} & D2 & 10 & 33.86 & 10.28 & \textbf{4.23}\\\cline{2-2}
  & D7 & 10 & 12.69 & 7.21 & \textbf{2.92}\\
\hline
\multirow{2}{*}{RNA-Mt2} & D2 & 10 & 31.45 & 13.32 & \textbf{4.04}\\\cline{2-2}
  & D7 & 10 & 11.58 & 7.89 & \textbf{1.50}\\
\hline\hline
\end{tabular}
\label{tab:PPer}
\vspace{-1.5em}
\end{table}

\textbf{Results:}
As shown in Table \ref{tab:PPer}, when compared to other
baselines, our model achieves lower Wasserstein error on both Mt1 and Mt2 data, which proves that our model is capable of learning the hidden dynamics of the two studied gene expressions. In Figure \ref{fig:gene} (a) to (d), we visualized the predicted distributions of the two genes. The distributions of Mt1 and Mt2 predicted by our model (curves in blue) are closer to the true distributions (curves in red) on both D2 and D7. Furthermore, our model precisely indicates the correlations between Krt8 and Krt18 on D4 and D7, as shown in Figure \ref{fig:gene} (e) and (f), which also demonstrates the effectiveness of our model since closer to the true correlation represents better performance (more results in Appendix). In Figure \ref{fig:gene} (g) and (h), we see the training process of our model is easier with least computation time. 

\subsection{Realistic Data – Daily Trading Volume}
\begin{figure}[t!]
     \centering
    \subfloat[][14:35]{\includegraphics[width=.35\linewidth]{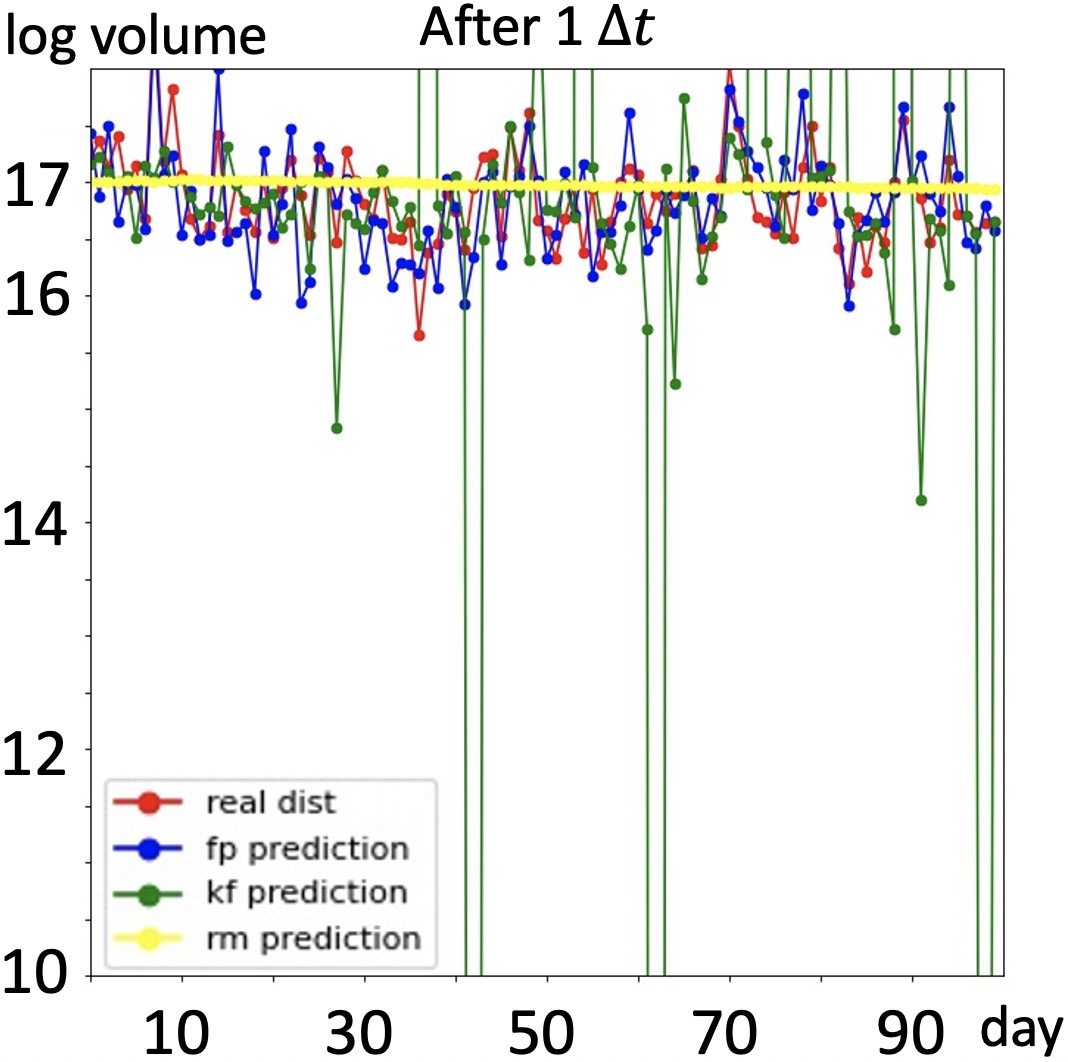}}\qquad
     \subfloat[][15:15]{\includegraphics[width=.35\linewidth]{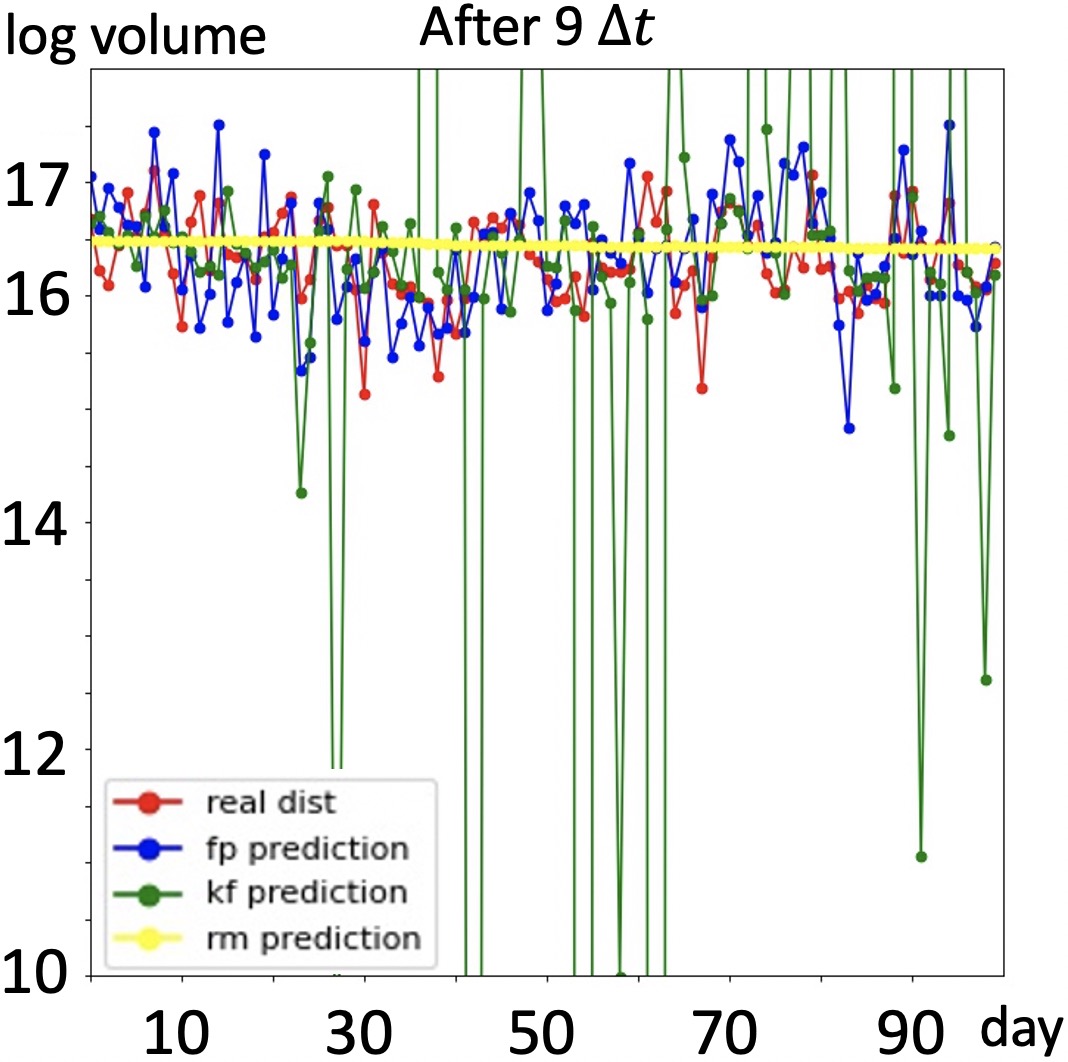}}\\
     \subfloat[][15:35]{\includegraphics[width=.35\linewidth]{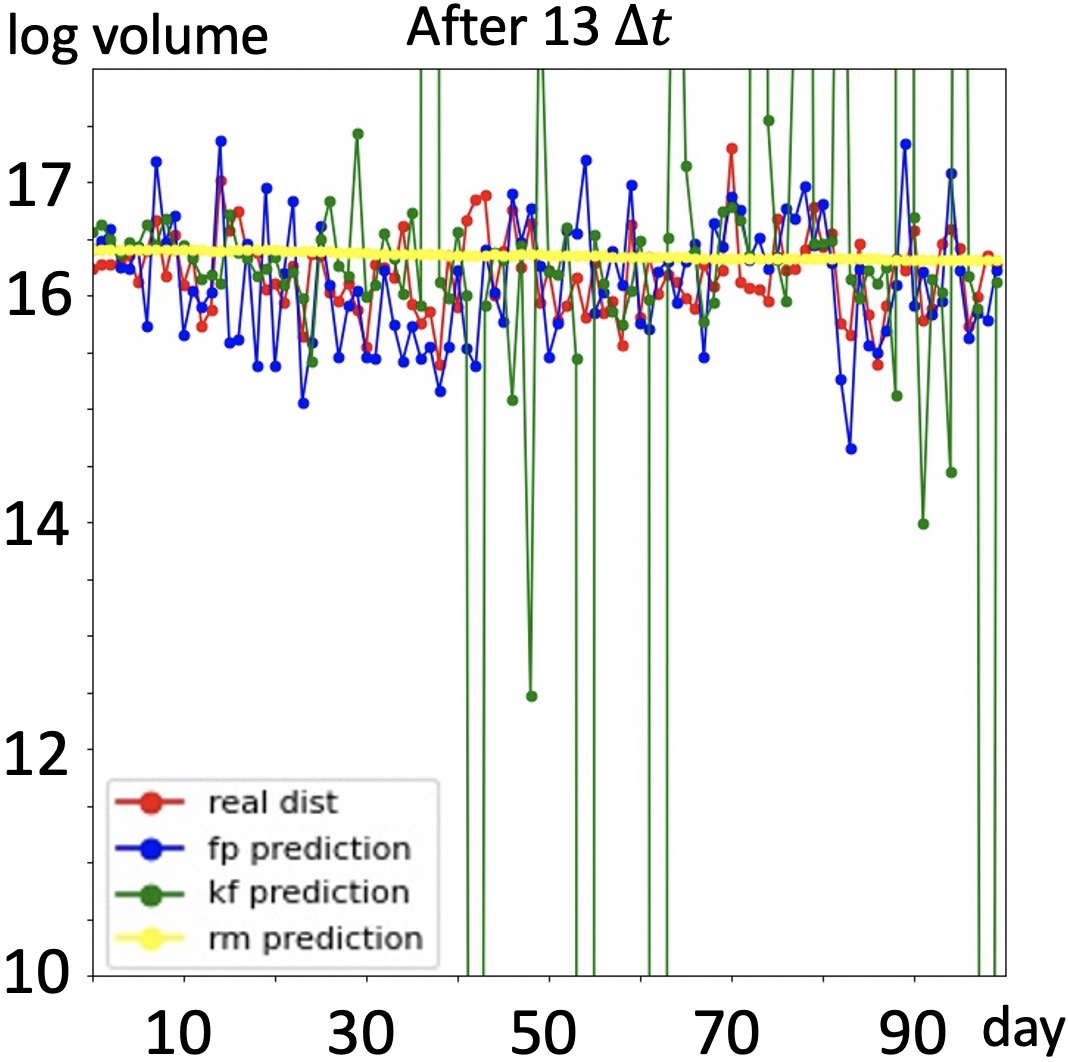}}\qquad
     \subfloat[][16:15]{\includegraphics[width=.35\linewidth]{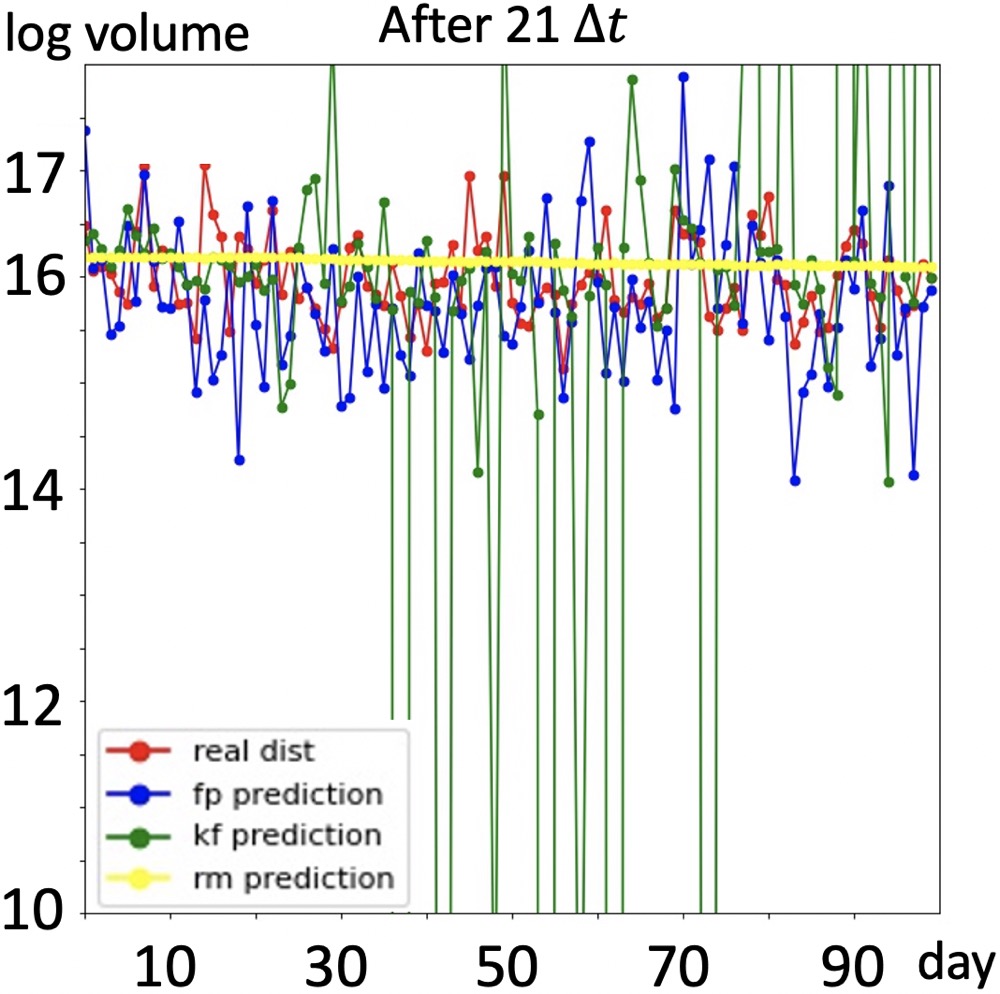}}\\
     \vspace{-1em}
     \subfloat[][14:35]{\includegraphics[width=.35\linewidth]{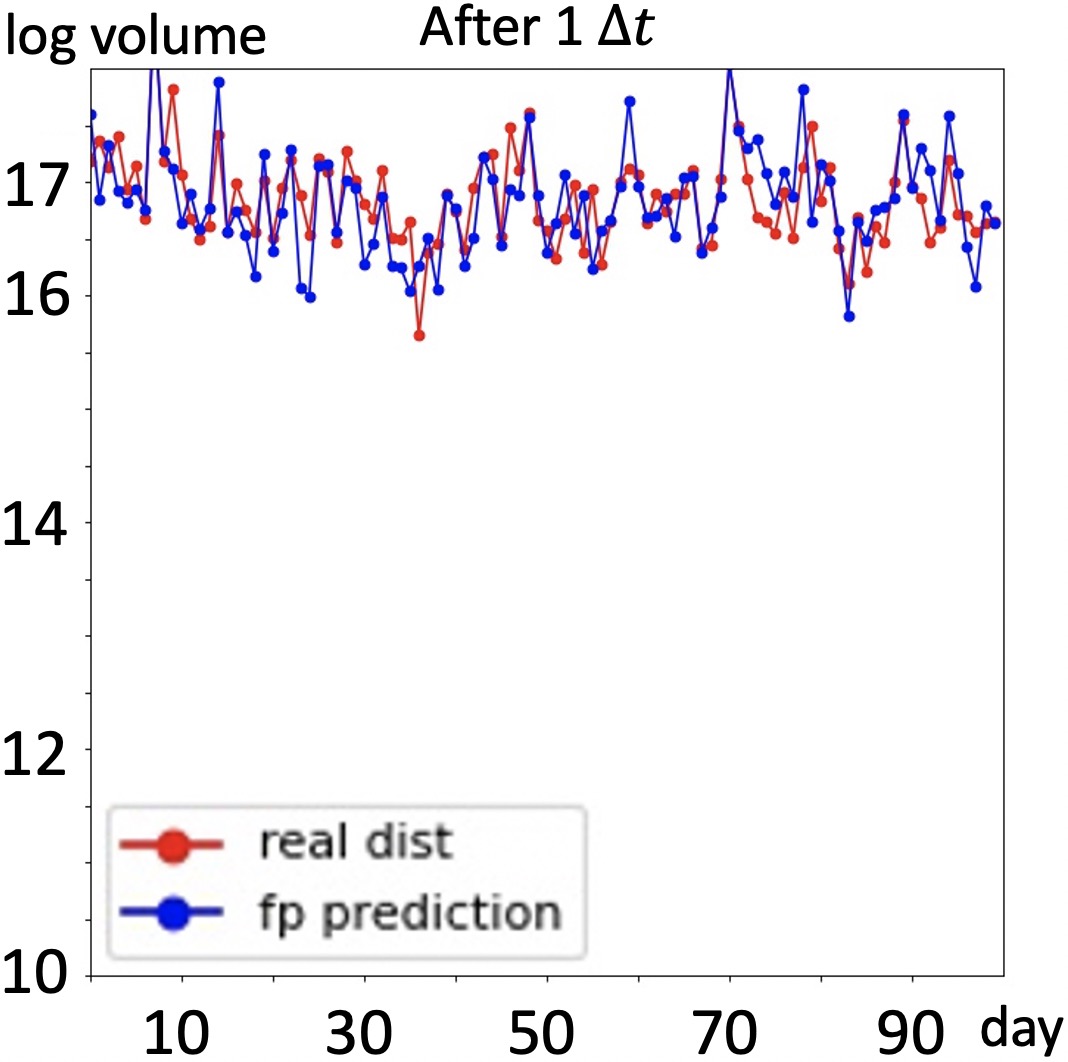}}\qquad
     \subfloat[][15:15]{\includegraphics[width=.35\linewidth]{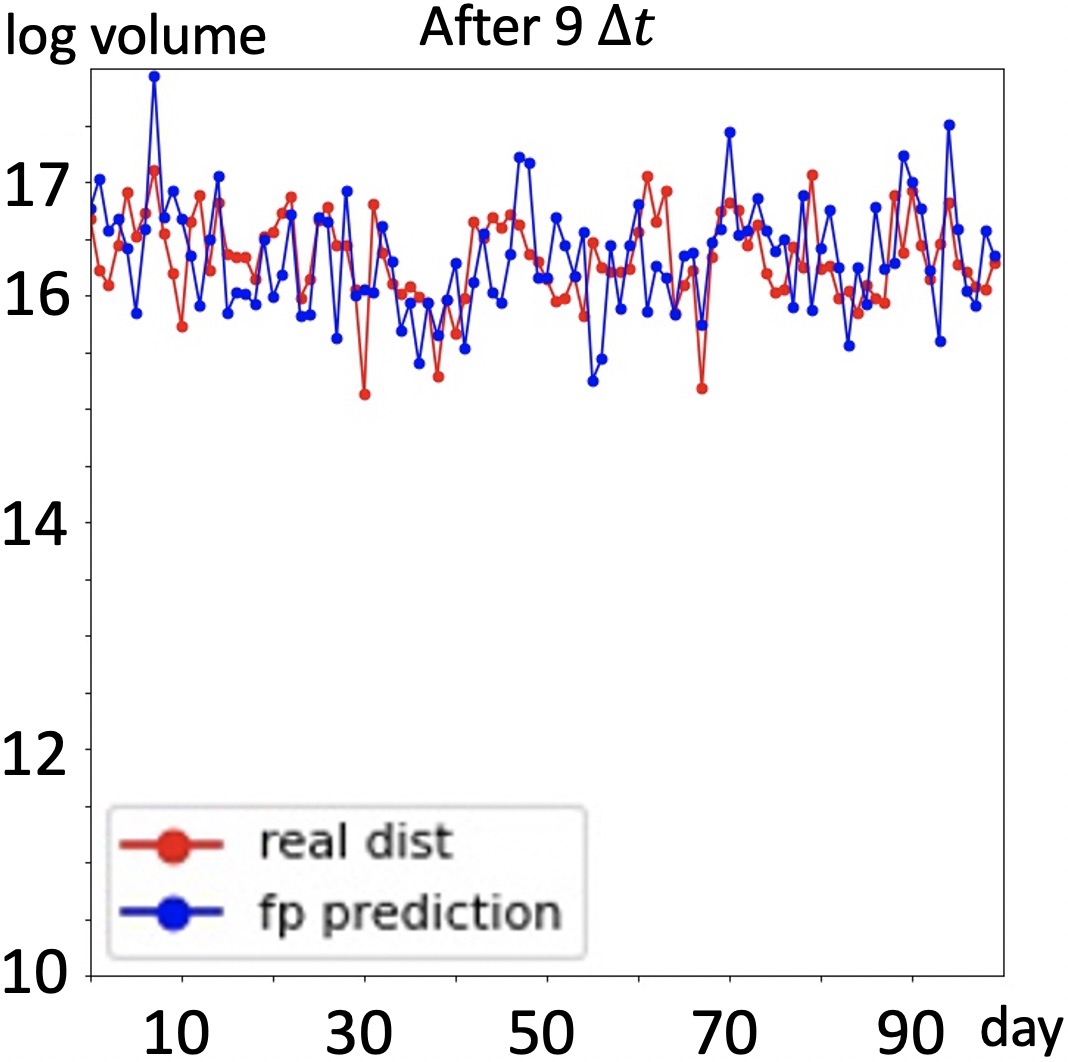}}\\
     \subfloat[][15:35]{\includegraphics[width=.35\linewidth]{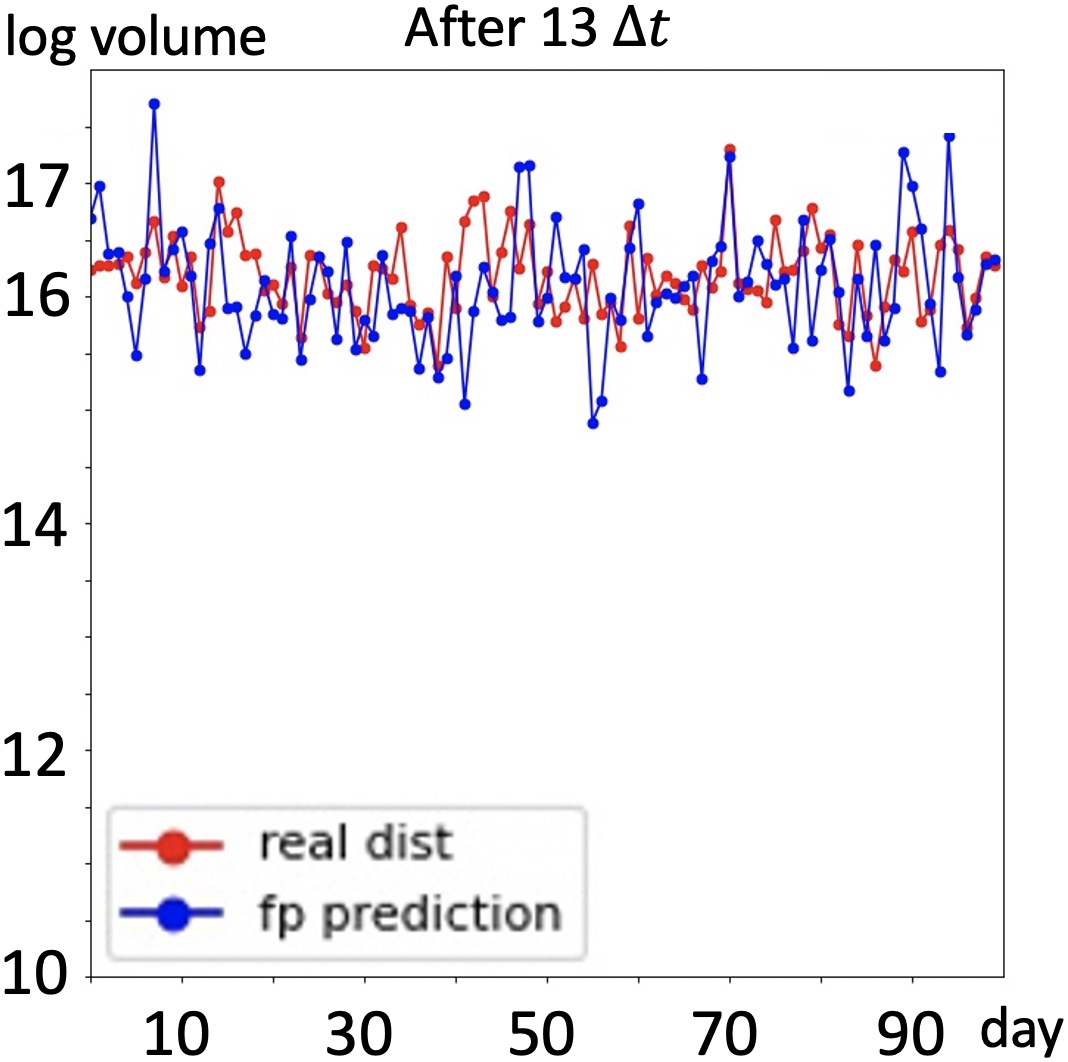}}\qquad
     \subfloat[][16:15]{\includegraphics[width=.35\linewidth]{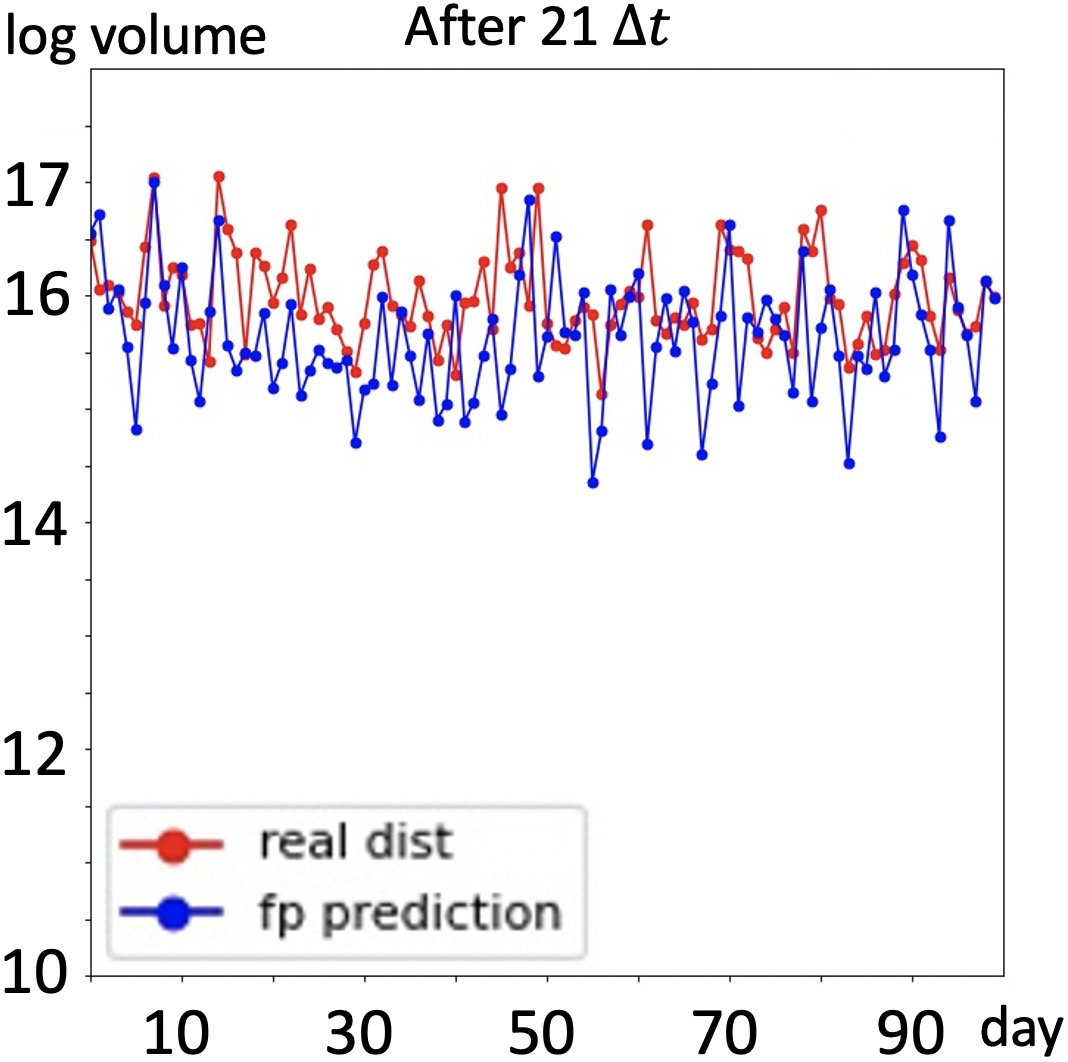}}
\caption{(a) to (d): Group A: with full trajectory of training data, predictions of traded volume in next 100 days, RM(yellow) fails to capture the regularities of traded volume in time series, kalman filter based model(green) fails to capture noise information and make reasonable predictions, our model(blue) is able to seize the movements of traded volume and yield better predictions. (e) to (h): Group B: predictions of our model without full trajectory.}
\label{fig:tradingvolume}
\end{figure}

In this section we would like to demonstrate the performance of our model in financial area. Trading volume is the total quantity of shares or contracts traded for specified securities such as stocks, bonds, options contracts, future contracts and all types of commodities. It can be measured on any type of security traded during a trading day or a specified time period. In our case, daily volume of trade is measured on stocks. Predicting traded volume is an essential component in financial research since the traded volume, as a basic component or input of other financial algorithms, tells investors the market's activity and liquidity.
The data set we use is the historical traded volume of the stock "JPM". The data covers period from January 2018 to January 2020 and is obtained from Bloomberg. Each day from 14:30 to 20:55, we have 1 observation every 5 minutes, totally 78 observations everyday. Our task is described as follows: we treat historical traded volume at 14:30, 14:40, 15:05, 15:20 and 16:20, namely, $\bm{x}_0, \bm{x}_2, \bm{x}_7, \bm{x}_{10}, \bm{x}_{22}$ as training data, each time point includes 730 samples. Then for next 100 days we predict traded volume at 14:35, 15:15, 15:35 and 16:15, namely, $\bm{x}_1, \bm{x}_9, \bm{x}_{13}, \bm{x}_{21}$. One of baselines we choose is classical rolling means(RM) method, which predicts intraday volume of a particular time interval by the average volume traded in the same interval over the past days. The other one baseline is a kalman filter based model \citep{kffortrading} that outperforms all available models in predicting intrady trading volume.


\textbf{Experiment Setup:}
Following similar setup as we did for RNA data set, we utilize the same structures for neural networks here. For hyperparameters we set $\Delta t = 0.02$, $T/\Delta t = 22$, it takes one single $\Delta t$ from $\bm{x}_t$ to $\bm{x}_{t+1}$. For preprocessing, we rescale data by taking natural logarithm of trading volume, which is a common way in trading volume research. We conduct experiments on two groups(A\&B) to show advantages of our method, for group A we train our model on complete data set, in this case the data has full trajectory; for group B we manually delete some trajectories of the data, for instance, we randomly kick out some samples of $\bm{x}_0, \bm{x}_2, \bm{x}_7, \bm{x}_{10}, \bm{x}_{22}$ then follow the same procedures of training and prediction.


\textbf{Results:}
We present prediction results in Figure \ref{fig:tradingvolume}. As shown in first four figures, with full trajectory, prediction made by RM is almost a straight line, the prediction value bouncing up and down within a very small range, thus this model cannot capture the volume movements, namely, regularities existing in the time series; prediction made by the Kalman filter based model captures the regularities better than RM model, but it fails to deal with noise component existing in the time series, thus some predictions are out of a reasonable range. Traded volume predicted by our model is closer to the real case, moreover, our model captures regularities meanwhile gives stable predictions. Furthermore, without full trajectory, Kalman filter based model fails to be applied here and RM model still fails to capture the regularities, we randomly drop half of the training samples and display predictions made by our model in last four figures of Figure \ref{fig:tradingvolume}, we see our model still works well.


\section{Discussions}
In this section we discuss the limitations and extension of our model.

\textbf{The challenge for non-uniqueness:}
Mathematically it is impossible to recover the exact drift term of an SDE if we are only given the information of density evolution on certain time intervals, because there might be infinitely many drift functions to induce the same density evolution. More precisely, suppose $p(x,t)$ solves FPE (\ref{eqn:fpe}), consider
\begin{align}
  0 = -\sum_{i=1}^D \frac{\partial }{\partial x_i} (u^i(\bm{x},t) p(\bm{x},t)) + \frac{\sigma^2}{2} \sum_{i=1}^D \frac{\partial^2}{\partial x_i^2}p(\bm{x},t).  \nonumber \label{hodg ex}
\end{align}

\begin{figure}[t!]
\centering
\subfloat[][True vector field]{\includegraphics[width=.4\linewidth]{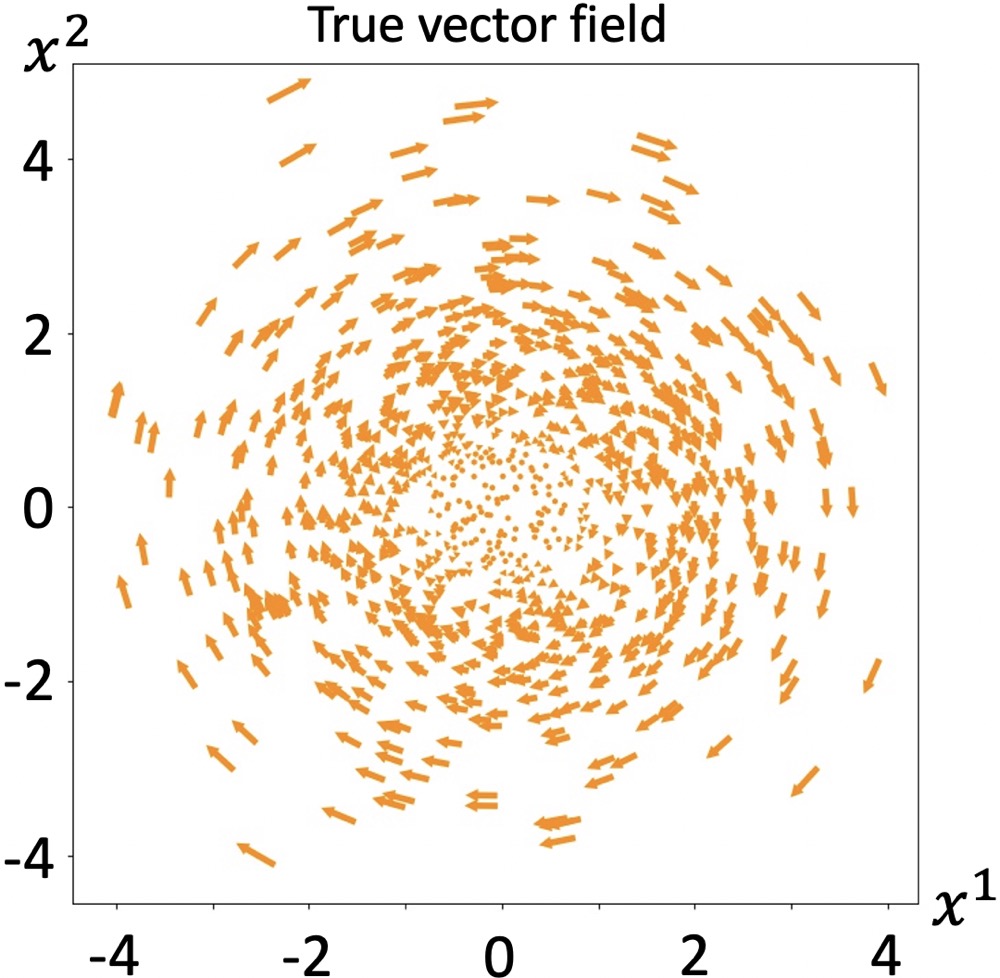}}\qquad
\subfloat[][Learned vector field]{\includegraphics[width=.4\linewidth]{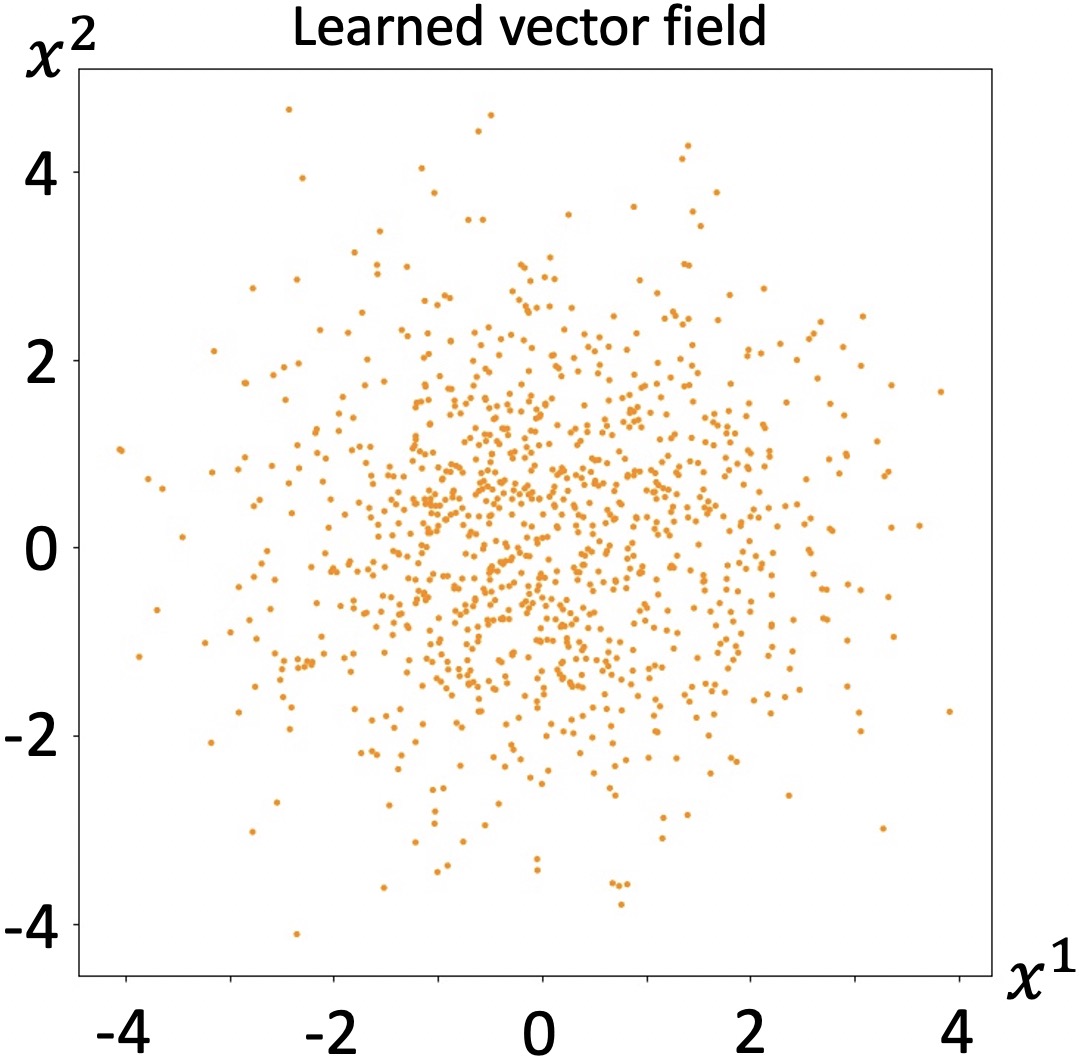}}
\caption{Results of learning curl field}
\label{fig:addround}
\end{figure}

\begin{figure}[t!]
     \centering
    \subfloat[][Prediction at $t_{10}$]{\includegraphics[width=.33\linewidth]{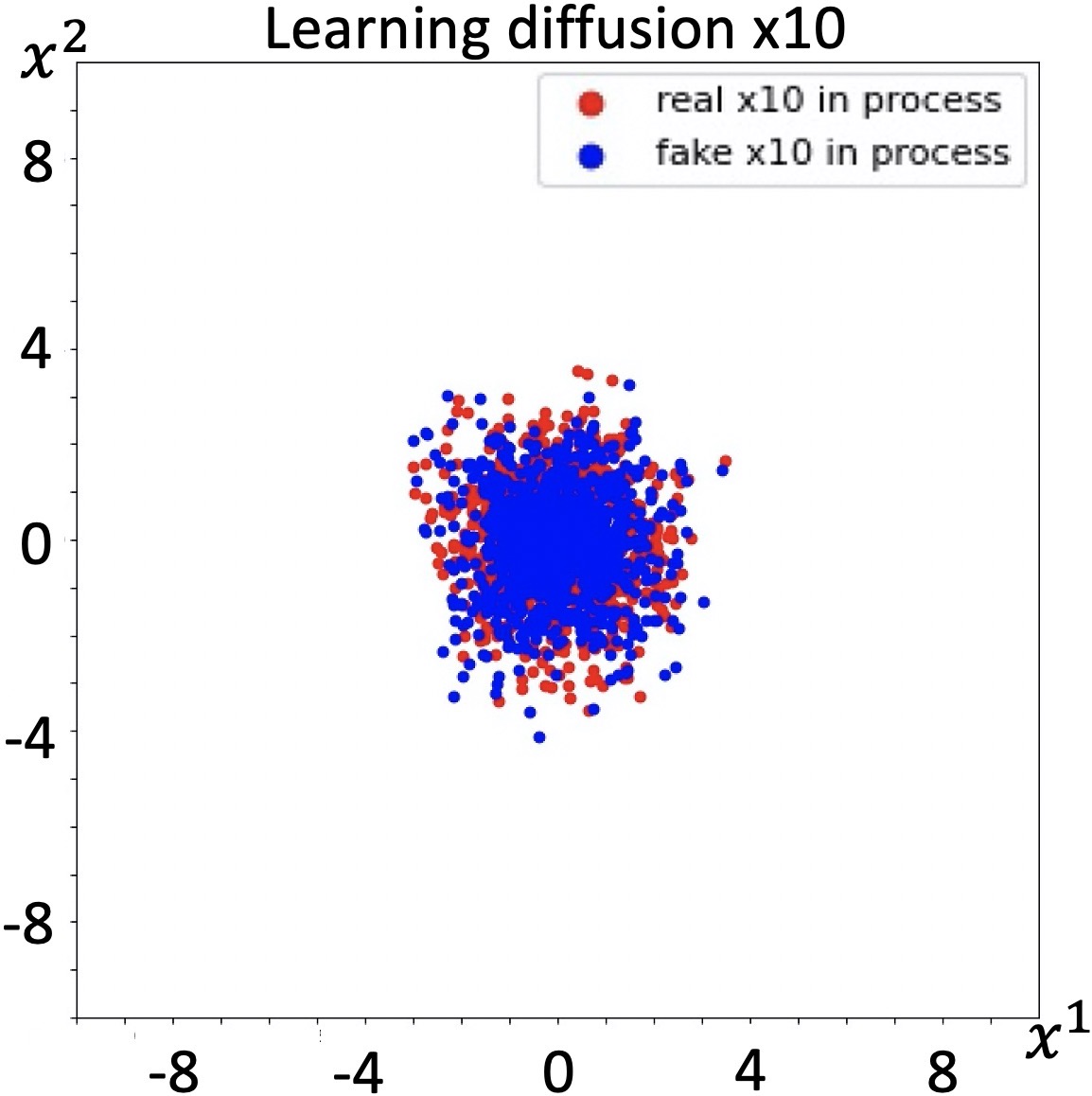}}
     \subfloat[][Prediction at $t_{30}$]{\includegraphics[width=.33\linewidth]{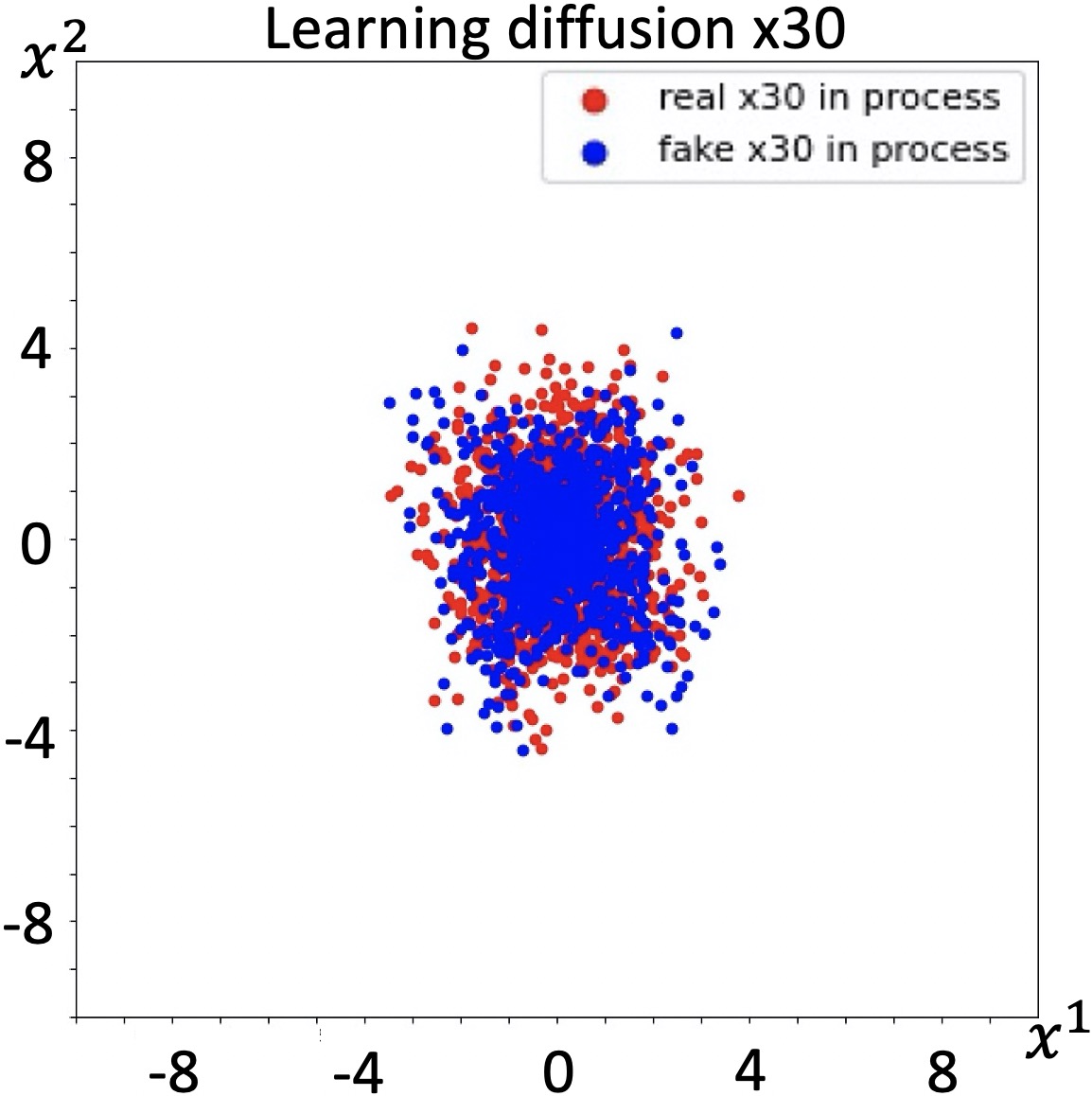}}
     \subfloat[][Prediction at $t_{50}$]{\includegraphics[width=.33\linewidth]{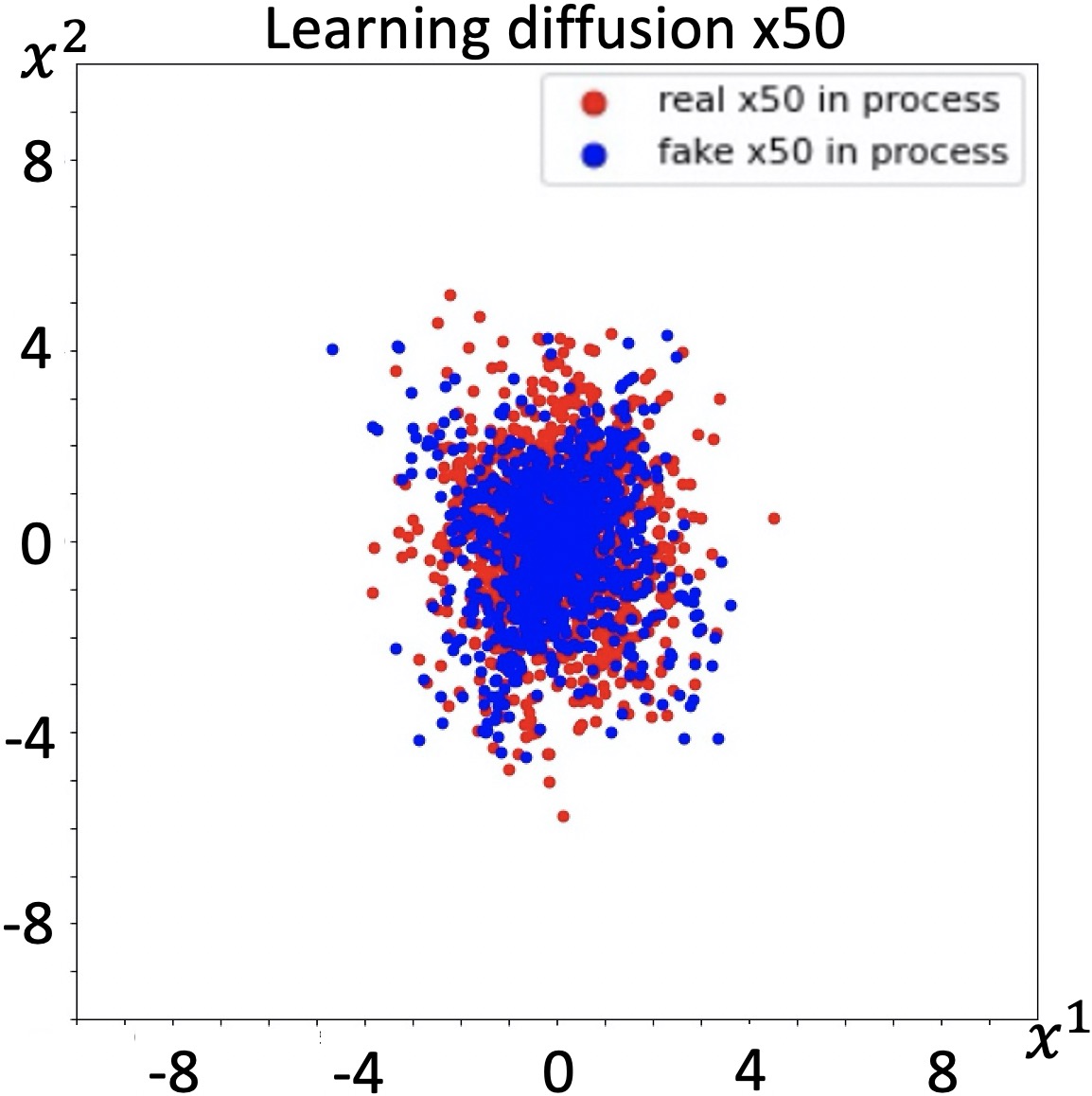}}\\
     \caption{Results of learning diffusion function}
     \label{fig:diff}
\end{figure}

One can prove, under mild assumptions, that there may be infinitely many vector fields $u(\bm{x},t)=(u^1(\bm{x},t),...,u^D(\bm{x},t))$ solving above equation. Therefore the solution to FPE (\ref{eqn:fpe}) with drift term $g(\bm{x},t)+u(\bm{x},t)$ is still $p(\bm{x},t)$, i.e. the vector field $u(\bm{x},t)$ never affects the density evolution of the dynamic. This illustrates that given the density evolution $p(\cdot, t)$, the solution for drift term is not necessarily unique. This clearly poses an essential difficulty of determining the exact drift term from the density. In this study, the main goal is to recover the entire density evolution (i.e. interpolate the density between observation time points) and predict how the density evolves in the future. As a result, although we cannot always acquire the exact drift term of the dynamic, we can still accurately recover and predict the density evolution. This is still meaningful and may find its application in various scientific domains.


\textbf{Curl field:} The drift function we showed in the synthetic experiments will apparently cause the evolution of the distribution. If the drift function is a curl, namely $g=\nabla \times \bm{F}$, then the distribution does not change, under this situation we cannot learn the density evolution since our algorithm depends on the change of the whole distribution. To demonstrate this point of view, we simulate a curl field $(y, -x)$ induced by $\bm{A} = [0,10;-10,0]$ on a Gaussian distribution that mean$=(0,0)$, covariance$=[2,0;0,2]$. Here we set noise part as 0. As shown in Figure \ref{fig:addround}, true and learned vector fields are indicated in $(a)$ and $(b)$ respectively. We see that the learned vectors are all "points", meaning the length of the vectors is "0", the algorithm fails to recover true vector field.


\textbf{Learning diffusion function:} Our framework also works for learning unknown diffusion function in the It\^{o} process. As an extension of our work, if we approximate the diffusion function with a neural network $\sigma_\eta$ (with parameters $\eta$), we revise the operator $\mathcal{F}$ as:
\begin{align}
    \mathcal{F}_{f}(X)=&\frac{1}{N}\sum\limits_{k=1}^N\Biggl(\sum_{i=1}^Dg^i_\omega(\bm{x}^{(k)})\frac{\partial}{\partial x_i}f(\bm{x}^{(k)}) \nonumber\\ &+\sum_{i=1}^D\biggl(\sum_{j=1}^D\frac{1}{2}(\sigma_{\eta}^{ij}(\bm{x}^{(k)}))^2\biggr)\frac{\partial^2}{\partial x_i^2}f(\bm{x}^{(k)})\Biggr),\nonumber
\end{align}
which can be derived by the same technique we used to derive Proposition \ref{prop:loss1}.

We test this formulation on a synthetic data set, where we only consider diffusion influence, namely, drift term in Equation \ref{lem:fpe} is ignored. We set the ground truth of diffusion coefficient as $\bm{\sigma} = [(1,0),(0,2)]$. We design the neural network as a simple one fully connected layer with 32 nodes, then show our result in Figure \ref{fig:diff}, we see that the predictions(blue) follow the same patterns as the ground truth(red) does.

\textbf{Future directions} It worth mentioning that our proposed algorithm \ref{alg:1} requires the gradient with respect to the parameter $\omega$ of drift term $g_\omega$ (i.e. line 10 of Algorithm \ref{alg:1}). Notice that each sample $\tilde{x}_{t_n}^{(k)}$ is computed from \eqref{n-step Euler Maruyama} for $n$ steps, thus each sample $\tilde{x}_{t_n}^{(k)}$ can be treated as $n$ compositions of drift term $g_\omega$, which may lead to more expensive computation. However, we cann avoid direct computation of gradient $\nabla_\omega$ by applying the adjoint method with Fokker-Planck equation \eqref{eqn:fpe} as the constraint \cite{pontryagin2018mathematical},\cite{zahr2016adjoint}. This is one of our future research directions.
Moreover, our model can also readily handle high dimensional cases by leveraging deep neural networks. Providing more numerical analysis such as compare trapezoidal rule and Runge-Kutta method in our scheme, as well as exploring super high dimensional applications are also appealing future directions.

\section{Conclusion}
In this paper, we formulate a novel method to recover the hidden dynamics from aggregate data. In particular, our work shows one
can simulate the evolving process of aggregate data as an It\^{o} process, in order to investigate aggregate data, we derive a new model that employs the weak form of FPE as well as the framework of WGAN. Furthermore, in Appendix we prove the theoretical guarantees of the error bound of our model. Finally we demonstrate our model through experiments on three synthetic data sets and two real-world data sets.

\bibliography{paper}

\begin{thebibliography}{45}
\providecommand{\natexlab}[1]{#1}
\providecommand{\url}[1]{\texttt{#1}}
\expandafter\ifx\csname urlstyle\endcsname\relax
  \providecommand{\doi}[1]{doi: #1}\else
  \providecommand{\doi}{doi: \begingroup \urlstyle{rm}\Url}\fi

\bibitem[Alshamaa et~al.(2019)Alshamaa, Chkeir, Mourad-Chehade, and
  Honeine]{hmm1}
Alshamaa, D., Chkeir, A., Mourad-Chehade, F., and Honeine, P.
\newblock Hidden markov model for indoor trajectory tracking of elderly people.
\newblock In \emph{IEEE Sensors Applications Symposium (SAS)}, 2019.

\bibitem[Arjovsky et~al.(2017)Arjovsky, Chintala, and Bottou]{wgan}
Arjovsky, M., Chintala, S., and Bottou, L.
\newblock Wasserstein gan.
\newblock In \emph{arXiv preprint arXiv:1701.07875}, 2017.

\bibitem[Atkinson(2008)]{atkinson2008introduction}
Atkinson, K.~E.
\newblock \emph{An introduction to numerical analysis}.
\newblock John wiley \& sons, 2008.

\bibitem[Beck et~al.(2018)Beck, Becker, Grohs, Jaafari, and
  Jentzen]{deepforsde1}
Beck, C., Becker, S., Grohs, P., Jaafari, N., and Jentzen, A.
\newblock Solving stochastic differential equations and kolmogorov equations by
  means of deep learning.
\newblock 2018.

\bibitem[Chen et~al.(2016)Chen, Feng, and Palomar]{kffortrading}
Chen, R., Feng, Y., and Palomar, D.
\newblock Forecasting intraday trading volume: a kalman filter approach.
\newblock In \emph{Available at SSRN 3101695}, 2016.

\bibitem[Cuturi(2013)]{sinkhorn}
Cuturi, M.
\newblock Sinkhorn distances: Lightspeed computation of optimal transport.
\newblock In \emph{Advances in Neural Information Processing Systems}, 2013.

\bibitem[Deriche et~al.(2020)Deriche, Absa, Amin, and Liu]{hmmvar}
Deriche, M., Absa, A.~A., Amin, A., and Liu, B.
\newblock A novel approach for salt dome detection and tracking using a hybrid
  hidden markov model with an active contour model.
\newblock \emph{Journal of Electrical Systems}, 16(3):\penalty0 276--294, 2020.

\bibitem[Djuric et~al.(2003)Djuric, Kotecha, Zhang, Huang, Ghirmai, Bugallo,
  and Miguez]{djuric2003particle}
Djuric, P.~M., Kotecha, J.~H., Zhang, J., Huang, Y., Ghirmai, T., Bugallo,
  M.~F., and Miguez, J.
\newblock Particle filtering.
\newblock \emph{IEEE Signal Processing Magazine}, 20\penalty0 (5):\penalty0
  19--38, 2003.

\bibitem[Eddy(1996)]{eddy1996hidden}
Eddy, S.~R.
\newblock Hidden markov models.
\newblock \emph{Current Opinion in Structural Biology}, 6\penalty0
  (3):\penalty0 361--365, 1996.

\bibitem[Fang et~al.(2019)Fang, Wang, Yao, Zhao, Zhao, and Zha]{pf2}
Fang, Y., Wang, C., Yao, W., Zhao, X., Zhao, H., and Zha, H.
\newblock On-road vehicle tracking using part-based particle filter.
\newblock \emph{IEEE Transactions on Intelligent Transportation Systems},
  20(12):\penalty0 4538--4552, 2019.

\bibitem[Farahi \& Yazdi(2020)Farahi and Yazdi]{kf1}
Farahi, F. and Yazdi, H.~S.
\newblock Probabilistic kalman filter for moving object tracking.
\newblock \emph{Signal Processing: Image Communication}, 82, 2020.

\bibitem[Glorot \& Bengio(2010)Glorot and Bengio]{xvaier}
Glorot, X. and Bengio, Y.
\newblock Understanding the difficulty of training deep feedforward neural
  networks.
\newblock In \emph{International Conference on Artificial Intelligence and
  Statistics}, 2010.

\bibitem[Greydanus et~al.(2019)Greydanus, Dzamba, and Yosinski]{hnn}
Greydanus, S., Dzamba, M., and Yosinski, J.
\newblock Hamiltonian neural networks.
\newblock \emph{arXiv preprint arXiv:1906.01563}, 2019.

\bibitem[Harvey(1990)]{harvey1990forecasting}
Harvey, A.~C.
\newblock \emph{Forecasting, structural time series models and the Kalman
  filter}.
\newblock Cambridge University Press, 1990.

\bibitem[Hashimoto et~al.(2016)Hashimoto, Gifford, and Jaakkola]{hashinn}
Hashimoto, T., Gifford, D., and Jaakkola, T.
\newblock Learning population-level diffusions with generative rnns.
\newblock In \emph{International Conference on Machine Learning}, pp.\
  2417--2426, 2016.

\bibitem[Hefny et~al.(2015)Hefny, Downey, and Gordon]{hefny2015supervised}
Hefny, A., Downey, C., and Gordon, G.~J.
\newblock Supervised learning for dynamical system learning.
\newblock In \emph{Neural Information Processing Systems}, 2015.

\bibitem[Hicks et~al.(2015)Hicks, Teng, and Irizarry]{Hicks025528}
Hicks, S.~C., Teng, M., and Irizarry, R.~A.
\newblock On the widespread and critical impact of systematic bias and batch
  effects in single-cell rna-seq data.
\newblock \emph{bioRxiv}, 2015.

\bibitem[Jordan et~al.(1998)Jordan, Kinderlehrer, and Otto]{jko}
Jordan, R., Kinderlehrer, D., and Otto, F.
\newblock The variational formulation of the fokker--planck equation.
\newblock \emph{SIAM journal on mathematical analysis}, 29(1):\penalty0 1--17,
  1998.

\bibitem[Kalman(1960)]{originalkf}
Kalman, R.~E.
\newblock A new approach to linear filtering and prediction problems.
\newblock \emph{arXiv preprint arXiv:1805.04099}, 1960.

\bibitem[Kingma \& Ba(2014)Kingma and Ba]{kingma2014adam}
Kingma, D.~P. and Ba, J.
\newblock Adam: A method for stochastic optimization, 2014.

\bibitem[Klein et~al.(2015)Klein, Mazutis, Akartuna, Tallapragada, Veres, Li,
  Peshkin, Weitz, and Kirschner]{klein2015droplet}
Klein, A., Mazutis, L., Akartuna, I., Tallapragada, N., Veres, A., Li, V.,
  Peshkin, L., Weitz, D., and Kirschner, M.
\newblock Droplet barcoding for single-cell transcriptomics applied to
  embryonic stem cells.
\newblock \emph{Cell}, 161\penalty0 (5):\penalty0 1187--1201, 2015.

\bibitem[Kloeden \& Platen(2013)Kloeden and Platen]{kloeden2013numerical}
Kloeden, P.~E. and Platen, E.
\newblock \emph{Numerical solution of stochastic differential equations},
  volume~23.
\newblock Springer Science \& Business Media, 2013.

\bibitem[Langford et~al.(2009)Langford, Salakhutdinov, and
  Zhang]{langford2009learning}
Langford, J., Salakhutdinov, R., and Zhang, T.
\newblock Learning nonlinear dynamic models.
\newblock In \emph{International Conference on Machine Learning}, pp.\
  593–600, 2009.

\bibitem[Li et~al.(2019)Li, Liu, Zha, and Zhou]{parafpe}
Li, W., Liu, S., Zha, H., and Zhou, H.
\newblock Parametric fokker-planck equation.
\newblock In \emph{Geometry science of information}, 2019.

\bibitem[Li(2018)]{vanderpol}
Li, Y.
\newblock A data-driven method for the steady state of randomly perturbed
  dynamics.
\newblock \emph{arXiv preprint arXiv:1805.04099}, 2018.

\bibitem[Liu \& Wang(2016)Liu and Wang]{variationalgd}
Liu, Q. and Wang, D.
\newblock Stein variational gradient descent: A general purpose bayesian
  inference algorithm.
\newblock In \emph{Neural Information Processing Systems}, pp.\  2378--2386,
  2016.

\bibitem[Milstein \& Tretyakov(2013)Milstein and Tretyakov]{numericalsde}
Milstein, G.~N. and Tretyakov, M.~V. (eds.).
\newblock \emph{Stochastic numerics for mathematical physics}.
\newblock Springer Science \& Business Media, 2013.

\bibitem[Miyato et~al.(2018)Miyato, Kataoka, Koyama, and Yoshida]{spectral}
Miyato, T., Kataoka, T., Koyama, M., and Yoshida, Y.
\newblock Spectral normalization for generative adversarial networks.
\newblock \emph{arXiv preprint arXiv:1802.05957}, 2018.

\bibitem[Nelson(1985)]{quantum1}
Nelson, E.
\newblock \emph{Quantum fluctuations}.
\newblock Princeton University Press, 1985.

\bibitem[{\O}ksendal(2003)]{oksendal2003stochastic}
{\O}ksendal, B.
\newblock Stochastic differential equations.
\newblock In \emph{Stochastic differential equations}, pp.\  65--84. Springer,
  2003.

\bibitem[Pavon et~al.(2018)Pavon, Tabak, and Trigila]{datadrivensb}
Pavon, M., Tabak, E.~G., and Trigila, G.
\newblock The data-driven schroedinger bridge.
\newblock \emph{arXiv preprint arXiv:1806.01364}, 2018.

\bibitem[Pontryagin(2018)]{pontryagin2018mathematical}
Pontryagin, L.~S.
\newblock \emph{Mathematical theory of optimal processes}.
\newblock Routledge, 2018.

\bibitem[Qi \& Majda(2016)Qi and Majda]{lowdim}
Qi, D. and Majda, A.
\newblock Low-dimensional reduced-order models for statistical response and
  uncertainty quantification: Two-layer baroclinic turbulence.
\newblock \emph{Journal of the Atmospheric Sciences}, 73(12):\penalty0
  4609--4639, 2016.

\bibitem[Rezende \& Mohamed(2015)Rezende and Mohamed]{varinf}
Rezende, D. and Mohamed, S.
\newblock Variational inference with normalizing flows.
\newblock In \emph{arXiv preprint arXiv:1505.05770}, 2015.

\bibitem[Risken(1989)]{thefpe}
Risken, H.
\newblock The fokker-planck equation.
\newblock \emph{Springer Series in Synergetics}, 18:\penalty0 4609--4639, 1989.

\bibitem[Risken \& Caugheyz(1991)Risken and Caugheyz]{fokkerplanckbook}
Risken, H. and Caugheyz, T. (eds.).
\newblock \emph{The fokker-planck equation: Methods of solution and
  application}.
\newblock Springer, 1991.

\bibitem[Santos et~al.(2019)Santos, Lobo, and Bernardino]{pf1}
Santos, N.~P., Lobo, V., and Bernardino, A.
\newblock Unmanned aerial vehicle tracking using a particle filter based
  approach.
\newblock In \emph{IEEE Underwater Technology (UT)}, 2019.

\bibitem[Singh et~al.(2020)Singh, Zhang, and Chen]{hmmaggregate}
Singh, R., Zhang, Q., and Chen, Y.
\newblock Learning hidden markov models from aggregate observations.
\newblock \emph{arXiv preprint arXiv:2011.11236}, 2020.

\bibitem[Sirignano \& Spiliopoulos(2018)Sirignano and
  Spiliopoulos]{sirignano2018dgm}
Sirignano, J. and Spiliopoulos, K.
\newblock Dgm: A deep learning algorithm for solving partial differential
  equations.
\newblock \emph{Journal of computational physics}, 375:\penalty0 1339--1364,
  2018.

\bibitem[Villani(2008)]{villani2008optimal}
Villani, C.
\newblock \emph{Optimal transport: old and new}, volume 338.
\newblock Springer Science \& Business Media, 2008.

\bibitem[Wang et~al.(2018)Wang, Dai, Kong, Erfani, Bailey, and
  Zha]{yisen2018aggregate}
Wang, Y., Dai, B., Kong, L., Erfani, S.~M., Bailey, J., and Zha, H.
\newblock Learning deep hidden nonlinear dynamics from aggregate data.
\newblock In \emph{Uncertainty in Artificial Intelligence}, 2018.

\bibitem[Weinan et~al.(2017)Weinan, Han, and Jentzen]{deepforsde}
Weinan, E., Han, J., and Jentzen, A.
\newblock Deep learning-based numerical methods for high-dimensional parabolic
  partial differential equations and backward stochastic differential
  equations.
\newblock In \emph{Communications in Mathematics and Statistics}, pp.\
  349--380, 2017.

\bibitem[Zahr \& Persson(2016)Zahr and Persson]{zahr2016adjoint}
Zahr, M.~J. and Persson, P.-O.
\newblock An adjoint method for a high-order discretization of deforming domain
  conservation laws for optimization of flow problems.
\newblock \emph{Journal of Computational Physics}, 326:\penalty0 516--543,
  2016.

\bibitem[Zang et~al.(2019)Zang, Bao, Ye, and Zhou]{wan}
Zang, Y., Bao, G., Ye, X., and Zhou, H.
\newblock Weak adversarial networks for high-dimensional partial differential
  equations.
\newblock In \emph{arXiv preprint arXiv:1907.08272}, 2019.

\bibitem[Zienkiewicz \& Cheung(1971)Zienkiewicz and
  Cheung]{zienkiewicz1971finite}
Zienkiewicz, O. and Cheung, I.
\newblock \emph{The Finite Element Method in Engineering Science}.
\newblock McGraw-Hill European Publishing Programme. McGraw-Hill, 1971.
\newblock ISBN 9780070941380.
\newblock URL \url{https://books.google.com/books?id=B99RAAAAMAAJ}.

\end{thebibliography}

\bibliographystyle{icml2021}

\newpage
\onecolumn
\appendix

\section{Supplementary Experiments}
\subsection{RNA-sequence}
\begin{figure}[ht]
     \centering
     \subfloat[][D2 of Mt1]{\includegraphics[width=.25\linewidth]{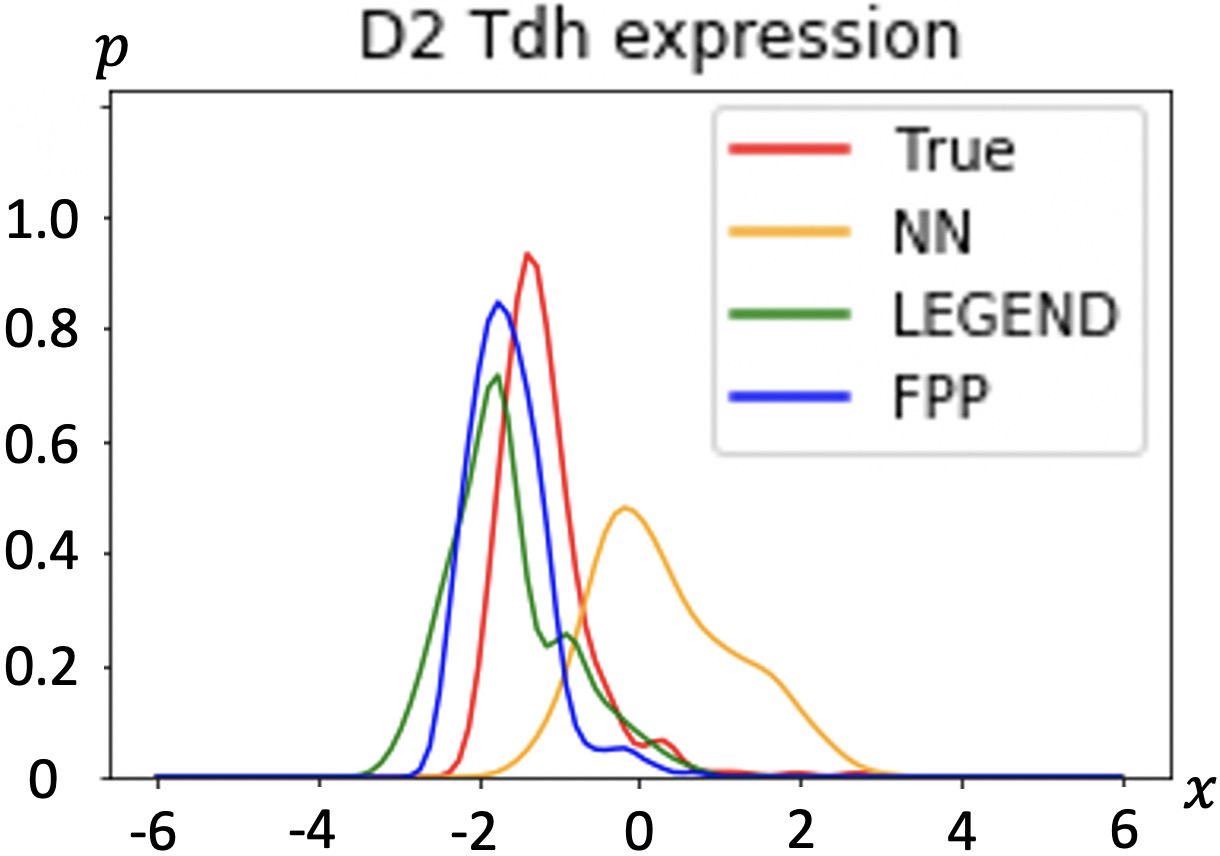}}
     \subfloat[][D7 of Mt1]{\includegraphics[width=.25\linewidth]{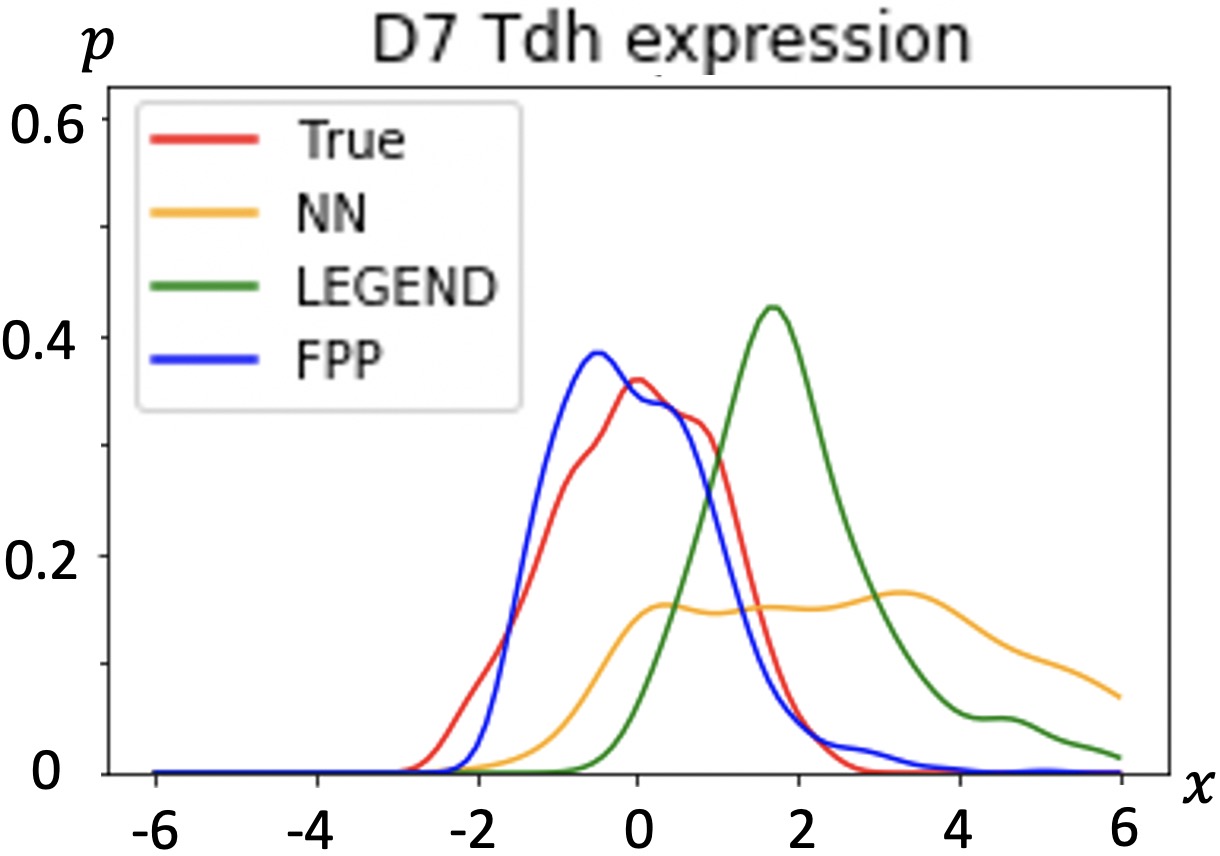}}
     \subfloat[][D2 of Mt2]{\includegraphics[width=.25\linewidth]{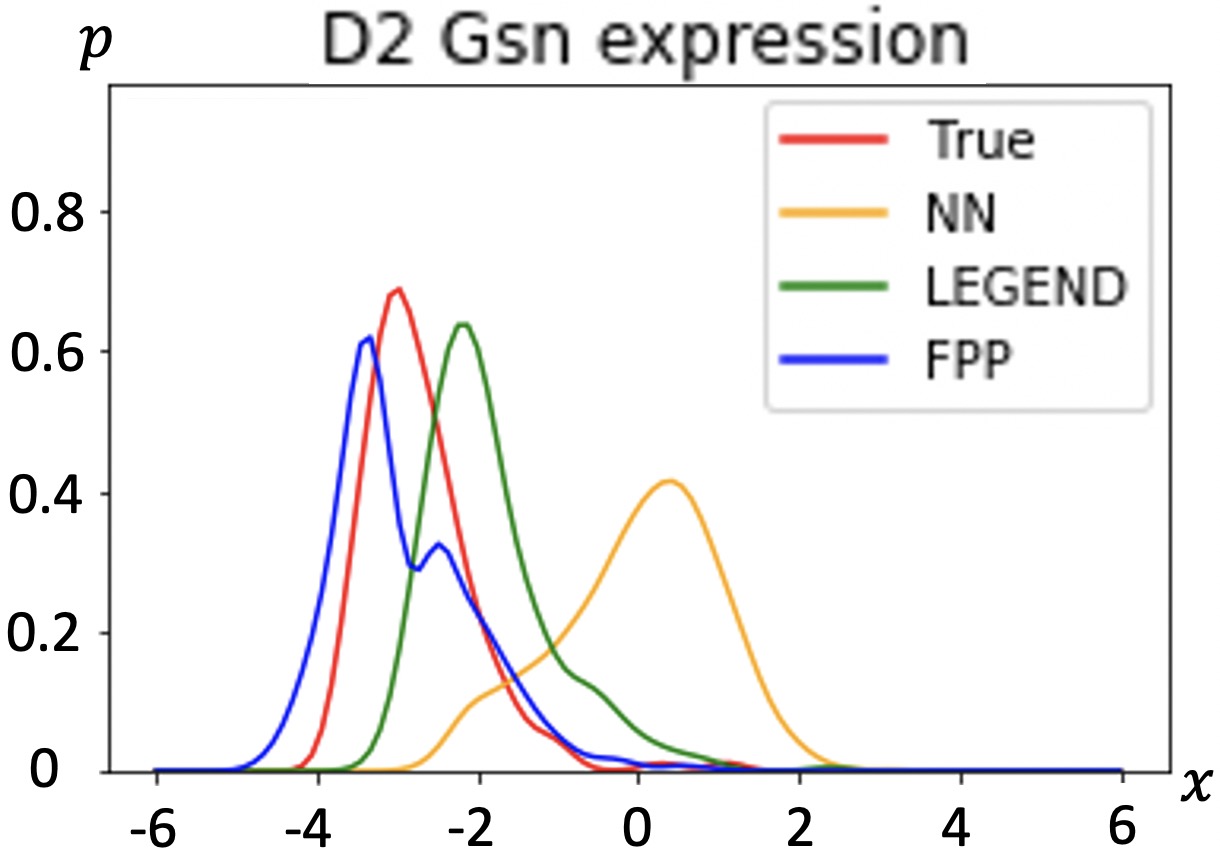}}
     \subfloat[][D7 of Mt2]{\includegraphics[width=.25\linewidth]{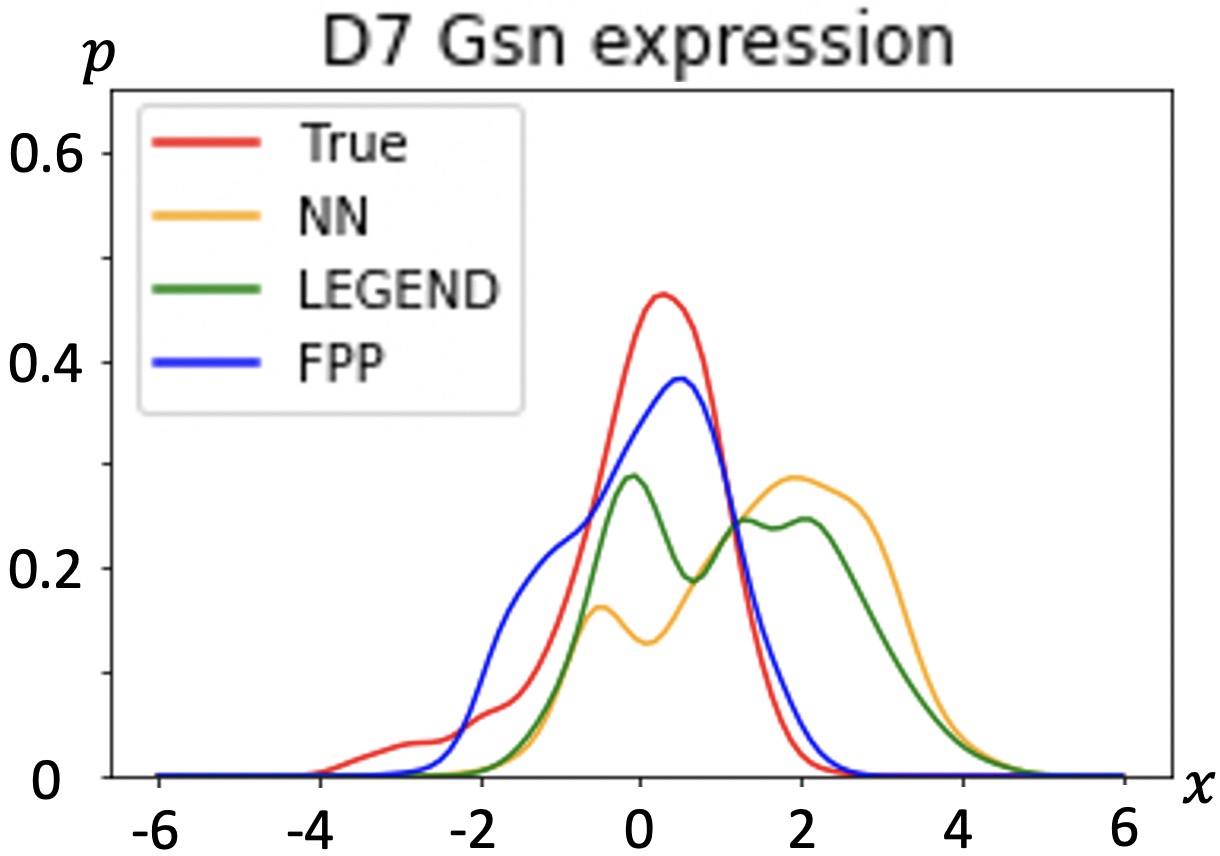}}
\caption{The performance comparisions among different models on D2 and D7 of Tdh and Gsn.}
\label{fig:gene_add}
\end{figure}
\begin{table}[ht]
\centering
\caption{The Wasserstein error of different models on
Supplementary RNA-sequence data sets.}
\setlength\tabcolsep{4pt}
\begin{tabular}{l|c|c|c c c}
\hline\hline
Data & Task & Dimension & NN & LEGEND & Ours
\\ [0.5ex]
\hline
\multirow{2}{*}{RNA-Tdh} & D2 & 10 & 16.28 & 5.75 & \textbf{2.15}\\\cline{2-2}
  & D7 & 10 & 28.19 & 22.49 & \textbf{1.03}\\
\hline
\multirow{2}{*}{RNA-Gsn} & D2 & 10 & 34.94 & 10.77 & \textbf{3.31}\\\cline{2-2}
  & D7 & 10 & 15.74 & 10.42 & \textbf{2.07}\\
\hline\hline
\end{tabular}
\label{tab:PPer_add}
\end{table}

\subsection{Daily Trading Volume}
\begin{figure}[h!]
     \centering
    \subfloat[][14:35]{\includegraphics[width=.2\linewidth]{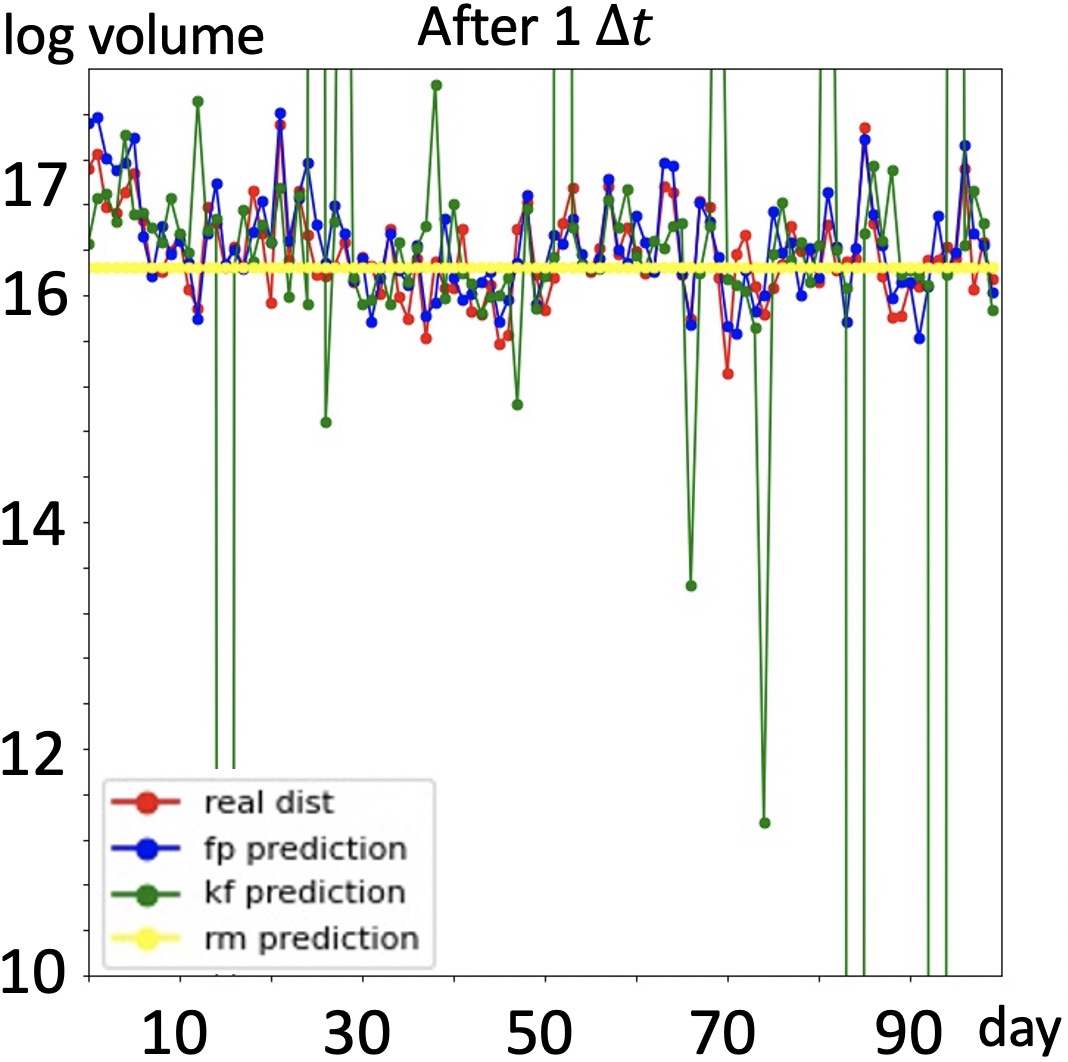}}
     \subfloat[][15:15]{\includegraphics[width=.2\linewidth]{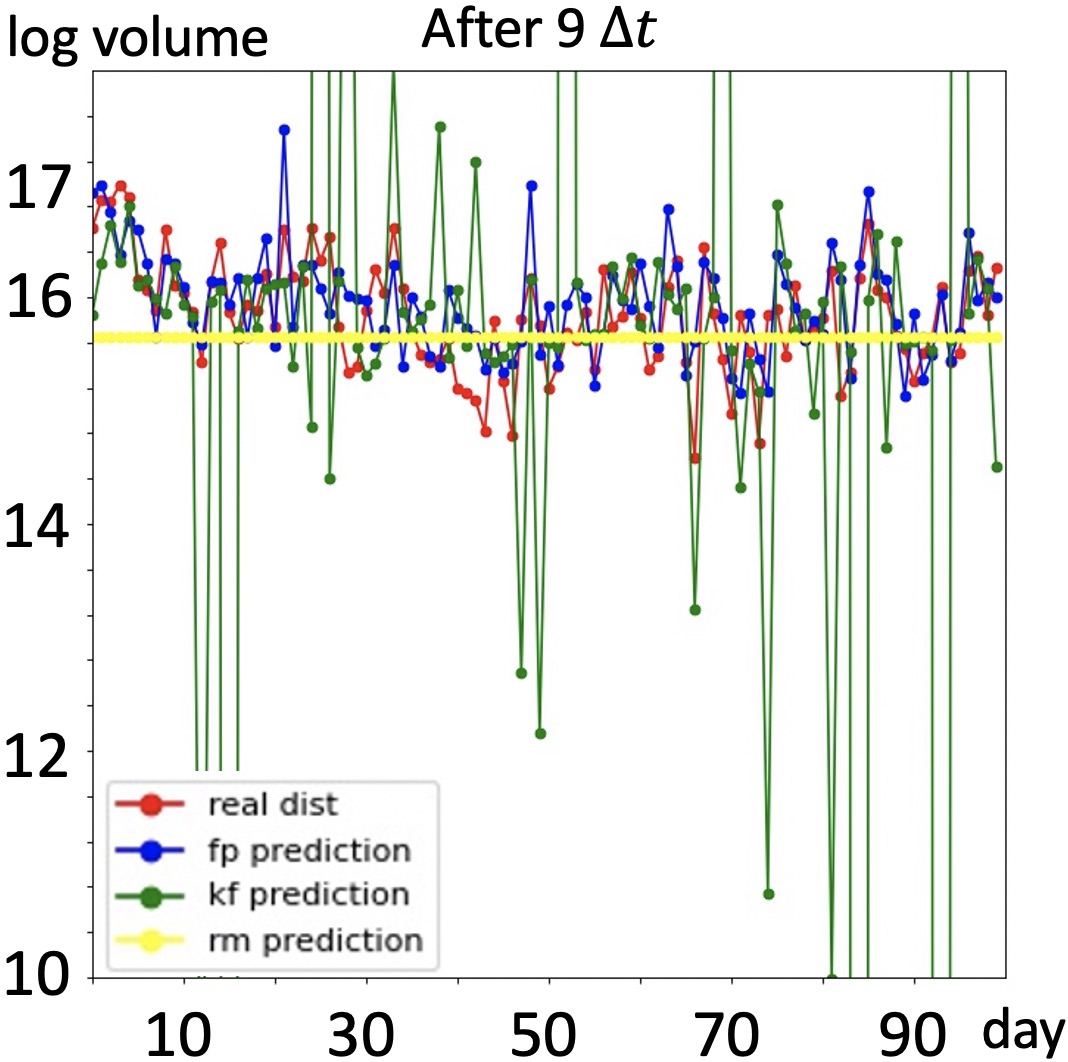}}
     \subfloat[][15:35]{\includegraphics[width=.2\linewidth]{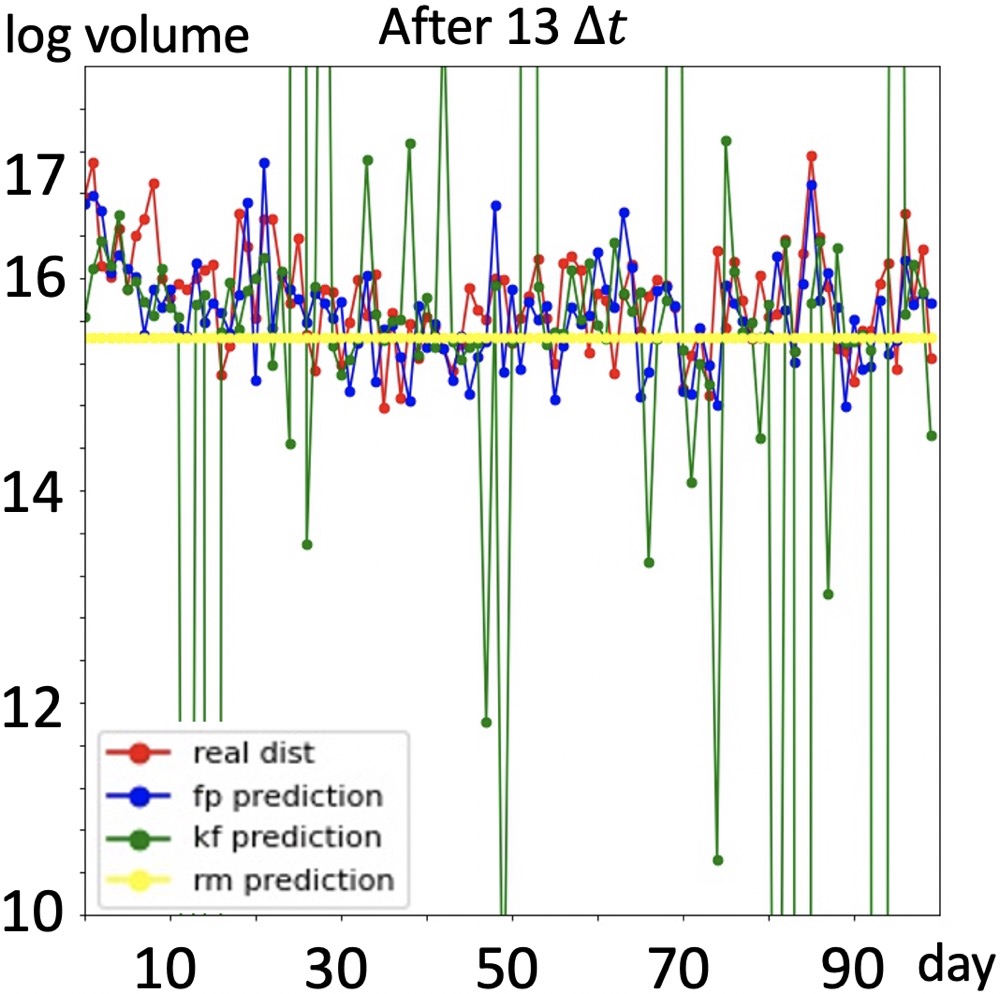}}
     \subfloat[][16:15]{\includegraphics[width=.2\linewidth]{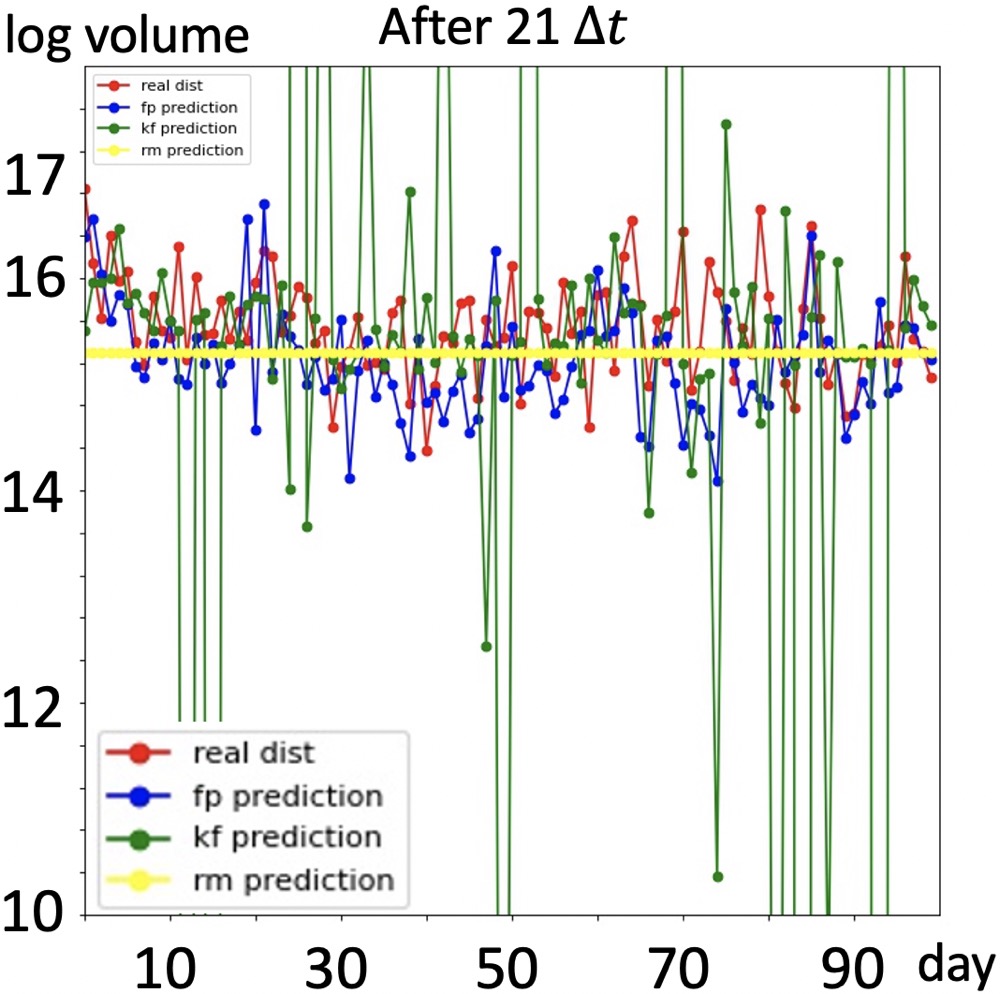}}\\
     \subfloat[][14:35]{\includegraphics[width=.2\linewidth]{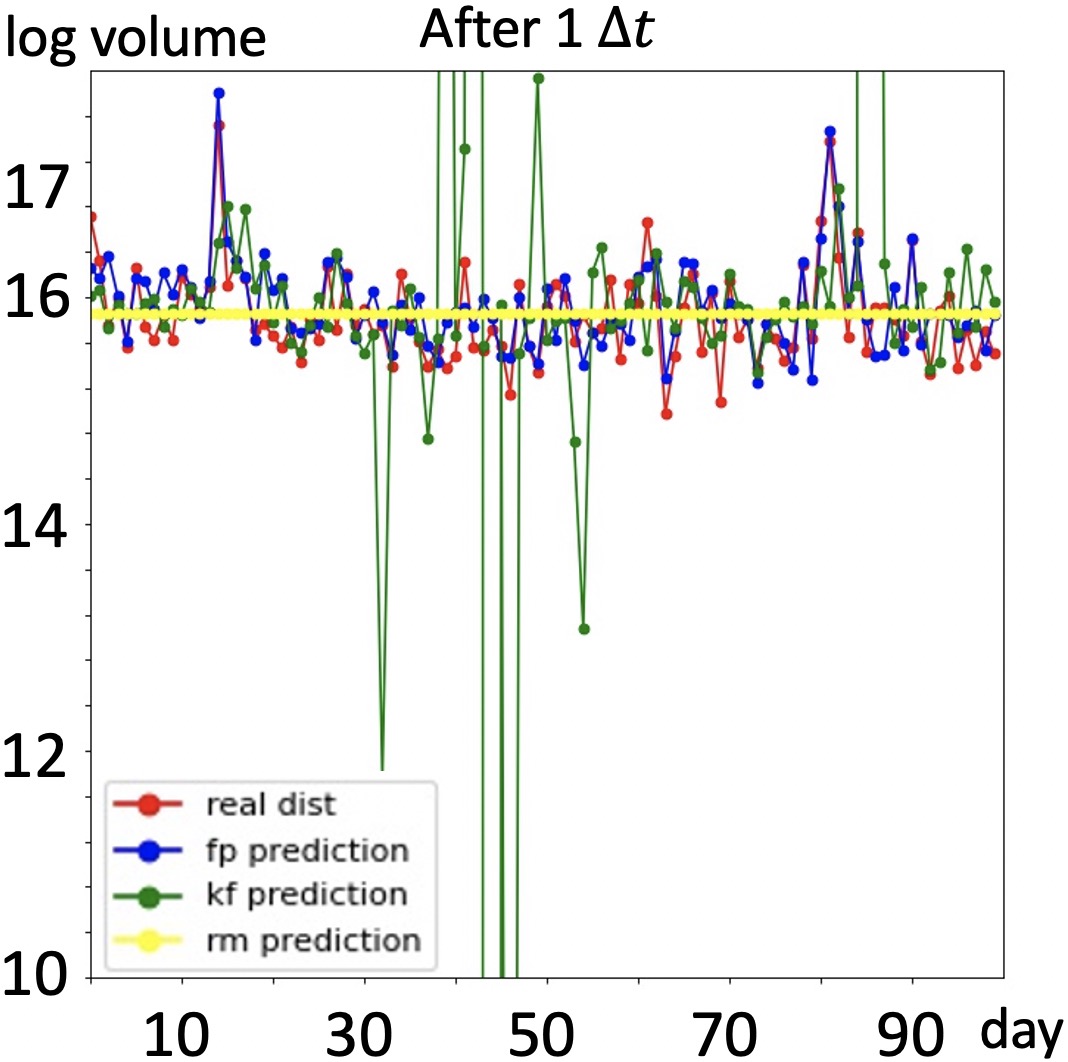}}
     \subfloat[][15:15]{\includegraphics[width=.2\linewidth]{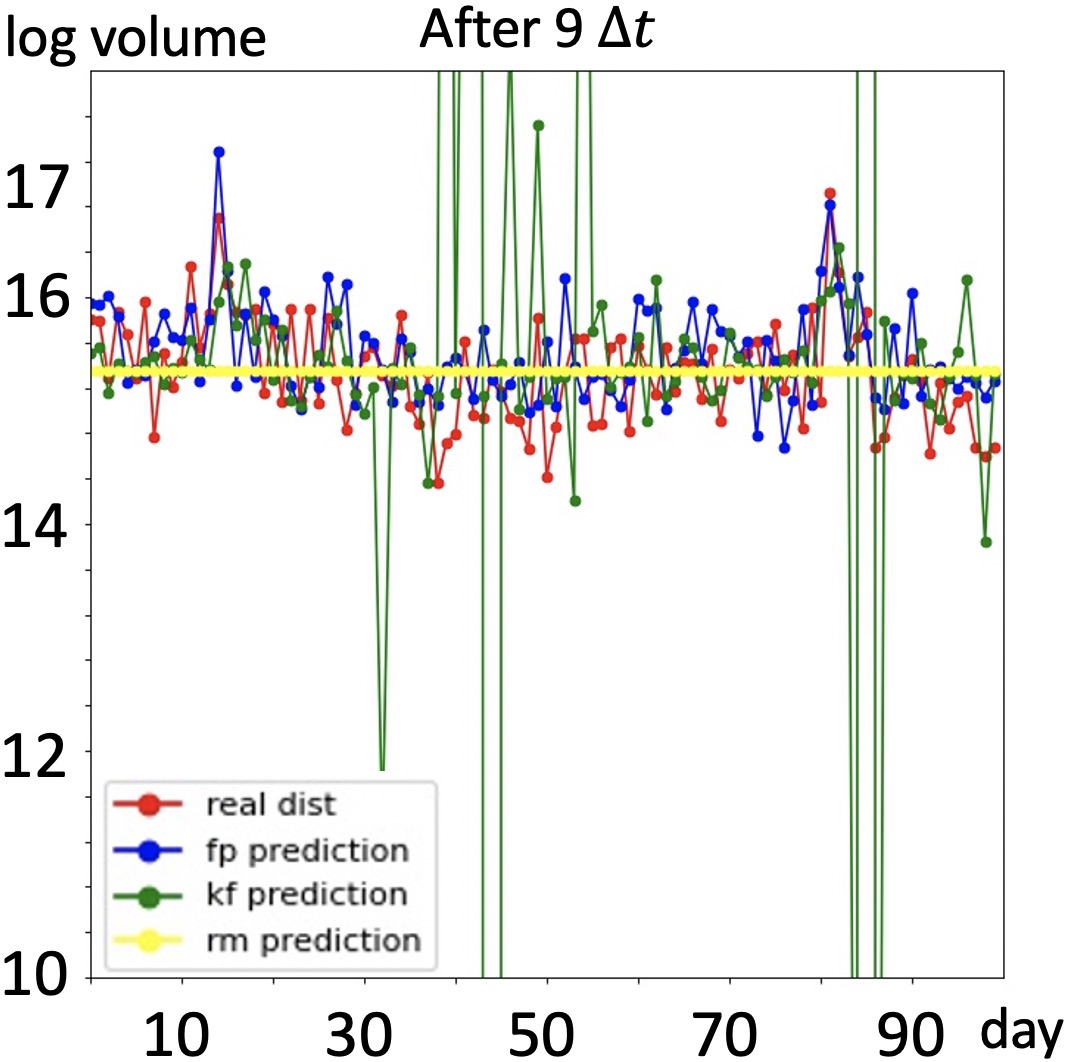}}
     \subfloat[][15:35]{\includegraphics[width=.2\linewidth]{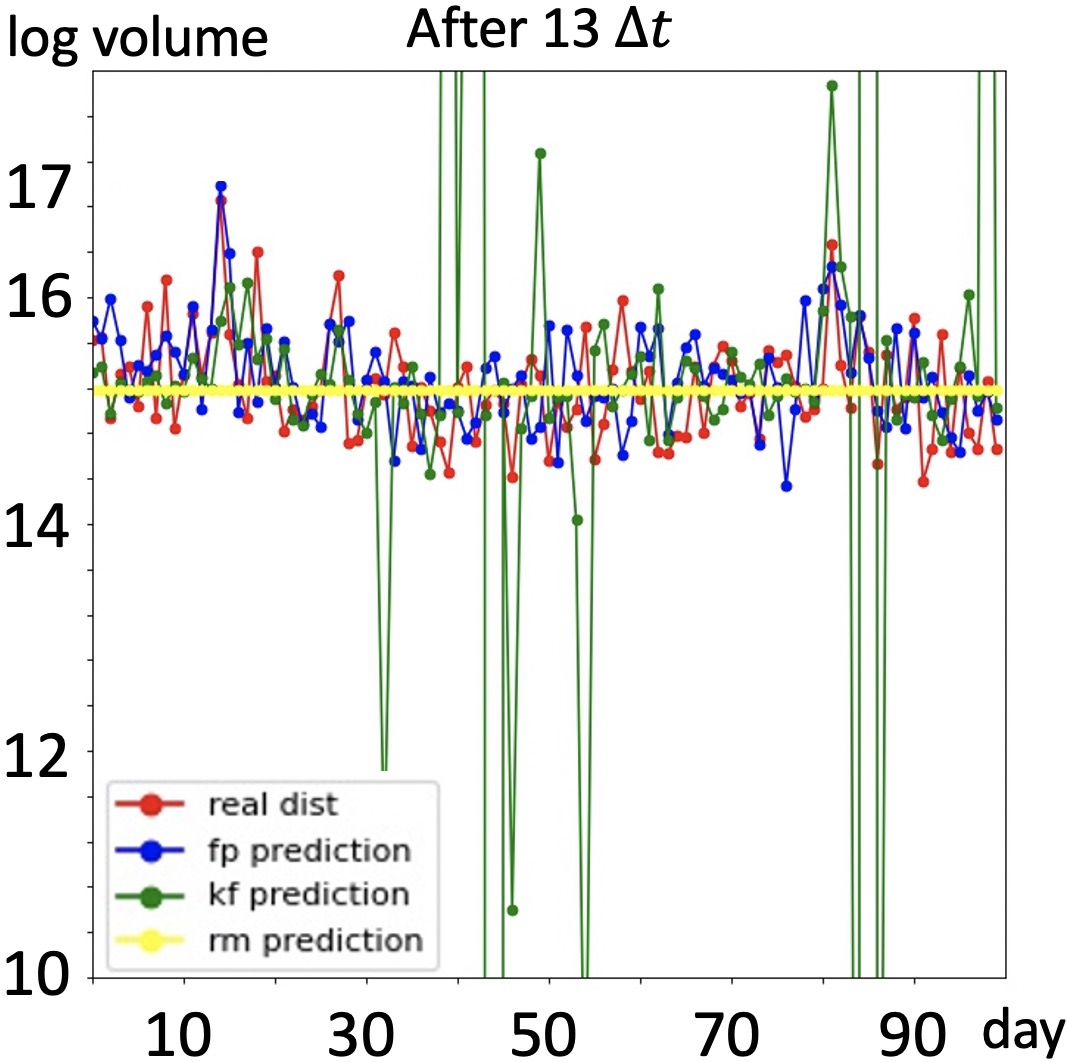}}
     \subfloat[][16:15]{\includegraphics[width=.2\linewidth]{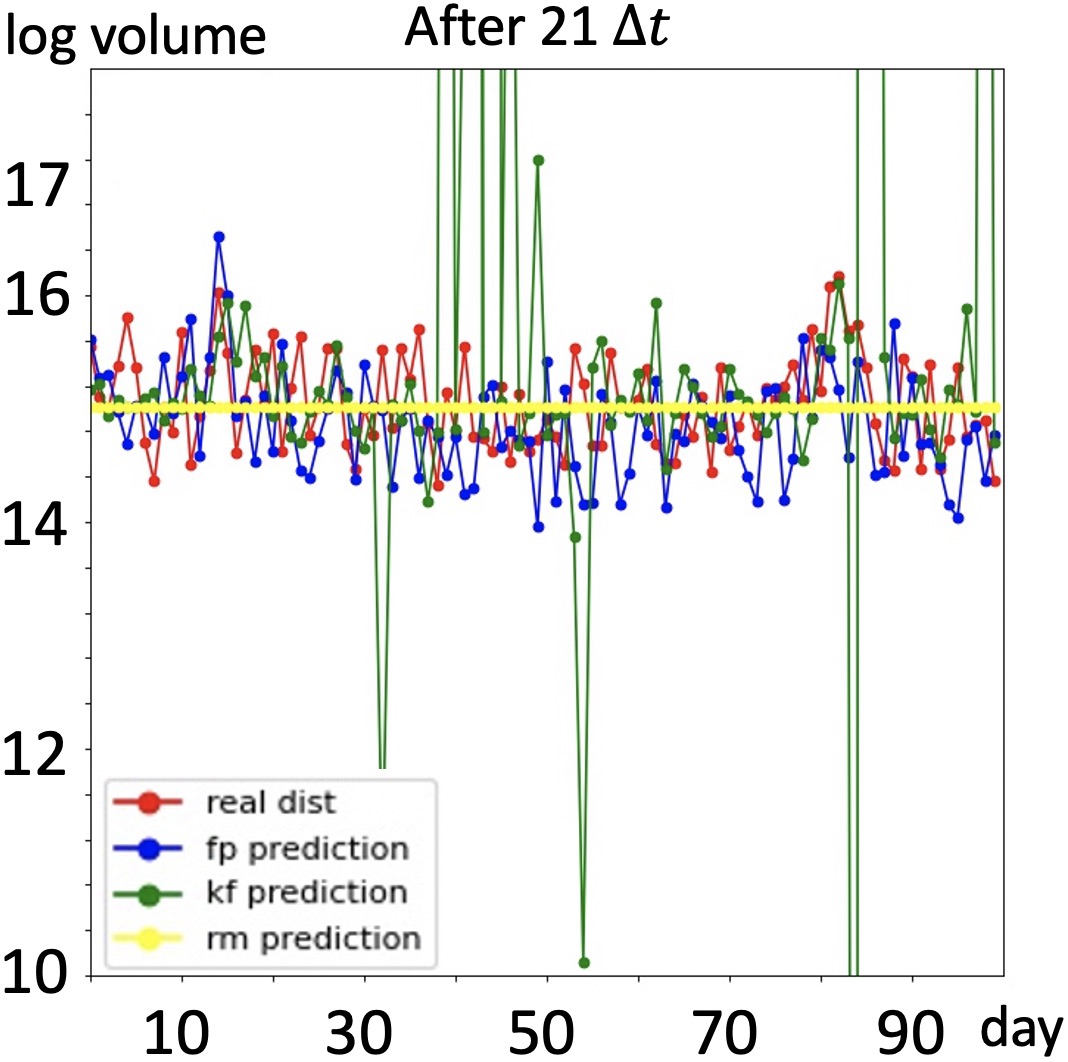}}
\caption{(a) to (d): TSLA stock. (e) to (h): GOOGL stock. We predictions of traded volume in next 100 days, RM(yellow) fails to capture the regularities of traded volume in time series, kalman filter based model(green) fails to capture noise information and make reasonable predictions, our model(blue) is able to seize the movements of traded volume and yield better predictions.}
\label{fig:tradingvolume_add}
\end{figure}

\begin{table}[ht!]
\centering
\caption{The Mean absolute percentage error(MAPE) of different models on
Daily Trading Volume data sets.}
\setlength\tabcolsep{4pt}
\begin{tabular}{l|c|c c c c}
\hline\hline
Stock & Time & RM & KF & Ours
\\ [0.5ex]
\hline
\multirow{4}{*}{JPM} & 14:35  & 0.52 & 0.28 & \textbf{0.01} \\ 
                            &   15:15  & 0.54 & 0.36 & \textbf{0.04} \\
                            &   15:35    & 0.51 & 0.42 & \textbf{0.06} \\
                            &   16:15    & 0.52   & 0.49 & \textbf{0.12} \\
\hline
\multirow{4}{*}{TSLA} &  14:35  & 0.53 & 0.31 & \textbf{0.02} \\
                            &   15:15  & 0.55 & 0.36 & \textbf{0.03} \\
                            &  15:35  & 0.53 & 0.39 & \textbf{0.08} \\ 
                            &  16:15  & 0.52   & 0.38 &\textbf{0.14}\\
\hline
\multirow{4}{*}{GOOGL} & 14:35  & 0.49 & 0.35 &\textbf{0.01} \\
                            &  15:15  & 0.51 & 0.38 &\textbf{0.03}\\
                            &  15:35   & 0.53 & 0.44 &\textbf{0.05}\\
                            &   16:15   & 0.51  & 0.42 & \textbf{0.11} \\
\hline\hline
\end{tabular}
\label{tab:PPer_numerical}
\end{table}

\section{Definition of $\bm{G}$ in Synthetic-2}
\textbf{Synthetic-2~(Nonlinear, converging to mixed-Gaussian):}
\begin{align}
    \thickhat{\bm{x}}_0 \sim \mathcal{N}(0,\bm{\Sigma}_0), \quad \thickhat{\bm{x}}_{t+\Delta t} = \thickhat{\bm{x}}_{t} - \bm{G} \Delta t + \sigma \sqrt{\Delta t}\mathcal{N}(0, 1)\nonumber
\end{align}
\vspace{-2em}
\begin{align}
&\bm{G}_{11} = 
    \frac{1}{\sigma_1}\frac{N_1}{N_1 + N_2}(\thickhat{x}_{t}^1 - \mu_{11}) + \frac{1}{\sigma_2}\frac{N_2}{N_1 + N_2}(\thickhat{x}_{t}^1 - \mu_{21}) \nonumber\\
&\bm{G}_{22} = 
\frac{1}{\sigma_1}\frac{N_1}{N_1 + N_2}(\thickhat{x}_{t}^2 - \mu_{12}) + \frac{1}{\sigma_2}\frac{N_2}{N_1 + N_2}(\thickhat{x}_{t}^2 - \mu_{22})
\nonumber\\
    &N_1 = \frac{1}{\sqrt{2\pi}\sigma_1}\exp\left(-\frac{(\thickhat{x}_{t}^1-\mu_{11})^2}{2\sigma_1^2}-\frac{(\thickhat{x}_{t}^1-\mu_{12})^2}{2\sigma_1^2}\right)\nonumber\\
    &N_2 = \frac{1}{\sqrt{2\pi}\sigma_2}\exp\left(-\frac{(\thickhat{x}_{t}^2-\mu_{21})^2}{2\sigma_2^2}-\frac{(\thickhat{x}_{t}^2-\mu_{22})^2}{2\sigma_2^2}\right)\nonumber
\end{align}

\section{Error Analysis}
In this section, we provide an error analysis of our model. Suppose the hidden dynamics is driven by $g_r(\bm{x})$, the dynamics that we learn from data is $g_f(\bm{x})$, then original It\^{o} process, Euler processes computed by true $g_r$ and estimated $g_f$ are:
\begin{align}
    d\bm{X} &= g(\bm{X})dt + \sigma d\bm{W}\nonumber\\
    \bm{x}_{t+\Delta t}^r &= \bm{x}_{t}^r + g_r(\bm{x}_t^r)\Delta t + \sigma\sqrt{\Delta t} \mathcal{N}(0,1)\nonumber\\
    \bm{x}_{t+\Delta t}^f &= \bm{x}_{t}^f + g_f(\bm{x}_t^f)\Delta t + \sigma\sqrt{\Delta t} \mathcal{N}(0,1)\nonumber
\end{align}
where $\bm{X}$ is the ground truth, $\bm{x}^r$ is computed by true $g_r$ and $\bm{x}^f$ is computed by estimated $g_f$. Estimating the error between original It\^{o} process and its Euler form can be very complex, hence we cite the conclusion from \cite{numericalsde} and focus more on the error between original form and our model.
\begin{customlemma}{2}
With the same initial $\bm{X}_{t_0} = \bm{x}_{t_0} = \bm{x}_0$, if there is a global Lipschitz constant $K$ which satisfies:
\begin{align}
    |g(\bm{x},t) - g(\bm{y},t)| \leq K|\bm{x}-\bm{y}|\nonumber
\end{align}
then after n steps, the expectation error between It\^{o} process $\bm{x}_{t_n}$ and Euler forward process $\bm{x}_{t_n}^r$ is:
\begin{align}
    \mathbb{E}|\bm{x}_{t_n} - \bm{x}_{t_n}^r| \leq K\Biggl(1+\mathbb{E}|X_0|^2\Biggr)^{1/2}\Delta t\nonumber
\end{align}
\label{lem:error}
\end{customlemma}
Lemma \ref{lem:error} illustrates that the expectation error between original It\^{o} process and its Euler form is not related to total steps $n$ but time step $\Delta t$.
\begin{customprop}{3}
With the same initial $\bm{x}_0$, suppose the generalization error of neural network $g$ is $\varepsilon$ and existence of global Lipschitz constant K:
\begin{align}
    |g(\bm{x}) - g(\bm{y})| \leq K|\bm{x}-\bm{y}|\nonumber
\end{align}
then after n steps with step size $\Delta t = T/n$, the expectation error between It\^{o} process $\bm{x}_{t_n}$ and approximated forward process $\bm{x}_{t_n}^f$ is bounded by:
\begin{align}
    \mathbb{E}|\bm{x}_{t_n} - \bm{x}_{t_n}^f| \leq \frac{\varepsilon}{K}(e^{KT}-1) + K(1+\mathbb{E}|\bm{x}_0|^2)^{1/2}\Delta t
\end{align}
\label{prop:error}
\end{customprop}
Proposition \ref{prop:error} implies that besides time step size $\Delta t$, our expectation error interacts with three factors, generalization error, Lipschitz constant of $g$ and total time length. In our experiments, we find the best way to decrease the expectation error is reducing the value of $K$ and $n$.

\section{Proofs}
\subsection{Proof of Proposition 1}

\begin{proof}
Suppose $\thickhat{x}^{(k)}_{t_m}$ and $\thickhat{x}^{(k)}_{t_{m-1}}$ are our observed samples at $t_m$ and $t_{m-1}$ respectively, then expectations could be approximated by:
\begin{align}
\mathbb{E}_{\bm{x}\sim \thickhat{p}(\bm{x},t_m)}[f(\bm{x})]&= \int f(\bm{x})\thickhat{p}(\bm{x},t_m)d\bm{x}
\approx \frac{1}{N}\sum_{k=1}^N f(\thickhat{x}^{(k)}_{t_m})
\label{eqn:prop11}
\end{align}

\begin{align}
\mathbb{E}_{\bm{x} \sim \thicktilde{p}(\bm{x},t_m)}[f(\bm{x})] &= \int^{}f(\bm{x})\thicktilde{p}(\bm{x},t_m)d\bm{x} =\int^{}f(\bm{x})\left[\thickhat{p}(\bm{x},t_{m-1})+\int_{t_{m-1}}^{t_m}\frac{\partial p(\bm{x},\tau)}{\partial t}d\tau\right]d\bm{x} \nonumber\\
&=\int^{}f(\bm{x})\thickhat{p}(\bm{x},t_{m-1})d\bm{x}+\int^{}f(\bm{x})\int_{t_{m-1}}^{t_m}\frac{\partial p(\bm{x},\tau)}{\partial t}d\tau d\bm{x}\nonumber\\
&\approx \frac{1}{N}\sum_{k=1}^N f(\thickhat{\bm{x}}_{t_{m-1}}^{(k)}) + \underbrace{\int f(\bm{x})\int_{t_{m-1}}^{t_m}\left\{-\sum_{i=1}^D\frac{\partial }{\partial x_i} \left[g_{\omega}^i(\bm{x})p(\bm{x},\tau)\right]+\frac{1}{2}\sigma^2\sum_{i=1}^D\frac{\partial^2 }{\partial x_i^2} p(\bm{x},\tau)\right\}d\tau d\bm{x}}_{I}
\label{eq:wd}
\end{align}

Then for the second term $I$ above, it is difficult to calculate directly, but we can use integration by parts to rewrite $I$ as:

\begin{align}
I&=\int_{t_{m-1}}^{t_m}\int^{}\left[\sum_{i=1}^D-f(\bm{x})\frac{\partial }{\partial x_i} g_{\omega}^i(\bm{x})p(\bm{x},\tau)+\frac{1}{2}\sigma^2 \sum_{i=1}^Df(\bm{x}) \frac{\partial^2}{\partial x_i^2}p(\bm{x},\tau)\right]d\bm{x} d\tau \nonumber \\
&=\int_{t_{m-1}}^{t_m}\int^{}\left[\sum_{i=1}^Dg_{\omega}^i(\bm{x})p(\bm{x},\tau)\frac{\partial}{\partial x_i}f(\bm{x})+\frac{1}{2}\sigma^2\sum_{i=1}^Dp(\bm{x},\tau)\frac{\partial^2}{\partial x_i^2}f(\bm{x})\right]d\bm{x} d\tau\nonumber\\
&=\int_{t_{m-1}}^{t_m}\left(\mathbb{E}_{\bm{x}\sim {p}(\bm{x},\tau)}\left[\sum_{i=1}^Dg_{\omega}^i(\bm{x})\frac{\partial}{\partial x_i}f(\bm{x})\right]+\mathbb{E}_{\bm{x}\sim {p}(\bm{x},\tau)}\left[\frac{1}{2}\sigma^2\sum_{i=1}^D\frac{\partial^2}{\partial x_i^2}f(\bm{x})\right]\right)d\tau\nonumber\\
&\approx \int_{t_{m-1}}^{t_m}\frac{1}{N}\sum_{k=1}^N\left(\sum_{i=1}^Dg_{\omega}^i(x^{(k)})\frac{\partial}{\partial x_i}f(x^{(k)})+\frac{1}{2}\sigma^2\sum_{i=1}^D\frac{\partial^2}{\partial x_i^2}f(x^{(k)})\right)d\tau
\end{align}

To approximate the integral from $t_{m-1}$ to $t_m$, we adopt trapezoid rule, then we could rewrite the expectation in Equation (\ref{eq:wd}) as:
\begin{align}
\mathbb{E}_{\bm{x}\sim \thicktilde{p}(\bm{x},t_m)}[f(\bm{x})]&\approx \frac{1}{N}\sum\limits_{k=1}^Nf(\thickhat{x}_{t_{m-1}}^{(k)})+\frac{\Delta t}{2}\left[\frac{1}{N}\sum\limits_{k=1}^N\left(\sum_{i=1}^Dg_{\omega}^i(\thickhat{x}_{t_{m-1}}^{(k)})\frac{\partial}{\partial x_i}f(\thickhat{x}_{t_{m-1}}^{(k)})+\frac{1}{2}\sigma^2\sum_{i=1}^D\frac{\partial^2}{\partial x_i^2}f(\thickhat{x}_{t_{m-1}}^{(k)})\right)\right.\nonumber\\
&~+\frac{1}{N}\sum\limits_{k=1}^N\left.\left(\sum_{i=1}^Dg_{\omega}^i(\thicktilde{x}_{t_m}^{(k)})\frac{\partial}{\partial x}f(\thicktilde{x}_{t_m}^{(k)})+\frac{1}{2}\sigma^2\sum_{i=1}^D\frac{\partial^2}{\partial x_i^2}f(\thicktilde{x}_{t_m}^{(k)})\right)\right]\nonumber\\
&= \frac{1}{N}\sum\limits_{k=1}^Nf(\thickhat{x}_{t_{m-1}}^{(k)}) + \frac{\Delta t}{2}\left[\mathcal{F}_f(\thickhat{X}_{m-1}) + \mathcal{F}_f(\thicktilde{X}_m) \right]
\label{eqn:prop12}
\end{align}
We subtract (\ref{eqn:prop11}) by (\ref{eqn:prop12}) to finish the proof.
\end{proof}

\subsection{Proof of Proposition 2}
\begin{proof}
Given initial $\thickhat{\bm{x}}_{t_0}$, we generate $\thicktilde{\bm{x}}_{t_{1}}$, $\thicktilde{\bm{x}}_{t_{2}}$, $\thicktilde{\bm{x}}_{t_{3}}$ ... $\thicktilde{\bm{x}}_{t_{n}}$ sequentially by Euler-Maruyama scheme. Then the expectations can be rewritten as:
\begin{align}
\mathbb{E}_{\bm{x} \sim \thickhat{p}(\bm{x},t_{n})}[f(\bm{x})]&= \int f(\bm{x})\thickhat{p}(\bm{x},t_{n})d\bm{x}
\approx \frac{1}{N}\sum_{k=1}^N f(\thickhat{x}^{(k)}_{t_{n}})
\label{eqn:prop21}
\end{align}

\begin{align}
    \mathbb{E}_{\bm{x} \sim \thicktilde{p}(\bm{x}, t_{n})}[f(\bm{x})] &\approx \frac{1}{N}\sum_{k=1}^N f(\thickhat{x}_{t_{0}}^{(k)}) + \int_{t_{0}}^{t_{1}} \frac{1}{N}\sum_{k=1}^N\left[\sum_{i=1}^D g_{\omega}^i(x^{(k)})\frac{\partial}{\partial x_i}f(x^{(k)}) + \frac{1}{2}\sigma^2\sum_{i=1}^D\frac{\partial^2}{\partial x_i \partial x_j}f(x^{(k)})\right]d\tau \nonumber \\
    &+\int_{t_{1}}^{t_{2}} \frac{1}{N}\sum_{k=1}^N\left[\sum_{i=1}^D g_{\omega}^i(x^{(k)})\frac{\partial}{\partial x_i}f(x^{(k)}) + \frac{1}{2}\sigma^2\sum_{i=1}^D\frac{\partial^2}{\partial x_i^2}f(x^{(k)})\right]d\tau +... \nonumber \\
     &+\int_{t_{n-1}}^{t_{n}} \frac{1}{N}\sum_{k=1}^N\left[\sum_{i=1}^n g_{\omega}^i(x^{(k)})\frac{\partial}{\partial x_i}f(x^{(k)}) + \frac{1}{2}\sigma^2\sum_{i=1}^n\frac{\partial^2}{\partial x_i^2}f(x^{(k)})\right]d\tau
\end{align}

which is:
\begin{align}
\mathbb{E}_{\bm{x}\sim \thicktilde{p}(\bm{x},t_{n})}[f(\bm{x})]&\approx \frac{1}{N}\sum\limits_{k=1}^Nf(\thickhat{x}_{t_{0}}^{(k)})+ \frac{\Delta t}{2}\left[\mathcal{F}_f(\thickhat{X}_{0}) + \mathcal{F}_f(\thicktilde{X}_1) \right] + \frac{\Delta t}{2}\left[\mathcal{F}_f(\thicktilde{X}_{1}) + \mathcal{F}_f(\thicktilde{X}_2) \right] + ... \nonumber \\
&+ \frac{\Delta t}{2}\left[\mathcal{F}_f(\thicktilde{X}_{n-1}) + \mathcal{F}_f(\thicktilde{X}_n) \right]
\end{align}

Finally it comes to:
\begin{align}
    \mathbb{E}_{\bm{x}\sim \thicktilde{p}(\bm{x},t_{n})}[f(\bm{x})]&\approx 
    \frac{1}{N}\sum\limits_{k=1}^Nf(\thickhat{x}_{t_{0}}^{(k)})+ \frac{\Delta t}{2}\left(\mathcal{F}_f(\thickhat{X}_{0}) + \mathcal{F}_f(\thicktilde{X}_{n}) + 2\sum_{s=1}^{n-1}\mathcal{F}_f(\thicktilde{X}_{s}) \right)
    \label{eqn:prop22}
\end{align}
We subtract (\ref{eqn:prop21}) by (\ref{eqn:prop22}) to finish the proof.
\end{proof}

\subsection{Proof of Error Analysis}
\begin{proof}
The proof process of Lemma \ref{lem:error} is quite long and out of the scope of this paper, for more details please see first two chapters in reference book \citep{numericalsde}. While for the proof of Proposition \ref{prop:error}, with initial $X$ and first one-step iteration:
\begin{align}
    &\begin{cases}
      \bm{x}_{t_0}^r = \bm{x}_{t_0} \\
      \bm{x}_{t_0}^f = \bm{x}_{t_0} \\
    \end{cases}\\
    &\begin{cases}
      \bm{x}_{t_1}^r = \bm{x}_{t_0}^r + g_r(\bm{x}_{t_0}^r)\Delta t + \sigma\sqrt{\Delta t} \mathcal{N}(0,1) \\
       \bm{x}_{t_1}^f = \bm{x}_{t_0}^f + g_f(\bm{x}_{t_0}^f)\Delta t + \sigma\sqrt{\Delta t} \mathcal{N}(0,1) \\
    \end{cases}
\end{align}
Then we have:
\begin{align}
   \mathbb{E}|\bm{x}_{t_0}^r - \bm{x}_{t_0}^f| &= \mathbb{E}|\bm{x}_{t_0} - \bm{x}_{t_0}| = 0 \\
   \mathbb{E}|\bm{x}_{t_1}^r - \bm{x}_{t_1}^f| &= \mathbb{E}|\bm{x}_{t_0}^r - \bm{x}_{t_0}^f + g_r(\bm{x}_{t_0}^r)\Delta t - g_f(\bm{x}_{t_0}^f)\Delta t + \sigma\sqrt{\Delta t} \mathcal{N}(0,1) - \sigma\sqrt{\Delta t} \mathcal{N}(0,1)|\nonumber\\
   &\leq \mathbb{E}|\bm{x}_{t_0}^r - \bm{x}_{t_0}^f| + \mathbb{E}|g_r(\bm{x}_{t_0}^r) - g_f(\bm{x}_{t_0}^f)|\Delta t\nonumber\\
   &=\mathbb{E}|g_r(\bm{x}_{t_0}^r) - g_f(\bm{x}_{t_0}^r) + g_f(\bm{x}_{t_0}^r) - g_f(\bm{x}_{t_0}^f)|\Delta t\nonumber\\
   &\leq \mathbb{E}|g_r(\bm{x}_{t_0}^r) - g_f(\bm{x}_{t_0}^r)|\Delta t + \mathbb{E}|g_f(\bm{x}_{t_0}^r) - g_f(\bm{x}_{t_0}^f)|\Delta t\nonumber\\
   &\leq \varepsilon\Delta t + \mathbb{E}|g_f(\bm{x}_{t_0}^r) - g_f(\bm{x}_{t_0}^f)|\Delta t\nonumber\\
   &=\varepsilon\Delta t + \mathbb{E}|g_f^{\prime}(\bm{x}_{t_0}^{\xi})(\bm{x}_{t_0}^r -\bm{x}_{t_0}^f)|\Delta t \qquad \qquad \qquad (\bm{x}_{t_0}^{\xi} \in [\bm{x}_{t_0}^r, \bm{x}_{t_0}^f])\nonumber\\
   &\leq \varepsilon\Delta t + K\mathbb{E}|\bm{x}_{t_0}^r - \bm{x}_{t_0}^f|\Delta t\nonumber\\
   & = \varepsilon\Delta t
\end{align}

Follow the pattern we have:
\begin{align}
    &\begin{cases}
       \bm{x}_{t_2}^r = \bm{x}_{t_1}^r + g_r(\bm{x}_{t_1}^r)\Delta t + \sigma\sqrt{\Delta t} \mathcal{N}(0,1) \\
       \bm{x}_{t_2}^f = \bm{x}_{t_1}^f + g_f(\bm{x}_{t_1}^f)\Delta t + \sigma\sqrt{\Delta t} \mathcal{N}(0,1) \\
    \end{cases}\\ 
    ...\nonumber\\
    &\begin{cases}
      \bm{x}_{t_n}^r = \bm{x}_{t_{n-1}}^r + g_r(\bm{x}_{t_{n-1}}^r)\Delta t + \sigma\sqrt{\Delta t} \mathcal{N}(0,1) \\
      \bm{x}_{t_n}^f = \bm{x}_{t_{n-1}}^f + g_f(\bm{x}_{t_{n-1}}^f)\Delta t + \sigma\sqrt{\Delta t} \mathcal{N}(0,1)
    \end{cases}
\end{align}

Which leads to:
\begin{align}
    \mathbb{E}|\bm{x}_{t_2}^r - \bm{x}_{t_2}^f| &= \mathbb{E}|\bm{x}_{t_1}^r - \bm{x}_{t_1}^f + g_r(\bm{x}_{t_1}^r)\Delta t - g_f(\bm{x}_{t_1}^f)\Delta t + \sigma\sqrt{\Delta t} \mathcal{N}(0,1) - \sigma\sqrt{\Delta t} \mathcal{N}(0,1)| \nonumber\\
    &\leq \mathbb{E}|\bm{x}_{t_1}^r - \bm{x}_{t_1}^f| + \mathbb{E} |g_r(\bm{x}_{t_1}^r) - g_f(\bm{x}_{t_1}^f)|\Delta t \nonumber\\
    &\leq \mathbb{E}|\bm{x}_{t_1}^r - \bm{x}_{t_1}^f| + \varepsilon \Delta t + K\mathbb{E}|\bm{x}_{t_1}^r - \bm{x}_{t_1}^f|\Delta t\nonumber\\
    &\leq (1+K\Delta t)\varepsilon \Delta t+ \varepsilon \Delta t\\
    ...\nonumber\\
    \mathbb{E}|\bm{x}_{t_n}^r - \bm{x}_{t_n}^f| &\leq \varepsilon \Delta t \sum_{i=0}^{n-1}(1+K\Delta t)^i
\end{align}
Now let $ S = \sum_{i=0}^{n-1}(1+K\Delta t)^i$, then consider followings:
\begin{align}
    S(K\Delta t) &= S(1+K\Delta t) - S\nonumber\\
    &=\sum_{i=1}^{n}(1+K\Delta t)^i - \sum_{i=0}^{n-1}(1+K\Delta t)^i\nonumber\\
    &=(1+K\Delta t)^n - 1\nonumber\\
    &=(1+K\frac{T}{n})^n - 1\nonumber\\
    &\leq e^{KT}-1
\end{align}
Finally we have:
\begin{align}
    \mathbb{E}|\bm{x}_{t_n}^r - \bm{x}_{t_n}^f| \leq \frac{\varepsilon}{K}(e^{KT}-1)
\end{align}

\begin{align}
    \mathbb{E}|\bm{x}_{t_n} - \bm{x}_{t_n}^f| \leq \frac{\varepsilon}{K}(e^{KT}-1) + K(1+E|\bm{x}_0|^2)^{1/2}\Delta t
\end{align}
\end{proof}

\end{document}